%% file: main.tex
\documentclass[10pt,twocolumn,letterpaper]{article}

 \usepackage[pagenumbers]{iccv} 

\usepackage{graphicx}
\usepackage{amsmath}
\usepackage{amssymb}
\usepackage{booktabs}
\usepackage[bold]{hhtensor}
\usepackage{courier}
\usepackage{adjustbox}
\usepackage{listings}

\usepackage{multirow}
\usepackage{color, colortbl}
\newcommand{\blue}[1]{\textcolor{blue}{#1}}
\newcommand{\gray}[1]{\textcolor{gray}{#1}}
\newcommand{\green}[1]{\textcolor{green}{#1}}

\input{preamble}

\definecolor{iccvblue}{rgb}{0.21,0.49,0.74}
\usepackage[pagebackref,breaklinks,colorlinks,allcolors=iccvblue]{hyperref}

\definecolor{azure}{rgb}{0.0, 0.3, 1.0}
\newcommand{\pmerror}[2]{#1\scriptsize{\ensuremath{\pm}}{\scriptsize #2}}
\usepackage{pifont}   
\newcommand{\cmark}{\ding{51}} 
\newcommand{\xmark}{\ding{55}}

\newcommand{\squishlist}{
 \begin{list}{$\bullet$}
  { \setlength{\itemsep}{0pt}
     \setlength{\parsep}{1pt}
     \setlength{\topsep}{1pt}
     \setlength{\partopsep}{0pt}
     \setlength{\leftmargin}{1.5em}
     \setlength{\labelwidth}{1em}
     \setlength{\labelsep}{0.5em} } }
\newcommand{\squishend}{
  \end{list}  }

\title{AdaptCLIP: Adapting CLIP for Universal Visual Anomaly Detection  
}

\author{
\quad Bin-Bin Gao\textsuperscript{1} 
\quad Yue Zhou \textsuperscript{2,3} 
\quad Jiangtao Yan \textsuperscript{1}
\quad Yuezhi Cai \textsuperscript{2} \\
\quad Weixi Zhang \textsuperscript{2} 
\quad Meng Wang \textsuperscript{2} 
\quad Jun Liu \textsuperscript{1} 
\quad Yong Liu \textsuperscript{1} 
\quad Lei Wang \textsuperscript{2} 
\quad Chengjie Wang\textsuperscript{1,4} \\
\normalsize \textsuperscript{1}{Tencent YouTu Lab} \quad  
\textsuperscript{2}{Siemens AG}   \quad 
\textsuperscript{3}{Technical University of Munich} \quad 
\textsuperscript{4}{Shanghai Jiao Tong University} \\
}

\begin{document}

\maketitle

\input{sec/0_abstract}    
\input{sec/1_intro}
\input{sec/2_rework}
\input{sec/3_method}

\input{sec/4_exps}
\input{sec/5_conclusion}

{
    \small
    \bibliographystyle{ieeenat_fullname}
    \bibliography{main}
}

\input{sec/6_suppl}

\end{document}

%% file: preamble.tex
\newcommand{\red}[1]{{\color{red}#1}}


%% file: sec/0_abstract.tex
\begin{abstract}
Universal visual anomaly detection aims to identify anomalies from novel or unseen vision domains without additional fine-tuning, which is critical in open scenarios. Recent studies have demonstrated that pre-trained vision-language models like CLIP exhibit strong generalization with just zero or a few normal images. However, existing methods struggle with designing prompt templates, complex token interactions, or requiring additional fine-tuning, resulting in limited flexibility. In this work, we present a simple yet effective method called AdaptCLIP based on two key insights. First, adaptive visual and textual representations should be learned alternately rather than jointly. Second, comparative learning between query and normal image prompt should incorporate both contextual and aligned residual features, rather than relying solely on residual features. AdaptCLIP treats CLIP models as a foundational service, adding only three simple adapters, visual adapter, textual adapter, and prompt-query adapter, at its input or output ends. AdaptCLIP supports zero-/few-shot generalization across domains and possesses a training-free manner on target domains once trained on a base dataset. AdaptCLIP achieves state-of-the-art performance on 12 anomaly detection benchmarks from industrial and medical domains, significantly outperforming existing competitive methods. We will make the code and model of AdaptCLIP available at \url{https://github.com/gaobb/AdaptCLIP}.
\end{abstract}

%% file: sec/1_intro.tex
\section{Introduction}
\label{sec:intro}

Universal visual anomaly detection (AD) aims to identify anomaly images and segment anomaly pixels from novel or unseen visual objects after learning a single model on a base or seen dataset. This is a more challenging task as it requires strong generalization when facing cross-domain datasets. Meanwhile, it is a more practical topic as people are more interested in fast adaptability in real-world scenarios, especially in low data regimes (i.e., few-shot and even zero-shot). For example, in medical image diagnosis and industrial visual quality inspection, it is difficult to collect a large-scale dataset due to inherent scarcity and privacy protection. Recently, developing universal visual AD has attracted increasing attention because existing unsupervised ADs with either separated~\cite{cvpr2022patchcore,cvpr2022rd,cvpr2023simplenet} or unified models~\cite{neurips2022uniad, eccv2024onenip} perform poorly in unseen objects despite promising performance on seen objects.

\begin{figure}[htbp]
    \vspace{-8pt}
    \centering
    \begin{minipage}[b]{0.23\textwidth}
        \centering
        \setlength\tabcolsep{0.2pt}
        \small
        \vspace{-5pt}
        \resizebox{\textwidth}{!}{%
        \begin{tabular}{c cccc}
            \toprule
            Methods &\scriptsize{ZS} &\scriptsize{FS} &\scriptsize{OA} &\scriptsize{w/o FT} \\
            \toprule
            WinCLIP~\cite{cvpr2023winclip} & \red{\cmark} & \red{\cmark} & \red{\cmark} & \red{\cmark} \\
            AdaCLIP~\cite{eccv24adaclip} & \red{\cmark} & \green{\xmark} & \green{\xmark} & \red{\cmark} \\
            InCtrl~\cite{cvpr2024inctrl} & \green{\xmark} & \red{\cmark} & \red{\cmark} & \red{\cmark} \\
            AnomalyCLIP~\cite{iclr2024anomalyclip} & \red{\cmark} & \green{\xmark} & \green{\xmark} & \red{\cmark} \\
            PromptAD~\cite{cvpr2024promptad} & \green{\xmark} & \red{\cmark} & \red{\cmark} & \green{\xmark} \\
            \rowcolor[HTML]{EFEFEF} \textbf{AdaptCLIP} & \red{\cmark} & \red{\cmark} & \red{\cmark} & \red{\cmark} \\
            \bottomrule
        \end{tabular}
        }
        \vspace{-5pt}   
        \label{tab:subtable1}
    \end{minipage}
    \hspace{-5pt}   
    \begin{minipage}[b]{0.24\textwidth}
        \centering
        \includegraphics[width=1.0\textwidth, keepaspectratio]{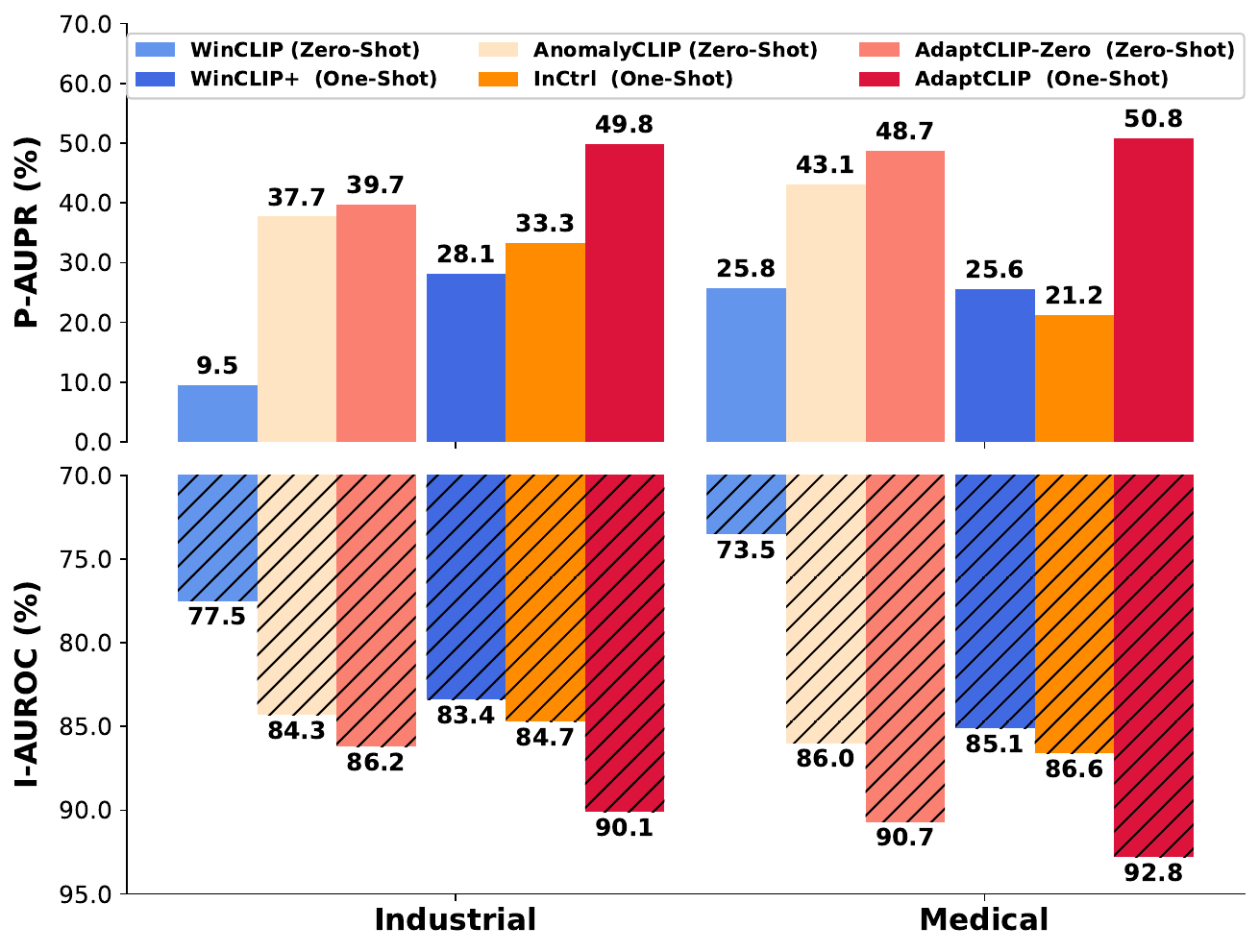}
        \vspace{-20pt}  
    \end{minipage}
    \hspace{-8pt}
    \caption{\small{Comparisons of state-of-the-arts and our \textbf{AdaptCLIP}. \red{\cmark}~means satisfied and \green{\xmark}~means not satisfied. Our method supports zero-/few-shot~(ZS and FS) visual AD across different domains without fine-tuning~(FT) on the target dataset. 
    It only adds simple adapters at CLIP's input or output ends without complex token interactions, thus preserving CLIP’s original ability (OA).
    The AdaptCLIP using only one normal image prompt achieves the best performance in image-level anomaly classification (I-AUROC) and pixel-level anomaly segmentation (P-AUPR) on 12 AD benchmarks from industrial and medical domains. Moreover, the zero-shot AdaptCLIP is also significantly better than existing zero-shot and even some one-shot approaches. The detailed results are reported in Tabs.~\ref{tab:sota_auroc} and ~\ref{tab:sota_aupr}. Best viewed in zoom.}}
    \label{fig:comzerofewshot}
\end{figure}

To address this fragmentation, recent works have attempted to design universal models to recognize anomalies for unseen objects. They typically build on vision-language models (i.e., CLIP~\cite{icml2021clip}) benefiting from strong generalization. 
WinCLIP~\cite{cvpr2023winclip} computes anomaly scores on dense patch windows. This brings large computational costs and memory burden, limiting high-resolution input or large pre-trained models. AnomalyCLIP~\cite{iclr2024anomalyclip} learns class-agnostic prompt embeddings to align patch-wise tokens thus avoiding dense window operations. It further refines vanilla CLIP by concatenating learnable tokens to intermediate layers of CLIP. AdaCLIP~\cite{eccv24adaclip} further integrates visual knowledge into textual prompt embeddings. However, they may destroy inherent representations of CLIP. Therefore, \emph{we want to explore whether we can achieve the same or even better AD performance while maintaining the original ability.}

In contrast, humans perceive anomalies when an input significantly deviates from those normal patterns stored in our brains. There is evidence to support this point in neuroscience~\cite{predictivecoding}. PatchCore~\cite{cvpr2022patchcore} builds a memory bank storing normal features and PaDiM~\cite{icpr2021padim} learns a multivariate Gaussian distribution of normal features. At inference, anomalies are recognized by comparing input features with the memory bank or the learned distribution. However, these methods usually require a certain number of normal images and thus are limited in universal (i.e., open-world) scenarios. Two recent works, i.e., InCtrl~\cite{cvpr2024inctrl} and PromptAD~\cite{cvpr2024promptad}, have studied how to further improve performance with few-shot normal images. However, InCtrl only considers anomaly classification, while PromptAD needs to learn a new model for each class. Different from them, \emph{we want to comprehensively explore a universal AD model, aiming to detect any anomalies in image-level and pixel-level from cross-domains without any training on target domains.} 

Toward this end, we propose a simple but effective universal visual anomaly detection framework, called AdaptCLIP. \emph{The philosophy of AdaptCLIP is that ``less and simpler could be better", and it contains three adapters designed by two key insights :} 
First, adaptive visual and textual representations should be learned alternately rather than jointly. Second, comparative learning between query and the corresponding normal image prompt should incorporate both contextual and aligned residual features, rather than relying solely on residual features. Our contributions are summarized as follows.

\squishlist 
\item We propose a simple but effective universal visual anomaly detection framework based on visual-language CLIP models, which is capable of detecting any visual anomalies at image- and pixel-level from cross-domain datasets without any training on target domains.

\item We propose visual and textual adapters, and find that they should alternately learn adaptive visual and textual representation guided by the powerful vision-language representations from CLIP models.

\item We propose a prompt-query adapter that aims to capture meta-perceptual capabilities between query image and the corresponding normal image prompt, based on their joint distribution of contextual features of the query and the aligned residual features between prompt and query.

\item AdaptCLIP outperforms zero- and few-shot AD methods on 8 industrial and 4 medical benchmarks, as shown in Fig.~\ref{fig:comzerofewshot}. Meanwhile, AdaptCLIP possesses simpler adapters, fewer parameters, and competitive efficiency.
\squishend

%% file: sec/2_rework.tex
\section{Related Works}
\label{sec:rework}

\textbf{Unsupervised ADs} target to identify anomalies given sufficient normal training images. Most unsupervised AD methods can be roughly grouped into three categories: embedding-, discrimination-, and reconstruction-based methods. Embedding-based methods, such as PaDiM~\cite{icpr2021padim}, MDND~\cite{icpr2021mdnd}, PatchCore~\cite{cvpr2022patchcore}, CS-Flow~\cite{wacv2022csflow} and PyramidFlow~\cite{cvpr2023pyramidflow}, assume that offline features extracted from a pre-trained model preserve discriminative information and thus help to separate anomalies from normal samples. Discrimination-based methods, such as CutPaste~\cite{cvpr2021cutpaste}, DRAEM~\cite{iccv2021draem}, and SimpleNet~\cite{cvpr2023simplenet}, typically convert unsupervised AD to supervised ones by introducing pseudo~(synthesized) anomaly samples. 
Reconstruction-based ADs, such as autoencoder~\cite{icml2008demae,neurips2013gdae,iccv2019memorize,iccv2021divideassemble}, generative adversarial networks~\cite{cvpr2019ocgan,aaai2021scadn,cvpr2020oldgold} and reconstruction networks~\cite{pr2021reconstruction,cvpr2022sspcab,cvpr2023diversitymeas}, assume that anomalous regions should not be able to be properly reconstructed and thus result in high reconstruction errors since they do not exist in normal training samples. The recent knowledge distillation~\cite{cvpr2020us, cvpr2021glancing,bmvc2021stfpm,cvpr2021mkd,cvpr2022rd} or feature reconstruction methods~\cite{neurips2022uniad, cvpr2023omnial,iccv2023focus,eccv2024onenip} train a student or reconstruction network to match a fixed pre-trained teacher network and achieve a good balance between effectiveness and efficiency. However, all these methods are limited to recognizing anomalies of seen classes but often perform poorly on unseen classes. For a novel scenario, people have to collect sufficient normal images first and then retrain a model. This is inefficient and lacks the rapid adaptability required for practical applications.

\noindent\textbf{Zero-Shot ADs} have achieved impressive performance by utilizing large vision-language models, e.g., CLIP~\cite{icml2021clip}. WinCLIP~\cite{cvpr2023winclip} designs two-class textual prompts and introduces multi-scale patch windows for accurate anomaly segmentation. It brings large computational costs and memory burden, limiting high-resolution input or large pre-trained models. AnomalyCLIP~\cite{iclr2024anomalyclip} learns class-agnostic prompt embeddings to align patch-wise tokens thus avoiding dense window operation. In addition, AnomalyCLIP refines vanilla CLIP representation by appending some learnable tokens to the middle layer of CLIP. Recently, AdaCLIP~\cite{eccv24adaclip} and VCP-CLIP~\cite{eccv24vcpclip} utilize similar ideas and further integrate visual knowledge into textual prompt embeddings. We argue that these additional operations make models more complex and may hurt the original capabilities of CLIP. Instead of visual-language models, ACR~\cite{nips2023acr} and MuSc~\cite{iclr2024musc} perform zero-shot AD only requiring batch-level and full-shot testing images, but they may be limited in privacy protection scenarios. Different from these methods, we explore whether the same or even better AD performance is achieved while retaining the original ability of CLIP without any information on test data distribution.

\begin{figure*}[h]
  \centering
  \vspace{-10pt}
\includegraphics[width=0.95\textwidth,keepaspectratio]{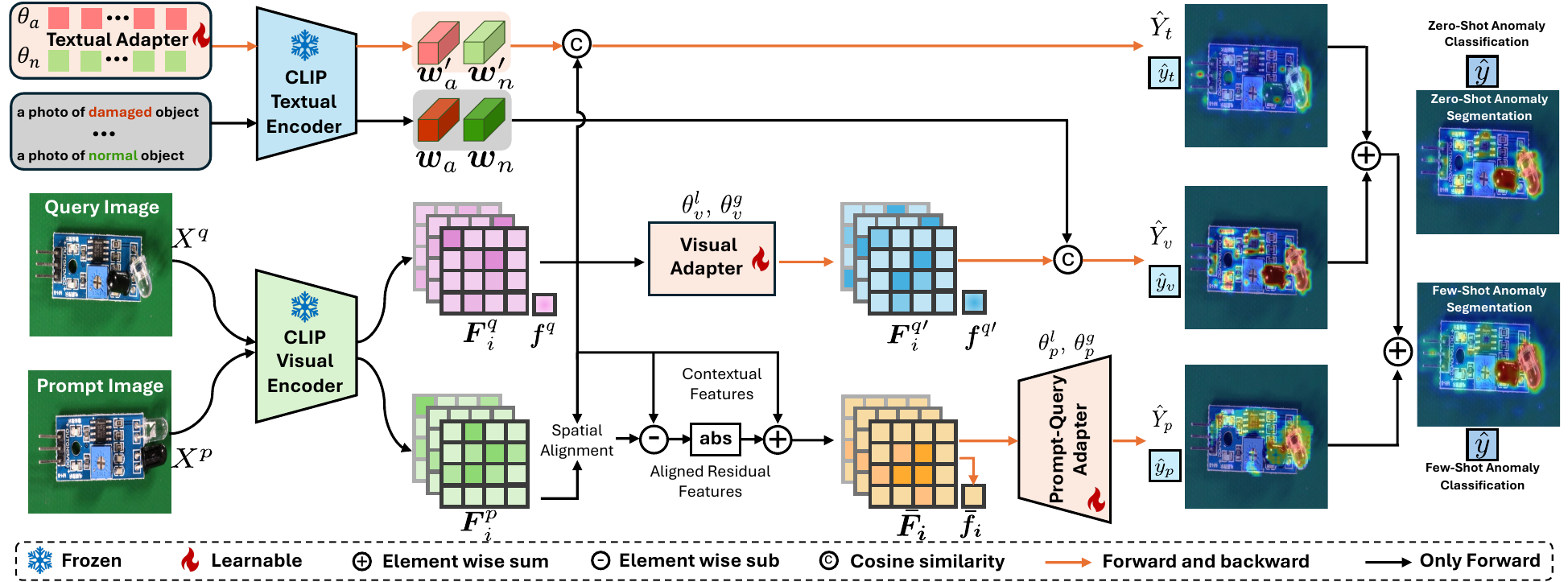} 
\vspace{-5pt}
  \caption{The framework of \textbf{AdaptCLIP}, which consists of three pluggable adapters, i.e., visual adapter, textual adapter, and prompt-query adapter. First, the first two adapters alternately learn visual and textual representations for zero-shot anomaly detection~(Sec.~\ref{subsec:va}). The prompt-query adapter further learns a comparison ability between query image and its corresponding normal prompt for few-shot anomaly detection~(Sec.~\ref{subsec:pqa}). Once trained, it can segment any anomalies providing only few-shot and even zero-shot normal image prompts.}
  \label{fig:AdaptCLIP}
  \vspace{-10pt}
\end{figure*}

\noindent\textbf{Few-Shot ADs} mainly pay attention to learning or using only a limited number of normal images, such as TDG~\cite{iccv2021tdg}, RegAD~\cite{eccv2022regad}, GraphCore~\cite{iclr2023graphcore} and FastRecon~\cite{iccv2023fastrecon}. Some works~\cite{cvpr2022catching,cvpr2023bgad} consider another few-shot setting where a limited number of samples is given from anomaly images. The performance of these methods lags behind unsupervised ADs. Recently, few-shot AD performance has been improved significantly by visual-language models. 
WinCLIP+~\cite{cvpr2023winclip} is the first work to apply CLIP models to few-shot AD, which stores normal tokens into a memory bank, then retrieves the nearest token for each query token using cosine similarity, and finally computes an anomaly map using the nearest distance. InCtrl~\cite{cvpr2024inctrl} further integrates multi-level information, including patch-level residual maps and image-level residual features, and prior knowledge score using two-class textual prompts, to learn a holistic scoring function for anomaly classification. It does not consider pixel-level anomaly segmentation. PromptAD~\cite{cvpr2024promptad} introduces the concept of explicit anomaly margin, which mitigates the training challenge caused by the absence of anomaly training images. However, it requires re-training models when applied to target datasets. In contrast, we explore jointly optimizing anomaly classification and segmentation in a unified model, which can quickly adapt to novel scenarios only given few-shot normal image prompts, not involving additional re-training.

%% file: sec/3_method.tex
\section{Methods}\label{sec:methods}
\noindent\textbf{Problem Formulation}:
Our objective is to learn a universal AD model that detects any anomalies from diverse domains without any training on target dataset.
Thus, a reasonable assumption is that there is a different distribution between training and testing sets. 
Formally, let $\mathcal{D}_\text{base} =\{X_i, Y_i, y_i\}_{i=1}^N$ be a training dataset, that consists of $N$ normal and anomalous images, $X_i \in \mathcal{R}^{h \times w \times 3}$ is the $i$-th image, and $Y_i \in \mathcal{R}^{h \times w}$ and $y_i=\{0,1\}$ is the corresponding anomaly mask and anomaly label, with $y_i = 0$ indicates normal and $y_i = 1$ signifies anomaly. The testing set $\mathcal{T}$ may consist of multiple different domains with various objects and anomaly types. Here, we denote the $t$-th novel domain as $\mathcal{D}_\text{novel}^t=\{X_i, Y_i, y_i\}_{i=1}^{N_t}$. Under a few-shot setting, a few normal images $\mathcal{P}_c = \{X_i\}_{i=1}^k$ are randomly drawn from each class of the target domain, where $c$ is the class index and $k$ is typically a small number, e.g., $k=\{1,2,4\}$. It is worth noting that $\mathcal{P}_c$ is only available during inference, and cannot be used in any way during training phase.

\noindent\textbf{Overview}: 
As illustrated in Fig.~\ref{fig:AdaptCLIP}, the visual adapter adapts patch and image tokens with fixed two-class textual prompt embeddings. The textual adapter learns two-class prompt embeddings to align with the fixed patch and image tokens. The prompt-query adapter operates in a one-prompt meta-learning manner, leveraging the joint distribution of query context features and the aligned residual features between the prompt and query.
In a zero-shot scenario, image-level anomaly score and pixel-level anomaly map can be obtained using textual and visual adapters (Sec.~\ref{subsec:va}). In a few-shot scenario, anomaly score and map are derived by integrating predictions from zero-shot and prompt-query adapters~(Sec.~\ref{subsec:pqa}).
Below we present them in detail.

\subsection{Revisiting CLIP for Anomaly Detection}\label{sec:}
For a query image $X^q\in \mathcal{R}^{h \times w \times 3}$, we feed it to visual encoder $\mathcal{F}(\cdot)$ and obtain local patch tokens $\{\vec F^q_{i} \in \mathcal{R}^d\}_{~i=1}^{hw/p^2}$ and global image token $\vec f^q \in \mathcal{R}^d$, where $p$ is patch size. WinCLIP~\cite{cvpr2023winclip} introduces two-class prompts describing normal and abnormal states. For example, ``a photo of a normal object" and ``a photo of a damaged object". In practical application, one could design multiple textual descriptions for normal and abnormal states. Feeding these normal and abnormal descriptions to textual encoder $\mathcal{T(\cdot)}$, we can obtain the embeddings of normal $\vec w_n\in \mathcal{R}^d$ and abnormal $\vec w_a\in \mathcal{R}^d$. The pixel-level anomaly map is computed by measuring the cosine similarities between all patch tokens and the textual embeddings, that is
\begin{equation}\label{eq:localsoftmax}
 \hat Y = \big{[} 
\frac{\exp (\langle\vec w_a, \vec F^q_{i} \rangle)} {\exp (\langle \vec w_a, \vec F^q_{i}\rangle) + \exp (\langle\vec w_n, \vec F^q_{i}\rangle)}\big{]}, 
\end{equation}
where $\langle \cdot \rangle$ represents the cosine similarity, and $[\cdot]$ means that all patch-wise prediction scores are rearranged according to their spatial positions and interpolated to the original input resolution.
Replacing $\vec F^q_{i}$ with $\vec f^q$ in Eq.~\ref{eq:localsoftmax}, we can obtain an image-level anomaly score $\hat y$ for $X^q$, that is
\begin{equation}\label{eq:globalsoftmax}
\hat y = 
 \frac{\exp (\langle\vec w_a, \vec f^q \rangle)} {\exp (\langle \vec w_a, \vec f^q \rangle) + \exp (\langle\vec w_n, \vec f^q \rangle)}.
\end{equation}

\subsection{AdaptCLIP with Alternating Learning }\label{subsec:va}
To adapt CLIP for universal visual anomaly detection, we design visual and textual adapters to alternately learn visual and textual representations.
Specifically, the visual adapter learns adaptive visual tokens ($\vec F^{q\prime}_{i}$ and $\vec f^{q\prime}$) when fixing two-class static textual embeddings ($\vec w_a$ and $\vec w_n$), while the textual adapter learns two-class textual prompt embeddings ($\vec w_a^\prime$ and $\vec w_n^\prime$) when fixing visual tokens  ($\vec F^q_{i}$ and $\vec f^q$). 

\noindent\textbf{Visual Adapter} adapts vision tokens ($\vec F^q_{i}$ and $\vec f^q$) with fixed textual embeddings ($\vec w_a$ and $\vec w_n$).
It consists of two branches, global and local, which transform global image token and local patch tokens, respectively. Architecturally, the global and local branches are implemented using a simple residual multi-layer perception~(MLP), that is
\begin{equation}\label{eq:resmlp}
\vec F^{q\prime}_{i} = 
\vec F^q_{i} + \text{MLP}(\vec F^q_{i};\theta_v^l); \\
\vec f^{q\prime} = \vec f^q + \text{MLP}(\vec f^q; \theta_v^g),
\end{equation}
where $\theta_v^l$ and $\theta_v^g$ are learnable parameters. Replacing $\vec F^q_{i}$ and $\vec f^q$ in Eqs.~\ref{eq:localsoftmax} and \ref{eq:globalsoftmax} with $\vec F^{q\prime}_{i}$ and $\vec f^{q\prime}$, we obtain pixel-level anomaly map $\hat Y_v$  and image-level anomaly score $\hat y_v$.

\noindent\textbf{Textual Adapter} aims to directly learn two-class prompts $\vec \theta_a,  \vec \theta_n \in \mathcal{R}^{r \times d}$ without prompt templates, where $r>0$ is the length of prompts. We feed them into the frozen textual encoder $\mathcal{T}(\cdot)$ of CLIP, and obtain the corresponding embeddings $\vec w_a^\prime$ and $\vec w_n^\prime$, that is 
\begin{equation}\label{eq:lpt}
    \vec w_a^\prime = \mathcal{T}(\vec \theta_a),
    \vec w_n^\prime = \mathcal{T}(\vec \theta_n).
\end{equation}
Then, we replace the static $\vec w_a$ and $\vec w_n$ in Eqs.~\ref{eq:localsoftmax} and \ref{eq:globalsoftmax} with the learnable prompt embeddings $\vec w_a^\prime$ and $\vec w_n^\prime$ to derive local and global anomaly predictions, $\hat Y_t$ and $\hat y_t$.

\noindent\textbf{Alternating Learning or Joint Learning?}
A possible question is whether we can learn visual and textual representations jointly. That is, in Eqs.~\ref{eq:localsoftmax} and \ref{eq:globalsoftmax}, we simultaneously replace fixed textual embeddings and visual tokens with learnable prompt embeddings ($\vec w_a^\prime$ and $\vec w_n^\prime$) and adaptive visual tokens ($\vec F^{q\prime}_{i}$ and $\vec f^{q\prime}$). Indeed, this joint alignment mechanism is successful when a large-scale image-text dataset is available.
However, we empirically find that it does not work well in the AD field, as shown in Tab.~\ref{tab:ab} (Lines \gray{3} vs. \gray{4}). This is not surprising because the available training data scale is still relatively small and lacks fine-grained textual annotations. The joint learning easily overfits and leads to poor generalization on novel datasets. In contrast, the alternating learning helps us fully utilize the prior knowledge of the CLIP model and thus improve the cross-domain generalization.

\subsection{AdaptCLIP with Comparative Learning}\label{subsec:pqa}
Compared to static or learnable textual prompts, using a normal image as a visual prompt is more intuitive. 
Therefore, we expect to learn a comparison ability between a query image $X^q$ and its corresponding normal prompt $X^p$, which generalizes well to unseen objects. We find that applying multi-layer features yields better results. For simplicity, we use a single-layer feature in the following.

\noindent\textbf{Spatial Alignment:} 
A simple way is to directly measure their difference by the absolute value of their residual feature, that is $|\vec F^q_{i} -\vec F^p_{i}|$, where $\vec F^q_{i}$ and $\vec F^p_{i}$ are the patch token of $X^q$ and $X^p$, respectively.
It may fail if the query and prompt images are not aligned in pixel space~(e.g., due to rotation and translation). 
Therefore, we have to align query and prompt tokens for effective comparison. 
For any query token $\vec F_{i}^q$, we search the nearest one among all normal tokens $\{\vec F_{j}^p\}_{~j=1}^{hw/p^2}$ using euclidean distance, that is
\begin{equation}\label{eq:align}
    \vec F_{i}^{p\prime} = \vec F^p_k, k= \arg \min_{j}  \|\vec F_{i}^q - \vec F_{j}^p \|_2.
\end{equation}
Then, we take $\vec F_{i}^{p\prime}$ as aligned prompt token of $\vec F_{i}^q$. Now, we can derive the aligned residual feature, i.e., $|\vec F_{i}^q - \vec F_{i}^{p\prime}|$.

\noindent\textbf{Joint contextual and aligned residual feature:} 
The aligned residual feature highlights differences or anomaly regions well. However, it may lose contextual information or introduce noise. Intuitively, the contextual information is critical to identify anomalies. Therefore, we aggregate the original query tokens and the aligned residual features by an element-wise sum,
\begin{equation}\label{eq:absfeat}
    \vec {\bar {F}}_{i} =  \vec F_{i}^q + |\vec F_{i}^q -\vec F_{i}^{p\prime}|.
\end{equation}

\noindent\textbf{Prompt-Query Adapter:} 
The ultimate goal is to achieve pixel-level anomaly segmentation and image-level anomaly classification. Therefore, we propose a lightweight segmentation head $\mathcal{G}(\cdot; \theta_p^l)$ to learn anomaly segmentation based on the joint feature $\bar F$, that is 
\begin{equation}
    \hat Y_p = \mathcal{G} (\bar F; \theta_p^l),
\end{equation}
where $\theta_p^l$ is its parameters.
Specifically, the segmentation head consists of several transposed convolution blocks following a 1$\times$1 convolution layer. Here, each transposed convolution block upsamples input feature by 2$\times$, and it is composed of a 3$\times$3 convolution, a BatchNorm, a ReLU, and a 2$\times$2 deconvolution. 

Meanwhile, we need to obtain a global image-level prediction. First, we perform average-pooling and max-pooling on the joint feature $\bar F$ along the spatial dimension and then take their weighted average as the global image representation. Then, a simple MLP is used to map the global feature to an image-level prediction score, that is 
\begin{equation}
    \hat y_p = \text{MLP} \big((\text{AvgPool}(\bar F) + \text{MaxPool}(\bar F))/2; \theta_p^g\big),
\end{equation}
where $\theta_p^g$ is the parameter.

\begin{figure}[h]
\includegraphics[width=0.48\textwidth,keepaspectratio]{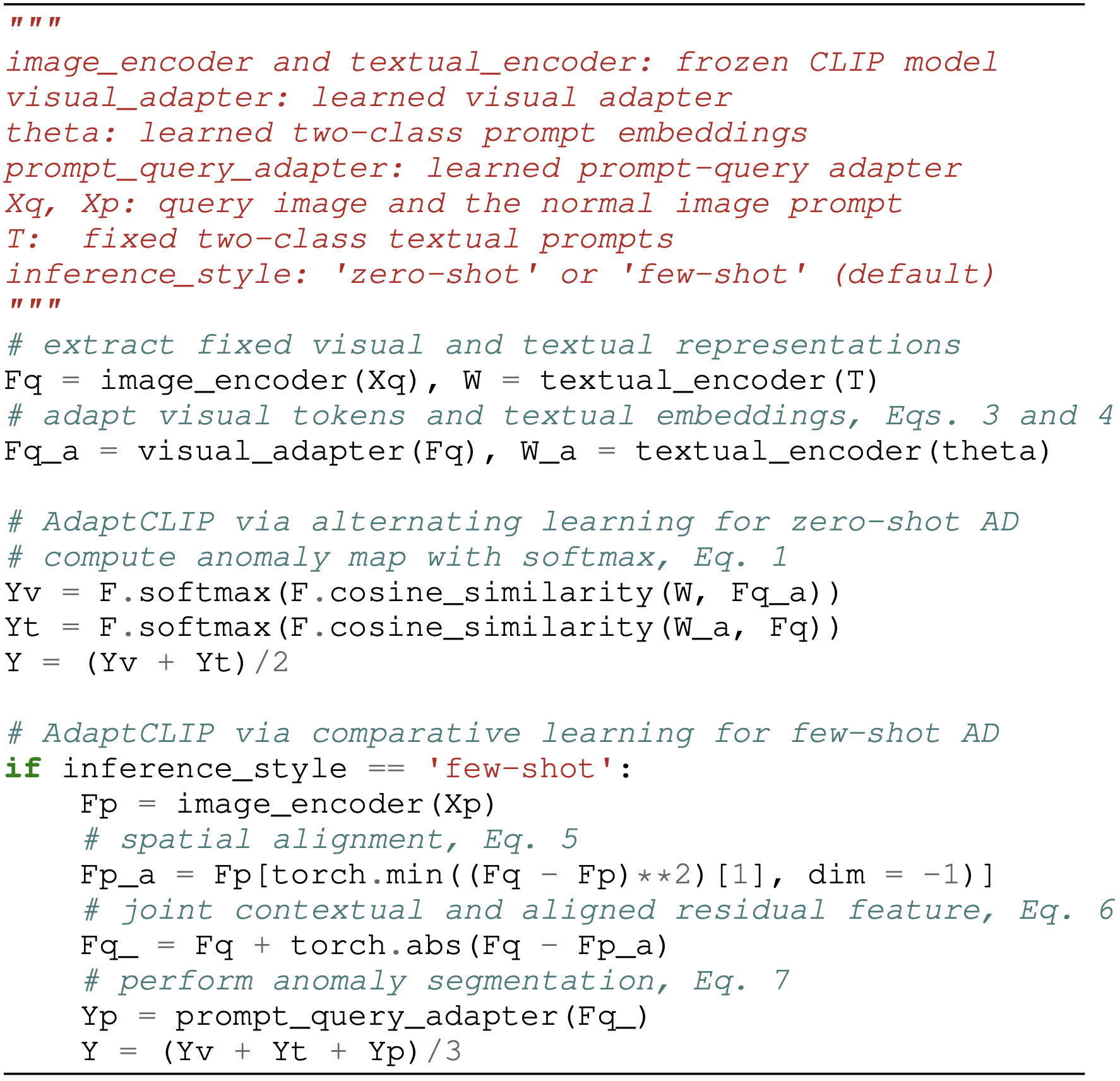} 
\vspace{-15pt}
\caption{PyTorch pseudocode for the inference of AdaptCLIP.}\label{fig:pscodeadaptclip}
\vspace{-5pt}
\end{figure}

\subsection{Training and Inference}
During training, we use cross-entropy loss for global image anomaly classification, and Focal and Dice losses for local patch anomaly segmentation, which is exactly the same as AnomalyCLIP~\cite{iclr2024anomalyclip}.
For zero-shot inference, we average the predictions from visual and textual adapters. For few-shot inference, we fuse (i.e., average) all results from three adapters, i.e., prompt-query, visual and textual adapters, as the final predictions of AdaptCLIP. Fig.~\ref{fig:pscodeadaptclip} shows PyTorch pseudocode for the pixel-level inference of AdaptCLIP, where $W=[{\vec w_a}^T; {\vec w_n}^T]$, $W_a = [{\vec w_a^\prime}^T; {\vec w_n^\prime}^T]$, $Fq$ and $Fp$ means the fixed $\vec F^q_i$ and $\vec F^p_i$, and $Fq_a$ and $Fp_a$ refers the adaptive $\vec F^{q\prime}$ and the aligned $\vec F^{p^\prime}$. Here, we omit the image-level inference since it can be easily obtained by replacing local patch tokens with a global image token.

%% file: sec/4_exps.tex
\section{Experiments}
\subsection{Experimental Setup}
\noindent\textbf{Datasets:}
We comprehensively evaluate AdaptCLIP on multiple datasets from industrial and medical domains.
For industrial domain, we use MVTec~\cite{cvpr2019mvtec}, VisA~\cite{eccv2022visa}, BTAD~\cite{isie2021btad}, MVTec3D~\cite{visapp2022mvtec3d}, DTD~\cite{wacv2023dtd}, KSDD~\cite{jim2020ksdd}, MPDD~\cite{icumt2021mpdd}, and large-scale Real-IAD~\cite{cvpr2024realiad}. 
In medical domain, we utilize brain tumor detection datasets, Br35H~\cite{Br35h} and COVID-19~\cite{cbm20covid19}, as well as gastrointestinal polyp datasets,~Kvasir~\cite{icmm2020kvasir} and Endo~\cite{jhe2017endo}. A detailed introduction to these datasets can be found in \texttt{Appendix}.

\noindent\textbf{Evaluation Metrics:} 
Following previous works, we use AUROC for image-level anomaly classification and AUPR for pixel-level anomaly segmentation in our main paper. Here, we emphasize that AUPR is better for anomaly segmentation, where the imbalance issue is very
extreme between normal and anomaly pixels~\cite{eccv2022visa}. In \texttt{Appendix}, we also provide detailed comparisons using all metrics, including AUROC, AUPR, and F1$_\text{max}$.

\noindent\textbf{Training and Testing Protocol:} 
Following AnomalyCLIP~\cite{iclr2024anomalyclip}, we train AdaptCLIP using the testing data from MVTec and evaluate zero-/few-shot performance on other datasets. As for the evaluation of MVTec, we train AdaptCLIP using the testing data of VisA. For fair comparison, all models are trained and evaluated using the same protocol.

\noindent\textbf{Competing Methods:} We compare our AdaptCLIP with diverse state-of-the-art zero-/few-shot AD methods including zero-shot WinCLIP~\cite{cvpr2023winclip}, AnomalyCLIP~\cite{iclr2024anomalyclip}, AdaCLIP~\cite{eccv24adaclip}, and few-shot WinCLIP+~\cite{cvpr2023winclip}, 
InCtrl~\cite{cvpr2024inctrl} and
AnomalyCLIP+. Here, AnomalyCLIP+ is a strong baseline we build on AnomalyCLIP~\cite{iclr2024anomalyclip} by adding patch-level feature associations like WinCLIP+. More implementation details about AdaptCLIP and competing methods can be found in \texttt{Appendix}.

\begin{table*}[t]
\setlength\tabcolsep{3pt}
\centering
\small
\caption{\small\textbf{Image-level anomaly classification comparisons with AUROC metric on industrial and medical domains.} The best and second-best results are highlighted in~\textcolor{red}{red} and \textcolor{blue}{blue}, respectively. The superscript$^\dag$
indicates that the results are our
re-implementation with the same training and testing protocol as AnomalyCLIP and our AdaptCLIP.
Note that the results are averaged over all categories on each dataset and the full results of each category are presented in \texttt{Appendix}, the same below.}
\label{tab:sota_auroc}
\vspace{-10pt}
\resizebox{1.0\textwidth}{!}{
\begin{tabular}{cc|cc cc cc cc |c| cc|c}
\toprule
  \multirow{2}{*}{\textbf{Shots}}  &\multirow{2}{*}{\textbf{Methods}}    &\multicolumn{9}{c|}{\textbf{Industrial}}    & \multicolumn{3}{c}{\textbf{Medical}}     \\ 
    \cmidrule{3-11}\cmidrule{12-14}
  &
  &\footnotesize{MVTec}	
  &\footnotesize{VisA}  	
  &\footnotesize{BTAD} 	
  &\footnotesize{MVTec3D}
  &\footnotesize{DTD} 
  &\footnotesize{KSDD}
  &\footnotesize{MPDD} 
  &\footnotesize{Real-IAD}	
  &\footnotesize{AVG}
 &\footnotesize{Br35H} 	&\footnotesize{Covid} &\footnotesize{AVG}\\
\midrule
\multirow{4}{*}{\textbf{0}}		&WinCLIP~\cite{cvpr2023winclip}	&90.4	&75.5	&68.2	&69.4	&95.1	&92.9	&61.5	&67.0	&77.5	&80.5	&66.4	&73.5\\
	&AdaCLIP$^\dag$~\cite{eccv24adaclip} 	&90.7	&81.7	&89.9	&76.2	&92.7	&96.6	&64.0	&73.3	&83.1	&96.7	&69.4	&83.0\\
	&AnomalyCLIP~\cite{iclr2024anomalyclip} 	&91.6	&82.0	&88.3	&73.9	&93.9	&97.8	&77.5	&69.5	&\blue{84.3}	&94.2	&77.7	&\blue{86.0}\\
	&\textbf{AdaptCLIP}-Zero	&93.5	&84.8	&91.0	&78.6	&96.0	&98.1	&73.6	&74.2	&\red{86.2}	&94.8	&86.5	&\red{90.7}\\
\midrule
\multirow{4}{*}{\textbf{1}}		&WinCLIP+~\cite{cvpr2023winclip}	&\pmerror{93.6}{0.4}	&\pmerror{80.0}{2.4}	&\pmerror{84.4}{1.5}	&\pmerror{74.1}{0.4}	&\pmerror{97.9}{0.2}	&\pmerror{93.8}{0.4}	&\pmerror{69.3}{2.9}	&\pmerror{74.7}{0.2}	&83.4	&\pmerror{80.1}{2.1}	&\pmerror{90.1}{3.6}	&85.1\\
	&InCtrl~\cite{cvpr2024inctrl} 	&\pmerror{91.3}{0.4}	&\pmerror{83.2}{2.4}	&\pmerror{88.5}{0.4}	&\pmerror{75.3}{1.3}	&\pmerror{97.9}{0.3}	&\pmerror{ 92.0}{0.9}	&\pmerror{ 73.0}{2.7}	&\pmerror{76.6}{0.0}	&84.7	&\pmerror{83.9}{6.4}	&\pmerror{89.2}{5.3}	&86.6\\
	&AnomalyCLIP+~\cite{iclr2024anomalyclip} 	&\pmerror{95.2}{0.2}	&\pmerror{86.1}{0.7}	&\pmerror{88.5}{0.8}	&\pmerror{76.7}{2.1}	&\pmerror{98.0}{0.2}	&\pmerror{ 97.5}{0.3}	&\pmerror{ 83.4}{2.6}	&\pmerror{78.2}{0.0}	&\blue{88.0}	&\pmerror{90.8}{5.1}	&\pmerror{87.3}{2.6}	&\blue{89.1}\\
	&\textbf{AdaptCLIP}	&\pmerror{94.5}{0.5}	&\pmerror{90.5}{1.2}	&\pmerror{93.4}{0.0}	&\pmerror{81.7}{1.5}	&\pmerror{98.0}{0.0}	&\pmerror{96.9}{0.3}	&\pmerror{83.8}{2.2}	&\pmerror{81.8}{0.3}	&\red{90.1}	&\pmerror{93.7}{2.4}	&\pmerror{91.8}{2.5}	&\red{92.8}\\
\midrule
\multirow{4}{*}{\textbf{2}}		&WinCLIP+~\cite{cvpr2023winclip}	&\pmerror{94.5}{1.0}	&\pmerror{82.7}{1.0}	&\pmerror{85.8}{1.8}	&\pmerror{74.3}{0.3}	&\pmerror{98.1}{0.2}	&\pmerror{93.8}{0.2}	&\pmerror{69.3}{2.3}	&\pmerror{76.1}{0.1}	&84.3	&\pmerror{81.6}{0.6}	&\pmerror{91.8}{2.5}	&86.7\\
	&InCtrl~\cite{cvpr2024inctrl} 	&\pmerror{91.8}{0.9}	&\pmerror{86.3}{1.4}	&\pmerror{86.2}{2.0}	&\pmerror{ 75.4}{0.5}	&\pmerror{98.3}{0.2}	&\pmerror{91.6}{0.9}	&\pmerror{ 74.2}{1.8}	&\pmerror{78.5}{0.0}	&85.3	&\pmerror{ 86.1}{1.7}	&\pmerror{89.7}{5.1}	&87.9\\
	&AnomalyCLIP+~\cite{iclr2024anomalyclip} 	&\pmerror{95.4}{0.1}	&\pmerror{87.8}{0.5}	&\pmerror{89.2}{1.1}	&\pmerror{78.3}{1.3}	&\pmerror{98.2}{0.1}	&\pmerror{ 97.9}{0.2}	&\pmerror{ 83.4}{1.5}	&\pmerror{78.3}{0.0}	&\blue{88.6}	&\pmerror{91.5}{4.0}	&\pmerror{89.3}{2.7}	&\blue{90.4}\\
	&\textbf{AdaptCLIP}	&\pmerror{95.7}{0.6}	&\pmerror{92.2}{0.8}	&\pmerror{93.4}{0.2}	&\pmerror{82.9}{1.1}	&\pmerror{98.3}{0.0}	&\pmerror{97.2}{0.0}	&\pmerror{84.4}{0.7}	&\pmerror{82.9}{0.2}	&\red{90.8}	&\pmerror{94.0}{1.7}	&\pmerror{94.9}{0.9}	&\red{94.5}\\
\midrule
\multirow{4}{*}{\textbf{4}}		&WinCLIP+~\cite{cvpr2023winclip}	&\pmerror{95.3}{0.1}	&\pmerror{84.3}{0.6}	&\pmerror{87.8}{0.8}	&\pmerror{75.7}{0.3}	&\pmerror{98.2}{0.0}	&\pmerror{94.0}{0.2}	&\pmerror{71.2}{1.6}	&\pmerror{77.0}{0.0}	&85.4	&\pmerror{82.3}{0.4}	&\pmerror{92.9}{2.1}	&87.6\\
	&InCtrl~\cite{cvpr2024inctrl} 	&\pmerror{93.1}{0.7}	&\pmerror{87.8}{0.2}	&\pmerror{67.5}{2.4}	&\pmerror{78.1}{1.1}	&\pmerror{  97.7}{0.1}	&\pmerror{91.6}{0.9}	&\pmerror{78.6}{2.3}	&\pmerror{81.8}{0.0}	&84.5	&\pmerror{89.1}{1.2}	&\pmerror{91.4}{4.1}	&90.3\\
	&AnomalyCLIP+~\cite{iclr2024anomalyclip} 	&\pmerror{96.1}{0.1}	&\pmerror{88.8}{0.5}	&\pmerror{90.5}{1.2}	&\pmerror{79.2}{1.3}	&\pmerror{98.4}{0.1}	&\pmerror{ 97.8}{0.1}	&\pmerror{86.3}{1.8}	&\pmerror{78.4}{0.0}	&\blue{89.4}	&\pmerror{ 91.1}{4.4}	&\pmerror{91.4}{3.0}	&\blue{91.3}\\
	&\textbf{AdaptCLIP}	&\pmerror{96.6}{0.3}	&\pmerror{93.1}{0.2}	&\pmerror{93.3}{0.3}	&\pmerror{84.2}{0.6}	&\pmerror{98.5}{0.1}	&\pmerror{97.0}{0.2}	&\pmerror{86.8}{1.1}	&\pmerror{83.9}{0.2}	&\red{91.7}	&\pmerror{93.7}{2.0}	&\pmerror{95.8}{0.9}	&\red{94.8}\\
\bottomrule
\end{tabular}
}
\vspace{-5pt}
\end{table*}

\subsection{Comparisons with Zero-/Few-Shot Methods}

\begin{table*}[t]
\setlength\tabcolsep{3pt}
\centering
\small
\caption{\small\textbf{Pixel-level anomaly segmentation comparisons with AUPR metric on industrial and medical domains.}}
\label{tab:sota_aupr}
\vspace{-10pt}
\resizebox{1.0\textwidth}{!}{
\begin{tabular}{cc|cc cc cc cc |c| cc|c}
\toprule
  \multirow{2}{*}{\textbf{Shots}}  &\multirow{2}{*}{\textbf{Methods}}    &\multicolumn{9}{c|}{\textbf{Industrial}}    & \multicolumn{3}{c}{\textbf{Medical}}     \\ 
    \cmidrule{3-11}\cmidrule{12-14}
  &
  &\footnotesize{MVTec}	
  &\footnotesize{VisA}  	
  &\footnotesize{BTAD} 	
  &\footnotesize{MVTec3D}
  &\footnotesize{DTD} 
  &\footnotesize{KSDD}
  &\footnotesize{MPDD} 
  &\footnotesize{Real-IAD}	
  &\footnotesize{AVG}
 &\footnotesize{Kvasir} 	&\footnotesize{Endo} &\footnotesize{AVG}\\
\midrule
\multirow{4}{*}{\textbf{0}}		&WinCLIP~\cite{cvpr2023winclip}	&18.2	&~~5.4	&12.9	&~~5.3	&~~9.8	&~~7.1	&14.1	&~~3.3	&~~9.5	&27.8	&23.8	&25.8\\
	&AdaCLIP$^\dag$~\cite{eccv24adaclip} 	&39.1	&31.0	&42.9	&37.5	&75.2	&48.2	&25.9	&30.5	&\red{41.3}	&36.6	&43.7	&40.1\\
	&AnomalyCLIP~\cite{iclr2024anomalyclip} 	&34.5	&21.3	&45.5	&30.5	&62.6	&51.9	&28.9	&26.7	&37.7	&39.6	&46.6	&\blue{43.1}\\
	&\textbf{AdaptCLIP}-Zero	&38.3	&26.1	&41.8	&31.4	&68.7	&58.3	&25.3	&28.2	&\blue{39.7}	&45.3	&52.0	&\red{48.7}\\
\midrule
\multirow{4}{*}{\textbf{1}}		&WinCLIP+~\cite{cvpr2023winclip}	&\pmerror{38.3}{0.8}	&\pmerror{15.8}{0.2}	&\pmerror{41.3}{2.6}	&\pmerror{18.4}{1.1}	&\pmerror{47.8}{0.9}	&\pmerror{19.2}{0.3}	&\pmerror{29.8}{2.0}	&\pmerror{13.9}{0.2}	&28.1	&\pmerror{27.6}{2.9}	&\pmerror{23.6}{0.1}	&25.6\\
	&InCtrl~\cite{cvpr2024inctrl} 	&\pmerror{47.8}{1.1}	&\pmerror{17.7}{0.6}	&\pmerror{44.1}{1.4}	&\pmerror{18.7}{0.5}	&\pmerror{64.3}{0.5}	&\pmerror{26.7}{0.7}	&\pmerror{27.9}{2.2}	&\pmerror{19.1}{0.0}	&33.3	&\pmerror{22.1}{1.7}	&\pmerror{20.3}{3.7}	&21.2\\
	&AnomalyCLIP+~\cite{iclr2024anomalyclip} 	&\pmerror{40.8}{0.1}	&\pmerror{24.8}{0.9}	&\pmerror{41.3}{1.1}	&\pmerror{30.6}{1.1}	&\pmerror{67.4}{0.4}	&\pmerror{47.5}{0.5}	&\pmerror{34.2}{0.8}	&\pmerror{27.9}{0.0}	&\blue{39.3}	&\pmerror{ 46.9}{3.9}	&\pmerror{47.8}{4.9}	&\blue{47.4}\\
	&\textbf{AdaptCLIP}	&\pmerror{53.7}{0.9}	&\pmerror{38.9}{0.3}	&\pmerror{60.6}{1.0}	&\pmerror{40.7}{0.6}	&\pmerror{76.9}{0.1}	&\pmerror{57.8}{1.2}	&\pmerror{33.5}{2.5}	&\pmerror{36.6}{0.1}	&\red{49.8}	&\pmerror{49.2}{4.7}	&\pmerror{52.4}{4.7}	&\red{50.8}\\
\midrule
\multirow{4}{*}{\textbf{2}}		&WinCLIP+~\cite{cvpr2023winclip}	&\pmerror{39.5}{0.6}	&\pmerror{17.2}{0.8}	&\pmerror{42.8}{1.3}	&\pmerror{19.1}{0.8}	&\pmerror{48.2}{0.9}	&\pmerror{19.0}{0.5}	&\pmerror{30.7}{1.1}	&\pmerror{14.8}{0.1}	&28.9	&\pmerror{29.1}{0.2}	&\pmerror{27.6}{2.3}	&28.4\\
	&InCtrl~\cite{cvpr2024inctrl} 	&\pmerror{49.2}{0.7}	&\pmerror{18.5}{0.2}	&\pmerror{44.2}{0.8}	&\pmerror{20.3}{0.6}	&\pmerror{64.4}{0.4}	&\pmerror{26.4}{2.5}	&\pmerror{29.2}{1.3}	&\pmerror{20.1}{0.0}	&34.0	&\pmerror{24.9}{1.9}	&\pmerror{24.5}{7.5}	&24.7\\
	&AnomalyCLIP+~\cite{iclr2024anomalyclip} 	&\pmerror{41.5}{0.1}	&\pmerror{26.2}{0.7}	&\pmerror{41.9}{0.6}	&\pmerror{32.4}{1.5}	&\pmerror{68.1}{0.2}	&\pmerror{47.6}{0.4}	&\pmerror{ 35.3}{1.1}	&\pmerror{28.1}{0.0}	&\blue{40.1}	&\pmerror{47.3}{2.9}	&\pmerror{49.6}{4.8}	&\blue{48.5}\\
	&\textbf{AdaptCLIP}	&\pmerror{55.1}{0.5}	&\pmerror{40.7}{0.6}	&\pmerror{61.0}{0.6}	&\pmerror{42.3}{1.1}	&\pmerror{77.4}{0.2}	&\pmerror{57.5}{1.1}	&\pmerror{35.0}{0.7}	&\pmerror{37.8}{0.1}	&\red{50.9}	&\pmerror{49.0}{4.1}	&\pmerror{53.1}{4.2}	&\red{51.1}\\
\midrule
\multirow{4}{*}{\textbf{4}}		&WinCLIP+~\cite{cvpr2023winclip}	&\pmerror{41.2}{0.9}	&\pmerror{18.1}{1.3}	&\pmerror{44.0}{0.4}	&\pmerror{19.9}{0.6}	&\pmerror{49.3}{0.1}	&\pmerror{19.1}{0.7}	&\pmerror{32.0}{0.2}	&\pmerror{15.4}{0.2}	&29.9	&\pmerror{29.6}{0.8}	&\pmerror{27.7}{0.5}	&28.7\\
	&InCtrl~\cite{cvpr2024inctrl} 	&\pmerror{50.9}{0.3}	&\pmerror{19.2}{0.6}	&\pmerror{44.0}{0.2}	&\pmerror{22.2}{1.2}	&\pmerror{64.9}{0.3}	&\pmerror{26.0}{1.4}	&\pmerror{31.4}{0.8}	&\pmerror{21.0}{0.0}	&35.0	&\pmerror{24.7}{1.6}	&\pmerror{22.3}{1.0}	&23.5\\
	&AnomalyCLIP+~\cite{iclr2024anomalyclip} 	&\pmerror{42.4}{0.0}	&\pmerror{27.5}{1.1}	&\pmerror{45.8}{3.0}	&\pmerror{33.4}{1.3}	&\pmerror{68.5}{0.2}	&\pmerror{46.4}{0.7}	&\pmerror{36.8}{1.0}	&\pmerror{28.2}{0.0}	&\blue{41.1}	&\pmerror{45.9}{1.5}	&\pmerror{49.2}{3.4}	&\blue{47.6}\\
	&\textbf{AdaptCLIP}	&\pmerror{57.2}{0.8}	&\pmerror{41.8}{0.6}	&\pmerror{62.3}{0.3}	&\pmerror{44.5}{0.3}	&\pmerror{78.2}{0.2}	&\pmerror{56.4}{1.4}	&\pmerror{37.4}{1.1}	&\pmerror{39.1}{0.3}	&\red{52.1}	&\pmerror{47.5}{2.7}	&\pmerror{52.2}{3.1}	&\red{49.9}\\
\bottomrule
\end{tabular}
}
\vspace{-5pt}
\end{table*}

Tabs.~\ref{tab:sota_auroc} and~\ref{tab:sota_aupr} present comparisons of AdaptCLIP to competing zero-/few-shot methods in image-level anomaly classification and pixel-level anomaly segmentation, respectively, on 8 real-world industrial and 4 medical AD datasets. 
Note that we only use image-level metrics to evaluate Br35H and Covid due to the lack of pixel-level annotations, and only report the results for Kvasir and Endo using pixel-level metrics since normal images are not included in these two datasets. Below we analyze these results in detail.

\noindent\textbf{Generalization on Industrial Domain:}
Generally, AdaptCLIP significantly outperforms all competing models on almost all industrial datasets across three few-shot settings, 1-shot, 2-shot and 4-shot.
The performance of all methods generally gets better with more image prompts. Specifically, InCtrl~\cite{cvpr2024inctrl} surpasses WinCLIP~\cite{cvpr2023winclip} due to additional fine-tuning on a base training dataset. AnomalyCLIP~\cite{iclr2024anomalyclip} further achieves better generalization, which verifies the importance of learning object-agnostic prompts. AdaptCLIP exhibits superior performance, outperforming AnomalyCLIP~\cite{iclr2024anomalyclip} by a large margin (about 10\%+ in pixel AUPR and 2\%+ in image AUROC), particularly on challenging and large-scale datasets like VisA and Real-IAD. This reveals the power of comparative learning based on the joint contextual and aligned residual features for universal anomaly detection. Under zero-shot setting, AdaptCLIP-Zero significantly outperforms SOTA AdaCLIP~\cite{eccv24adaclip} on anomaly classification, although it shows a slight weakness in industrial anomaly segmentation. 
However, AdaptCLIP is simpler, requires fewer learnable parameters (0.6M vs. 10.7M in Tab.~\ref{tab:sota_efficiency}), and generalizes better from the industrial to the medical domain. In addition, our one-shot AdaptCLIP easily outperforms zero-shot AdaCLIP~\cite{eccv24adaclip} if only one normal image prompt is available.

\noindent\textbf{Generalization on Medical Domain:}
Our AdaptCLIP performs strongly on medical AD regardless of zero-shot or few-shot settings when applying the same model trained on an industrial dataset (i.e., MVTec). Surprisingly, it significantly outperforms SOTA AdaCLIP on image anomaly classification~(i.e., 6.3\% in AUROC) and pixel anomaly segmentation~(i.e., 8.6\% in AUPR). Notably, our approach still works even when replacing normal image prompts with anomaly images. This is meaningful for some special datasets that don't contain any normal images, such as Kvasir and Endo. Here, this success is mainly due to the proposed spatial alignment mechanism, as well as a strong prior assumption that anomaly pixels are mostly sparse.

\noindent\textbf{Efficiency Comparison:}
We measure complexity and efficiency by the number of parameters and the forward inference time, as shown in Tab.~\ref{tab:sota_efficiency}. The evaluation is performed on one V100 GPU with batch size 32. The number of parameters of AdaCLIP and AnomalyCLIP is 17 times and 9 times that of our AdaptCLIP, respectively. Compared to SOTA, AdaptCLIP achieves competitive inference time yet better AD performance. When extending from zero-shot to one-shot, AnomalyCLIP+ and our AdaptCLIP require almost no additional inference time, unlike earlier WinCLIP.

\begin{table}[t]
    \centering
    \setlength\tabcolsep{2.0pt}
    \caption{Complexity and efficiency comparisons.} \label{tab:sota_efficiency}
    \vspace{-10pt}
    \resizebox{0.48\textwidth}{!}{ 
        \begin{tabular}{@{}c cccrc@{}}
            \toprule
Shots	&Methods	&CLIP Models	&Input Size	&\# Params (M)	&Inf.Time (ms) \\
\midrule
\multirow{7}{*}{0}	
&\multirow{2}{*}{WinCLIP~\cite{cvpr2023winclip}}	&ViT-B-16+240	&240$\times$240 	&208.4 + ~0.0	&~~201.3\\
&&ViT-B-16+240	&512$\times$512	&208.4 + ~0.0	&3912.6  \\
&AdaCLIP~\cite{eccv24adaclip}	&ViT-L/14@336px	&518$\times$518	&428.8 + 10.7	&~~212.0 \\	
&AnomalyCLIP~\cite{iclr2024anomalyclip}	&ViT-L/14@336px	 &518$\times$518 	&427.9 + ~5.6	&~~154.9 \\
&\multirow{2}{*}{\textbf{AdaptCLIP-Zero}}	&ViT-B-16+240	&512$\times$512	&208.4 + ~0.4	&~~~~49.9\\ 
&&ViT-L/14@336px	&518$\times$518	&427.9 + ~0.6	&~~162.2 \\
\midrule
\multirow{7}{*}{1}
&\multirow{2}{*}{WinCLIP+~\cite{cvpr2023winclip}}	&ViT-B-16+240	&240$\times$240	&208.4 + ~0.0	&~~339.5 \\
&&ViT-B-16+240	&512$\times$512	&208.4 + ~0.0	&7434.9 \\
&InCtrl~\cite{cvpr2024inctrl}	&ViT-B-16+240	&240$\times$240 		&208.4 + ~0.3 &~~337.0 \\
&AnomalyCLIP+~\cite{iclr2024anomalyclip}	&ViT-L/14@336px	&518$\times$518	&427.9 + ~5.6	&~~158.6 \\
&\multirow{2}{*}{\textbf{AdaptCLIP}}	
&ViT-B-16+240	&512$\times$512	&208.4 + ~1.4	&~~~~54.0\\ 
&&ViT-L/14@336px	&518$\times$518	&427.9 + ~1.8	&~~168.2\\ 
             \bottomrule
        \end{tabular}
    } 
\vspace{-12pt}
\end{table}

\begin{figure*}[t]
    \centering
    \begin{minipage}{0.48\linewidth}
        \centering
        \includegraphics[width=\linewidth]{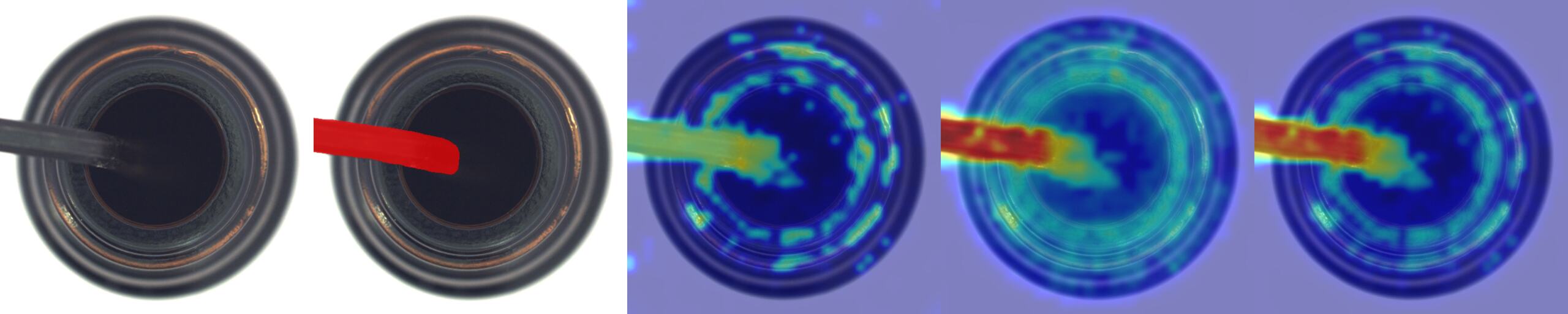}
        \includegraphics[width=\linewidth]{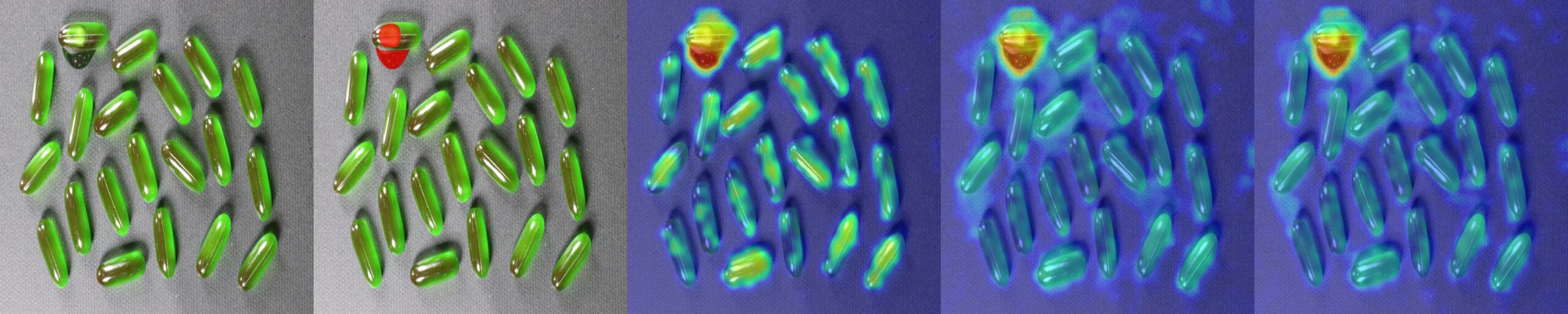}
        \includegraphics[width=\linewidth]{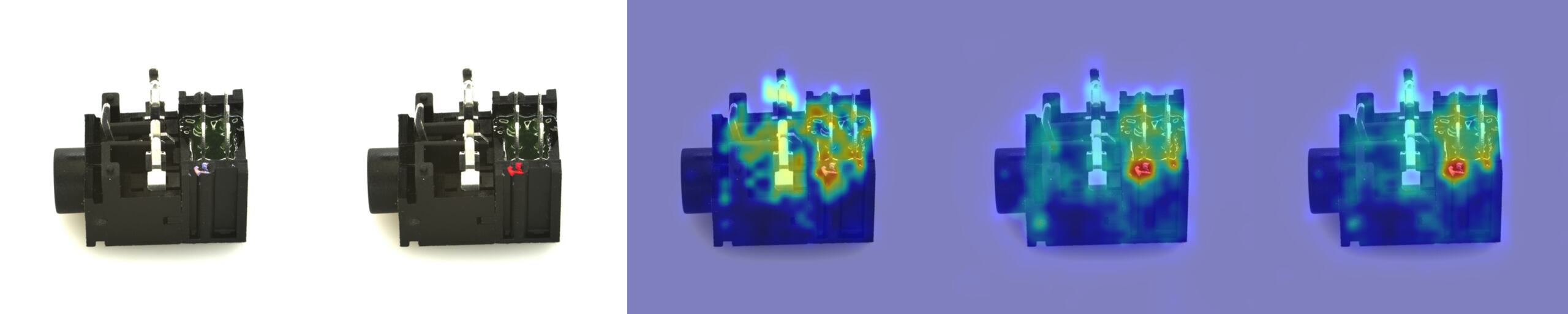}
        \includegraphics[width=\linewidth]{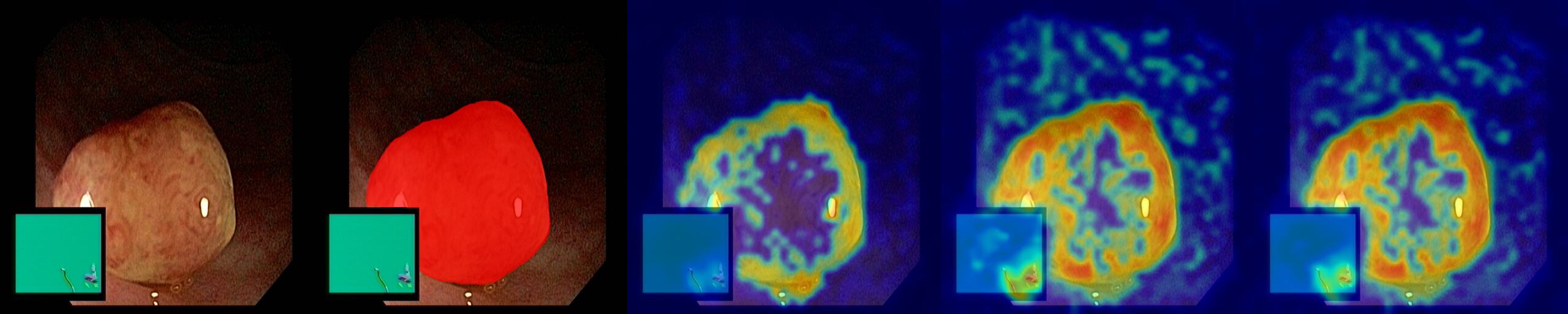}
        \vspace{-0.5cm} 
        
        \begin{tabular}{*{5}{>{\centering\arraybackslash}p{0.15\linewidth}}}
        \scriptsize{Query} &\scriptsize{GT Mask} &\scriptsize{0-shot}  &\scriptsize{1-shot} &\scriptsize{4-shot}  \\
        \end{tabular}
    \end{minipage}
    \hspace{0.0001\linewidth} 
    \begin{minipage}{0.48\linewidth}
        \centering
         \includegraphics[width=\linewidth]{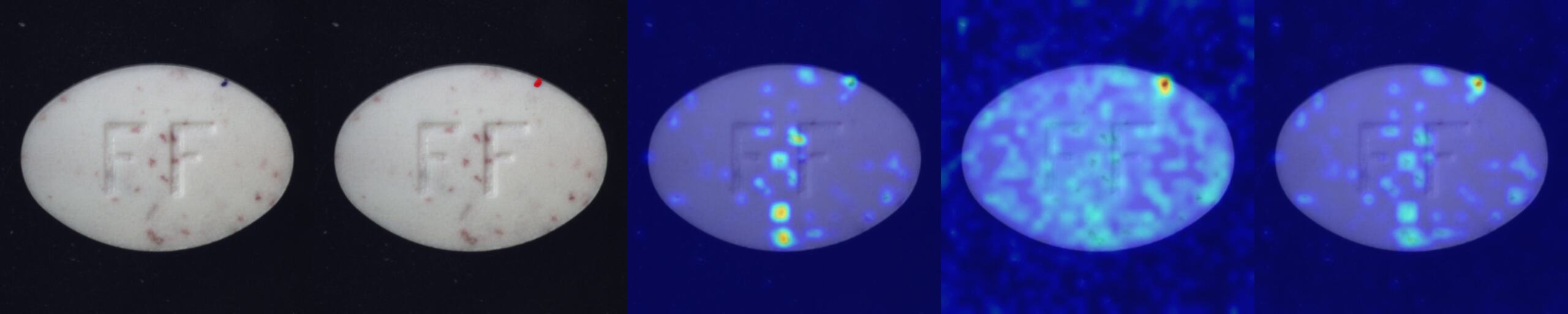}
         \includegraphics[width=\linewidth]{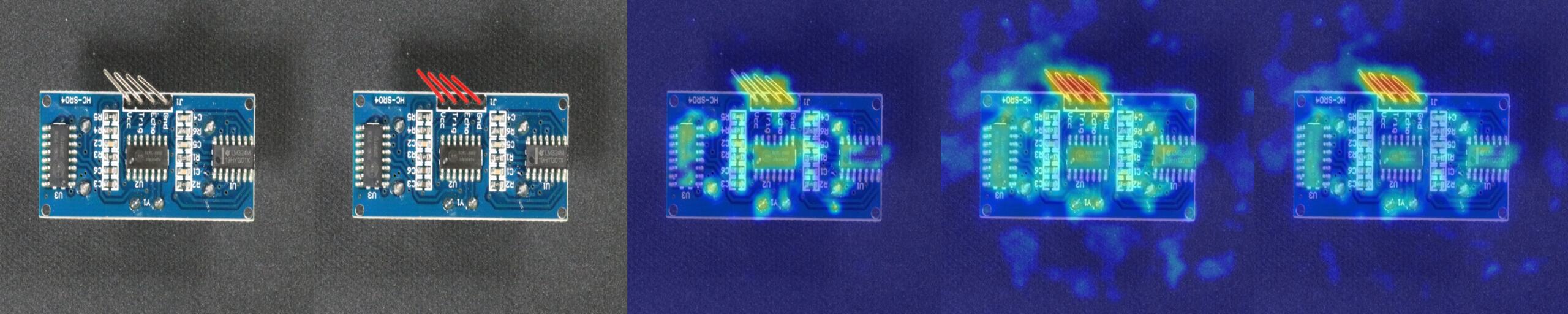}
        \includegraphics[width=\linewidth]{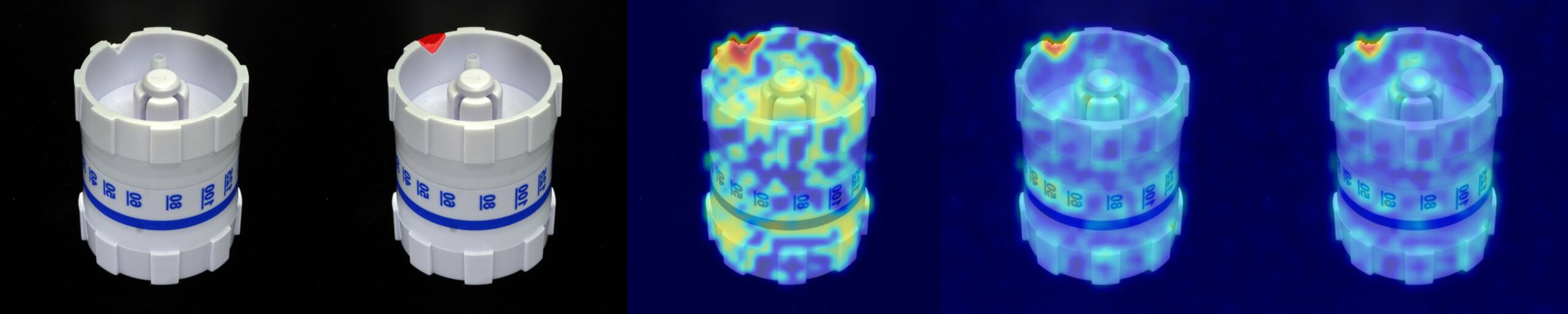}
        \includegraphics[width=\linewidth]{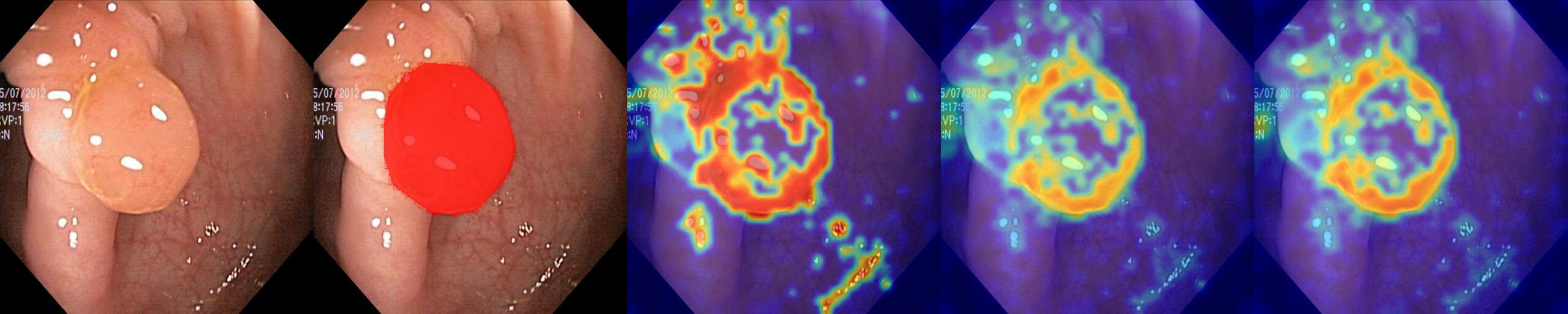}
        \vspace{-0.5cm} 
        
        \begin{tabular}{*{5}{>{\centering\arraybackslash}p{0.15\linewidth}}}
        \scriptsize{Query} &\scriptsize{GT Mask} &\scriptsize{0-shot}  &\scriptsize{1-shot} &\scriptsize{4-shot}  \\
        \end{tabular}
    \end{minipage}
 \vspace{-10pt}   
\caption{\small{Qualitative comparisons of our AdaptCLIP with different prompt numbers on MVTec, VisA, Real-IAD, Kvasir and Endo. More qualitative results of AdaptCLIP can be found in \texttt{Appendix}. Best viewed in color and zoom.}}\label{fig:comprompts}
\vspace{-10pt}
\end{figure*}

\noindent\textbf{Qualitative Results:} Fig.~\ref{fig:comprompts} shows some selected visualizations from industrial and medical testing images using AdaptCLIP. Generally, few-shot normal image prompts help AdaptCLIP segment
anomalies more accurately and produce fewer false positives than in a zero-shot manner.

\subsection{Comparisons with Many-Shot Methods}
In Tab.~\ref{tab:commanyshots}, we compare few-shot AdaptCLIP with many-shot and full-shot unified AD models.
It can be seen that AdaptCLIP is better than the early many-shot methods, RegAD~\cite{eccv2022regad}, and comparable to the latest PromptAD~\cite{cvpr2024promptad}. It is worth noting that PromptAD~\cite{cvpr2024promptad} requires re-training with few-shot normal images while our method remains training-free on target domains. Furthermore, our method outperforms full-shot methods, such as SimpleNet~\cite{cvpr2023simplenet} and UniAD~\cite{neurips2022uniad}, and is also competitive with the latest OneNIP~\cite{eccv2024onenip}. In short, our method has shown excellent performance, especially in the open-world scenario for universal anomaly detection, although there is still some small gap compared to state-of-the-art full-shot methods. 

\begin{table}[ht]
    \centering
    \setlength\tabcolsep{5.0pt}
    \caption{Comparisons of image-level anomaly classification and pixel-level anomaly segmentation~(using AUROC/AUPR metric, and the same as below) with many-shot and full-shot methods.}
    \vspace{-10pt}
    \resizebox{0.48\textwidth}{!}{
    \begin{tabular}{@{}lccccc@{}}
        \toprule
        Methods    &Shots  & MVTec         & VisA          & BTAD          \\ \midrule
        \multirow{2}{*}{\textbf{AdaptCLIP}} &1  & 94.5 / 53.7   & 90.5 / 38.9   & \red{93.4} / \blue{60.6}   \\
         &4  &\blue{96.6} / \blue{57.2}   & \red{93.1} / \blue{41.8}   & \blue{93.3} / \red{62.3}   \\
        \midrule
       RegAD~\cite{eccv2022regad} &8 & 91.2 / 51.1      & 79.7 / 28.6             & 90.7 / 40.5              \\
       PromptAD~\cite{cvpr2024promptad} &4  &\blue{96.6} /  52.9       & 89.1 / 31.5             & -             \\
    \midrule
    SimpleNet~\cite{cvpr2023simplenet} &full & 78.2 / 24.8 &89.2 / 33.1   & 90.3 / 36.2   \\
        UniAD~\cite{neurips2022uniad} &full &96.5 / 44.7  & 90.8 / 33.6   & 92.2 / 50.9   \\
    OneNIP~\cite{eccv2024onenip}	&full	&\red{97.9} / \red{63.7}	&\blue{92.5} / \red{43.3}	&92.6 / 56.8 \\
     \bottomrule
    \end{tabular}}
    \label{tab:commanyshots}
\end{table}

\subsection{Ablation Studies}
To demonstrate the effectiveness of the proposed three adapters in AdaptCLIP, TA: Texual Adapter, VA: Visual Adapter, and PQA: Prompt-Query Adapter, and two main insights, alternating learning, and comparative learning based on the joint contextual and aligned residual feature, we conduct experiments on MVTec and VisA, and report results in Tab.~\ref{tab:ab}. 

\noindent\textbf{Simple but effective baselines.} The first baseline is the naive CLIP~(Line \gray{0}), and it is simple and effective for zero-shot anomaly detection only using two-class textual prompts. However, it is still weak in pixel-level anomaly segmentation. The individual textual adapter and visual adapter are two addtional baselines. 
Specifically, the textual adapter can be seen as an extreme simplification of AnomalyCLIP~\cite{iclr2024anomalyclip}, removing the textual prompt template and textual prompt tuning. The simple textual adapter performs better than the original AnomalyCLIP and naive CLIP in anomaly classification, although it is slightly inferior in anomaly segmentation~(Lines \gray{0} vs. \gray{1}). The visual adapter learns adaptive local patch tokens and global image tokens to align textual representations from CLIP in both patch and image levels. This significantly improves pixel-level anomaly segmentation (Lines \gray{0} vs. \gray{2)}.

\begin{table}[t]
    \centering
    \setlength\tabcolsep{3pt}
    \caption{Ablation studies about different components. 
    }
    \vspace{-10pt}
    \resizebox{0.48\textwidth}{!}{ 
        \begin{tabular}{@{}c ccccc cccc@{}}
            \toprule
No. &Methods  &Shots    &TA  &VA &PQA &MVTec &VisA \\ 
\midrule
\gray{0} &\multirow{3}{*}{baselines}   
 &0    &\xmark &\xmark &\xmark &91.1 / 33.0 &82.1 / 18.0 \\
\gray{1} & &0    &\cmark &\xmark &\xmark &92.2 / 31.4 &82.9 / 19.7 \\
\gray{2} & &0    &\xmark &\cmark &\xmark &90.5 / 39.4 &81.0 / 22.1 \\
\midrule
\gray{3} &joint 
&0    &\cmark &\cmark  &\xmark &89.3 / 36.2 &81.6 / 21.5 \\
\gray{4} &\textcolor{azure}{alternating}  
&0    &\cmark &\cmark  &\xmark &93.5 / 38.3 &84.8 / 26.1 \\
\midrule
\gray{5} &w/o context   
&1    &\xmark &\xmark &\cmark  &62.6 / ~~7.0  &85.3 / 28.7 \\
\gray{6} &\textcolor{azure}{w context}    
&1    &\xmark &\xmark &\cmark  &88.1 / 50.2 &88.9 / 38.1 \\
\midrule
\gray{7} &\textcolor{azure}{AdaptCLIP}   
&1    &\cmark &\cmark &\cmark  &\red{94.2} / \red{52.5} & \red{92.0} / \red{38.8} \\ 
\bottomrule
        \end{tabular}
    } 
    \label{tab:ab}
\vspace{-10pt}
\end{table}

\noindent\textbf{Alternating learning is better than joint learning.}
We explore the impact of alternating learning and joint learning strategies on AdaptCLIP's performance. Alternating learning adapts visual or textual representations independently, while joint learning optimizes both representations simultaneously.
As shown (Lines \gray{3} vs. \gray{4}) in Tab.~\ref{tab:ab}, the alternating learning strategy significantly enhances the performance of AdaptCLIP compared to joint learning. Alternating learning not only fully leverages the strong prior guidance of CLIP's visual and textual representations but also mitigates the risk of over-fitting due to fine-tuning on a small training dataset. Additionally, we observe that the visual adapter alone excels in anomaly segmentation~(Line \gray{2}), whereas the textual adapter alone performs better in anomaly classification~(Line \gray{1}). By integrating the alternating learning into visual and textual adapters, AdaptCLIP generally achieves superior anomaly detection performance~(Line \gray{4}).

\noindent\textbf{The joint of contextual information and aligned residual features performs better than residual features alone.}
The aligned residual feature captures the distinctions between anomalous features and their corresponding normal counterparts. It effectively eliminates features related to individual objects and may improve generalization. However, we realize that isolated residual features may lose contextual information about visual objects, resulting in degraded model performance or even training failure (Line \gray{5}). Therefore, we propose a joint feature learning based on both contextual and aligned residual features, which further significantly boosts the model's performance~(Lines \gray{6} vs. \gray{5}). This means contextual information is equally important for anomaly identification.
Notably, the optimal performance for AdaptCLIP is achieved when all proposed components are integrated~(Line \gray{7}).

\noindent\textbf{Effects on pre-trained CLIP models.} We report zero- and one-shot results of AdaptCLIP using different CLIP models in Tab.~\ref{tab:ab-backbone}. It can be seen that a larger pre-trained model always brings better performance, especially in image-level classification. Furthermore, our method equipped with a lightweight model~(ViT-B-16+240) makes it possible to achieve competitive anomaly segmentation performance.

\begin{table}[ht]
    \centering
    \setlength\tabcolsep{5.0pt}
    \caption{Ablation studies about different pre-trained CLIP models.}
    \vspace{-10pt}
    \resizebox{0.48\textwidth}{!}{ 
        \begin{tabular}{@{}c ccccc cccc@{}}
            \toprule

 CLIP Models	&Input Size	&Shots	&MVTec			&VisA	\\		
\midrule
ViT-B-16+240	&512$\times$512	&0	&83.9 / 38.3	&75.4 / 19.5 \\
ViT-L/14@336px	&518$\times$518	&0	&93.5 / 38.3	&84.8 / 26.1 \\
\midrule
ViT-B-16+240	&512$\times$512	&1	&92.4 / 52.3	&85.2 / 30.3\\
ViT-L/14@336px	&518$\times$518	&1	&94.2 / 52.5	&92.0 / 38.8\\
             \bottomrule
        \end{tabular}
    } 
    \label{tab:ab-backbone}
\vspace{-16pt}
\end{table}

%% file: sec/5_conclusion.tex
\vspace{-8pt}
\section{Conclusion}
In this paper, we introduce a universal anomaly detection task, which focuses on generalizing anomaly detection models across domains, such as industrial and medical, and in open scenarios, such as zero- or few-shot settings. Once the universal anomaly detection model is trained, it does not need any fine-tuning on the target dataset. Compared with single zero-shot or few-shot AD models, the universal anomaly detection model is more flexible, supporting zero-/few-shot inference via fixed or learnable textual prompts and a few normal image prompts, while providing both image-level and pixel-level anomaly predictions. We propose a universal anomaly detection framework, AdaptCLIP, which alternately learns adaptive visual representations and text prompt embeddings, as well as jointly learns comparisons based on the contextual information of query image and the aligned residual features between the query and the prompt. Extensive experiments on 8 standard industrial and 4 medical datasets show that AdaptCLIP significantly outperforms current competitive models in multiple settings.

\noindent\textbf{Limitation:} 
AdaptCLIP achieves good AD performance only given zero-/few-shot normal image prompts. However, it could cause the model to confuse normal and abnormal instances and finally result in a decreased performance when we provide anomaly images as normal image prompts. Fortunately, normal images are generally relatively easy to obtain in practical applications. In addition, it may work using abnormal images as visual prompts because most of the pixels may be normal even in anomaly images.

%% file: sec/6_suppl.tex
\clearpage
\setcounter{page}{1}
\maketitlesupplementary

\section{Dataset Details}
To validate the effectiveness of our method, we conduct comprehensive experiments on 12 public anomaly detection datasets covering two domains, industrial and medical, and three modalities, including photography, radiology, and endoscopy. We only use two test datasets for model pre-training and generalization evaluation on other test datasets, and their relevant information is reported in Tab.~\ref{tab:datasets}. Specifically, we train models using the test data from MVTec and evaluate zero-/few-shot performance on other datasets. As for the evaluation of MVTec, we train models using VisA's test data. 

It should be noted that Real-IAD is the largest industrial anomaly detection dataset consisting of diverse categories (30 objects) and large-scale images (150k) among the utilized datasets. For the medical domain, we cannot find publicly available 2D medical AD datasets that include both image- and pixel-level annotations simultaneously. Therefore, we only report image-level classification performance on Br35H and Covid, while providing pixel-level anomaly segmentation performance on Kvasir and Endo. 

In addition, we note that MPDD and all four medical datasets are pose-agnostic, and KSDD may contain noise. We empirically find that the performance of few-shot AD methods may be limited and sometimes may be worse than zero-shot methods on these datasets. We believe this is a shortcoming of all few-shot normal image prompt-based methods.

\begin{table*}[t]
\setlength\tabcolsep{3pt}
\centering
\small
\caption{\small{Key statistics of industrial and medical datasets with different attributes. \cmark~means satisfied and \xmark~means not satified.}}\label{tab:datasets}
\vspace{-5pt}
\resizebox{1.0\textwidth}{!}{
\begin{tabular}{c|ccc cc cc cc cc}
\toprule
\multirow{2}{*}{\textbf{Domain}} 	&\multirow{2}{*}{\textbf{Dataset}} 	&\multirow{2}{*}{\textbf{Modality}}	&\multirow{2}{*}{\textbf{Category}}	&\multirow{2}{*}{\textbf{\# Classes}} &\multirow{2}{*}{\textbf{Pose-Agnostic}}	
&\multicolumn{2}{c}{\textbf{Anomaly Anotations}}		&\textbf{Train}	&\multicolumn{2}{c}{\textbf{Test}}	\\			
&&&&& &\textbf{Image-Level} &\textbf{Pixel-Level} &\textbf{\# Normal} &\textbf{\# Normal} &\textbf{\# Anomaly} \\
\toprule
\multirow{8}{*}{\textbf{Industrial}}
&MVTec 	&Photography	&Obj \& Texture	&15	&\xmark	&\cmark	&\cmark	&~~3,629	&~~~~467	&~~1,258\\
	&VisA 	&Photography	&Obj	&12	&\xmark	&\cmark	&\cmark	&~~8,659	&~~~~962	&~~1,200\\
	&BTAD 	&Photography	&Obj \& Texture	&3	&\xmark	&\cmark	&\cmark	&~~1,799	&~~~~451	&~~~~290\\
	&MVTec3D 	&Photography	&Obj	&10	&\xmark	&\cmark	&\cmark	&~~2,656	&~~~~249	&~~~~948\\
	&DTD 	&Photography	&Texture	&12	&\xmark	&\cmark	&\cmark	&~~1,200	&~~~~357	&~~~~947\\
	&KSDD 	&Photography	&Texture	&1	&\xmark	&\cmark	&\cmark	&~~~~857	&~~~~286	&~~~~~54\\
	&MPDD 	&Photography	&Obj	&6	&\cmark	&\cmark	&\cmark	&~~~~888	&~~~~176	&~~~~282\\
	&Real-IAD 	&Photography	&Obj	&30	&\xmark	&\cmark	&\cmark	&36,465	&63,256	&51,329\\
    \midrule

\multirow{4}{*}{\textbf{Medical}}
&Br35H 	&Radiology
(MRI)	&Brain	&1	&\cmark	&\cmark	&\xmark	&~~~~~~0	&~~~1500	&~~~1500\\
	&Covid	&Radiology (X-ray)	&Chest	&1	&\cmark	&\cmark	&\xmark	&~~~~~~0	&~~1,341	&~~~~219\\
	&Kvasir	&Endoscopy	&Gastrointestinal tract	&1	&\cmark	&\xmark	&\cmark	&~~~~~~0	&~~~~~~0	&~~1,000\\
	&Endo	&Endoscopy	&Gastrointestinal tract	&1	&\cmark	&\xmark	&\cmark	&~~~~~~0	&~~~~~~0	&~~~~200\\
\bottomrule
\end{tabular}
}
\vspace{-5pt}
\end{table*}

\begin{table*}[t]
\setlength\tabcolsep{1pt}
\centering
\small
\caption{\small{Comprehensive comparisons of state-of-the-art zero-/few-shot AD methods and our AdaptCLIP in terms of capabilities and complexity. \cmark~means satisfied and \xmark~means not satified.}}\label{tab:sotamethods}
\label{tab:capabilities}
\vspace{-5pt}
\resizebox{1.0\textwidth}{!}{
\begin{tabular}
{c cccccc | cccccc}
\toprule
\multirow{2}{*}{\textbf{Methods}} &\multicolumn{6}{c|}{\textbf{Capability}} &\multicolumn{6}{c}{\textbf{Complexity}} \\
&zero-shot	&few-shot	&image-cls. &pixel-seg. &unified	&ori-ability 	&pre-training	&post-finetuning	&sliding-wins	&class-names &learnable-prompts 	&image-prompts \\
\toprule
WinCLIP/WinCLIP+~\cite{cvpr2023winclip}	&\cmark	&\cmark	&\cmark	&\cmark	&\cmark	&\cmark	
&\xmark	&\xmark	&\cmark	&\cmark	&\xmark	&\cmark \\
AdaCLIP~\cite{eccv24adaclip}	&\cmark	&\xmark	&\cmark	&\cmark	&\cmark	&\xmark	
&\cmark	&\xmark	&\xmark	&\cmark	&\cmark	&\xmark \\
InCtrl~\cite{cvpr2024inctrl}	&\xmark	&\cmark	&\cmark	&\xmark	&\xmark	&\cmark	
&\cmark	&\xmark	&\xmark	&\cmark	&\xmark	&\cmark \\
AnomalyCLIP~\cite{iclr2024anomalyclip}	&\cmark	&\xmark	&\cmark	&\cmark	&\cmark	&\xmark	
&\cmark	&\xmark	&\xmark	&\xmark	&\cmark	&\xmark \\
PromptAD~\cite{cvpr2024promptad}	&\xmark	&\cmark	&\cmark	&\cmark	&\xmark	&\cmark	
&\cmark	&\cmark	&\xmark	&\cmark	&\cmark	&\cmark \\
\rowcolor[HTML]{EFEFEF}	\textbf{AdaptCLIP}		&\cmark	&\cmark	&\cmark	&\cmark	&\cmark	&\cmark	
&\cmark	&\xmark	&\xmark	&\xmark	&\cmark	&\cmark \\
\bottomrule
\end{tabular}
}
\vspace{-5pt}
\end{table*}

\section{Implementation Details}
We utilize the pre-trained CLIP (ViT-L/14@336) as the default CLIP model and extract local patch tokens from layers $\{6, 12, 18, 24\}$ and global image token from the last layer~$\{24\}$. All images are resized to a resolution of 518$\times$518 for training and testing.
Regarding the visual and textual adapters, we only use features from the last layer (i.e., 24) of the CLIP visual encoder, while for the prompt-query adapter, we use features from all 4 layers.
The visual adapter is a two-layer MLP, whose hidden layer dimension is 1/4 of the input layer, and the output layer dimension remains the same as the input layer. The length $r$ of learnable textual prompt embeddings is set to 12 in the textual adapter.  For the prompt-query adapter, the dimension of the first hidden layer is set to 128, and then the dimension of the next layer is halved until the last layer is set to 2 in both the lightweight segmentation head and the global MLP. We train models for 15 epochs with a learning rate of 0.001. All experiments are conducted using PyTorch with a single NVIDIA V100 GPU.

\subsection{Competing Methods}
For fair comparison, we compare state-of-the-art zero-shot methods, such as WinCLIP~\cite{cvpr2023winclip}, AnomalyCLIP~\cite{iclr2024anomalyclip}, and AdaCLIP~\cite{eccv24adaclip}, and few-shot methods, such as WinCLIP+~\cite{cvpr2023winclip}, InCtrl~\cite{cvpr2024inctrl}, AnomalyCLIP+, and PromptAD~\cite{cvpr2024promptad}, with our AdaptCLIP using the same training protocol and few-shot normal image prompts. It is worth noting that the original InCtrl~\cite{cvpr2024inctrl} only supports image-level few-shot AD, and we have appropriately extended it to allow pixel-level few-shot AD. In addition, AnomalyCLIP+ is an extension of AnomalyCLIP introducing feature association in WinCLIP+. In Tab.~\ref{tab:sotamethods}, we qualitatively analyze these methods in terms of \textbf{capability}, including zero-shot, few-shot, image-level anomaly classification, pixel-level anomaly segmentation, unified or separated models and original CLIP ability, and \textbf{complexity}, including pre-training, post-finetuning on target datasets, sliding windows, class names, learnable-prompts and few-shot normal image prompts. We summarize them in detail as follows.

\textbf{WinCLIP}~\cite{cvpr2023winclip} is the first zero-shot anomaly detection method based on a vision-language model, i.e., CLIP. WinCLIP designs two-class textual prompts and introduces multi-scale patch windows for accurate anomaly segmentation. 
However, it brings large computational costs and memory burden, limiting high-resolution input or large pre-trained models.  
Note
that no official implementation of WinCLIP is available, our results are based on an \href{https://github.com/zqhang/Accurate-WinCLIP-pytorch}{unofficial implementation}.

\textbf{WinCLIP+}~\cite{cvpr2023winclip} combines language- and visual-guided predictions for better anomaly classification and segmentation. 
The language-guided prediction is the same
as in WinCLIP. For visual-guided prediction, it first simply stores multi-scale features from few-shot normal images into a memory bank, and then measures the anomaly score using the distance or similarity between each query feature and the nearest feature from the memory bank. The final anomaly score is derived by averaging these two scores.

\textbf{AnomalyCLIP}~\cite{iclr2024anomalyclip} learns object-agnostic text prompts that capture generic normality and abnormality
in an image regardless of its foreground objects. However, AnomalyCLIP requires fine-tuning on
an auxiliary domain dataset including normal and anomaly images. AnomalyCLIP is a zero-shot anomaly detection method and it is capable of recognizing any anomalies. We
use the official model to report performance for anomaly classification and segmentation.

\textbf{AnomalyCLIP+} is one of our baselines. Note that the original AnomlayCLIP only supports zero-shot anomaly detection. To achieve few-shot anomaly detection, we introduce feature association based on a memory mechanism, similar to WinCLIP+, to AnomalyCLIP. The final anomaly results are also the average of zero-shot predictions guided by learnable textual prompts and few-shot predictions guided by few-shot normal image prompts.

\textbf{InCtrl}~\cite{cvpr2024inctrl} integrates multi-level information, including patch-level residual maps, image-level residual features, and prior knowledge score using two-class textual prompts, to learn a holistic scoring function for anomaly classification. However, it does not consider pixel-level anomaly segmentation.
In this paper, we simply use the patch-level residual map as the pixel-level anomaly prediction, which is essentially similar to WinCLIP+. It is worth noting that InCtrl provides multiple models for different shot numbers. We use its official models for extensive evaluation. In addition, since it does not provide a 1-shot model, we use the 2-shot model to evaluate 1-shot performance.

\textbf{AdaCLIP}~\cite{eccv24adaclip} further integrates visual knowledge from query images into textual prompt embeddings for enhancing the interaction of visual features and textual prompt embeddings. Different from AnomalyCLIP, AdaCLIP trains base models using more auxiliary datasets, including industrial and medical, which is not conducive to cross-domain evaluation. For a fair comparison, we retrain AdaCLIP models using the same training protocol as AnomalyCLIP, and conduct comprehensive evaluations on multiple datasets for zero-shot anomaly detection.

\textbf{PromptAD}~\cite{cvpr2024promptad} introduces an explicit anomaly margin to mitigate the training challenge caused by the absence of anomaly training images. Instead of using a unified paradigm (i.e., one model for all classes) in AnomalyCLIP and AdaCLIP, PromptAD uses a separate paradigm (i.e. one model for one class). Therefore, it needs to re-train a model with few-shot normal images when applied to each class of the target datasets. In addition, image-level anomaly classification and pixel-level segmentation models also need to be trained separately. In this paper, we only compare PromptAD with our method on MVTec and VisA because it involves fine-tuning for each class of target datasets.

\section{Compelete Experimental Results}

\begin{figure*}[tb]
  \centering
\begin{subfigure}{0.32\linewidth}
    \rotatebox{90}{\qquad\qquad\quad\textbf{Image-Level Classification}}\includegraphics[width=1.00\columnwidth,keepaspectratio]{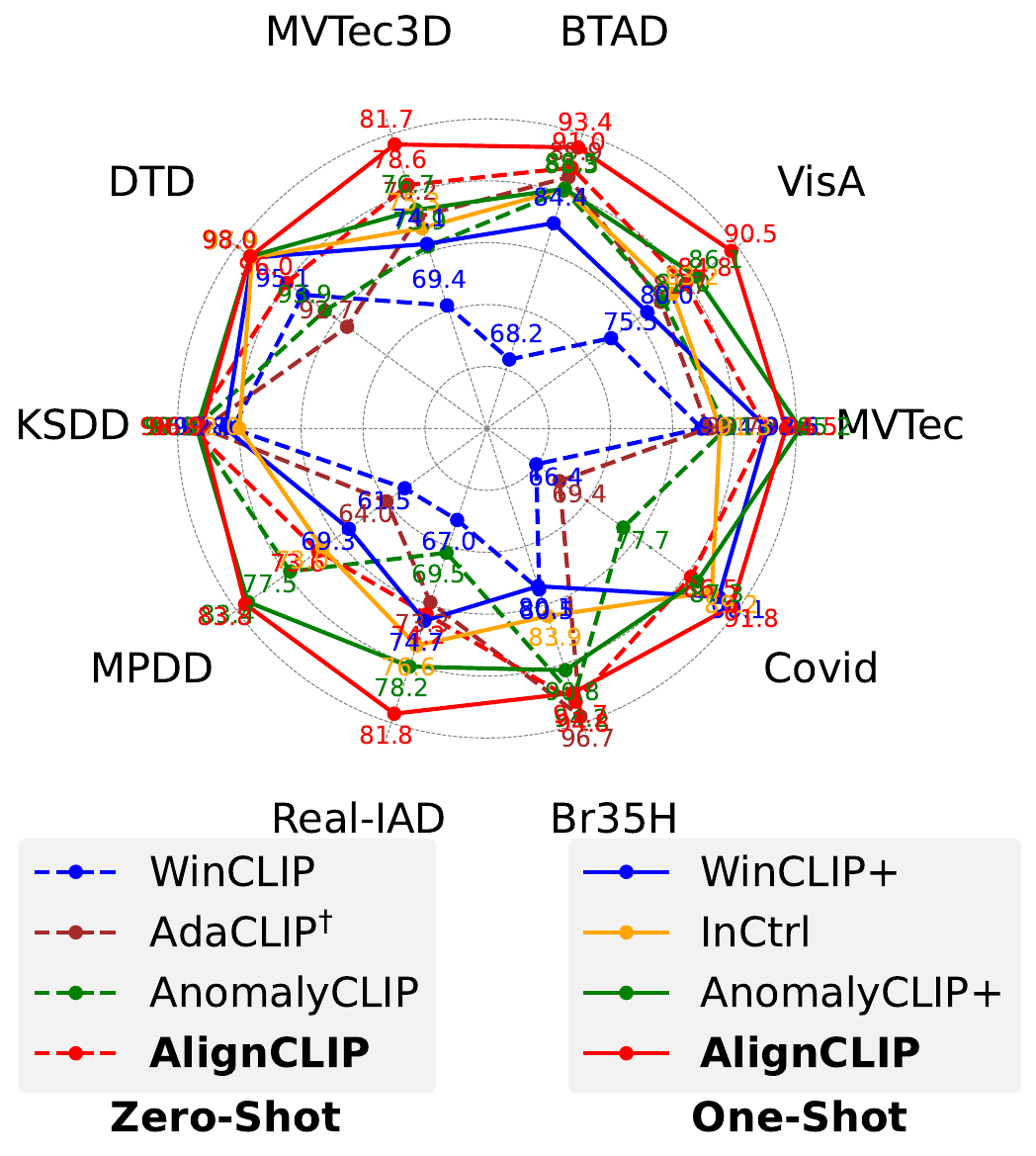} 
    \rotatebox{90}{\qquad\qquad\quad\textbf{Pixel-Level Segmentation}}\includegraphics[width=1.00\columnwidth,keepaspectratio]{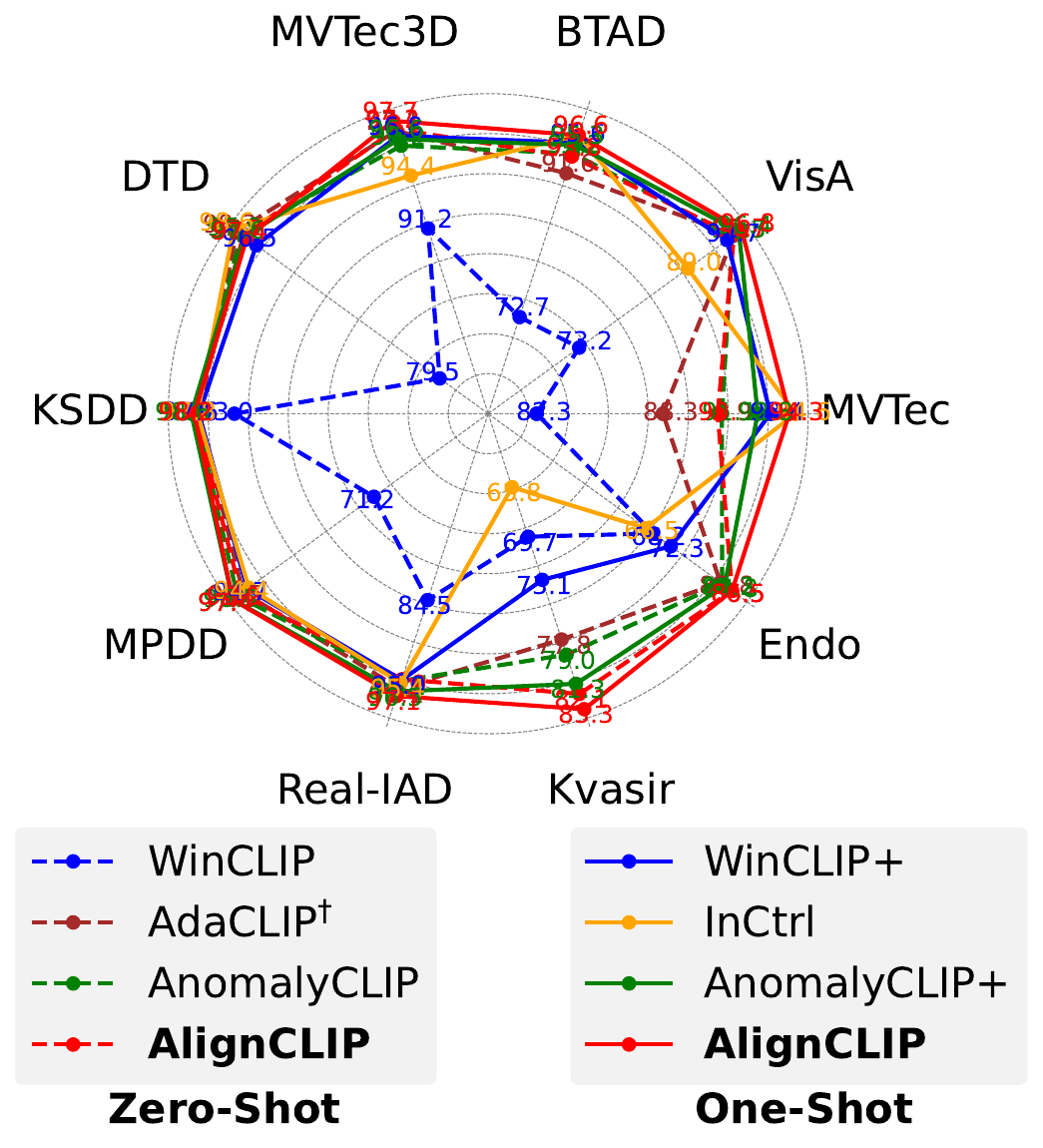} 
    \caption{AUROC.}
\end{subfigure}
\hspace{5pt}
\begin{subfigure}{0.32\linewidth}
    \includegraphics[width= 1.00\columnwidth]{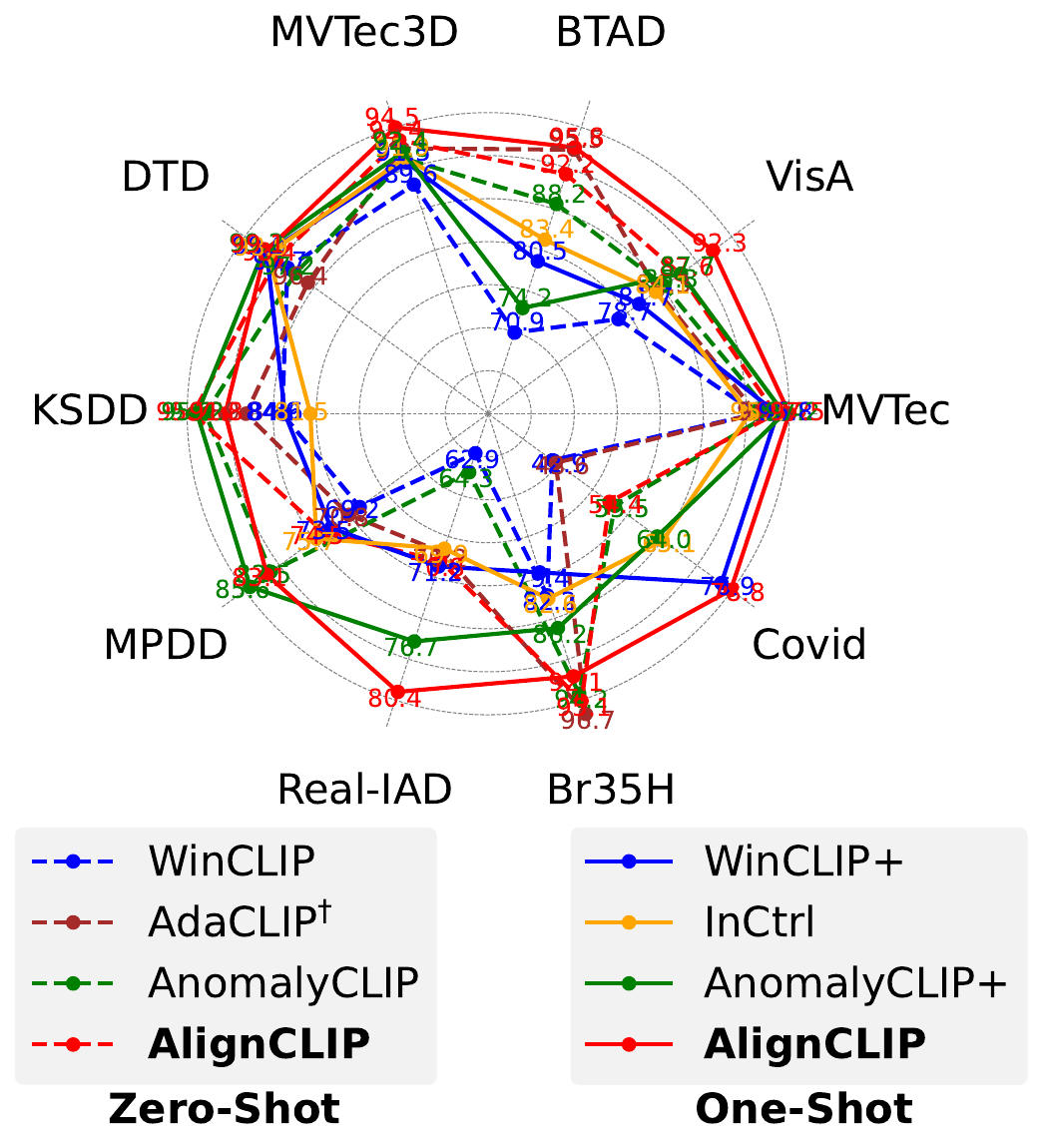} 
    \includegraphics[width= 1.00\columnwidth]{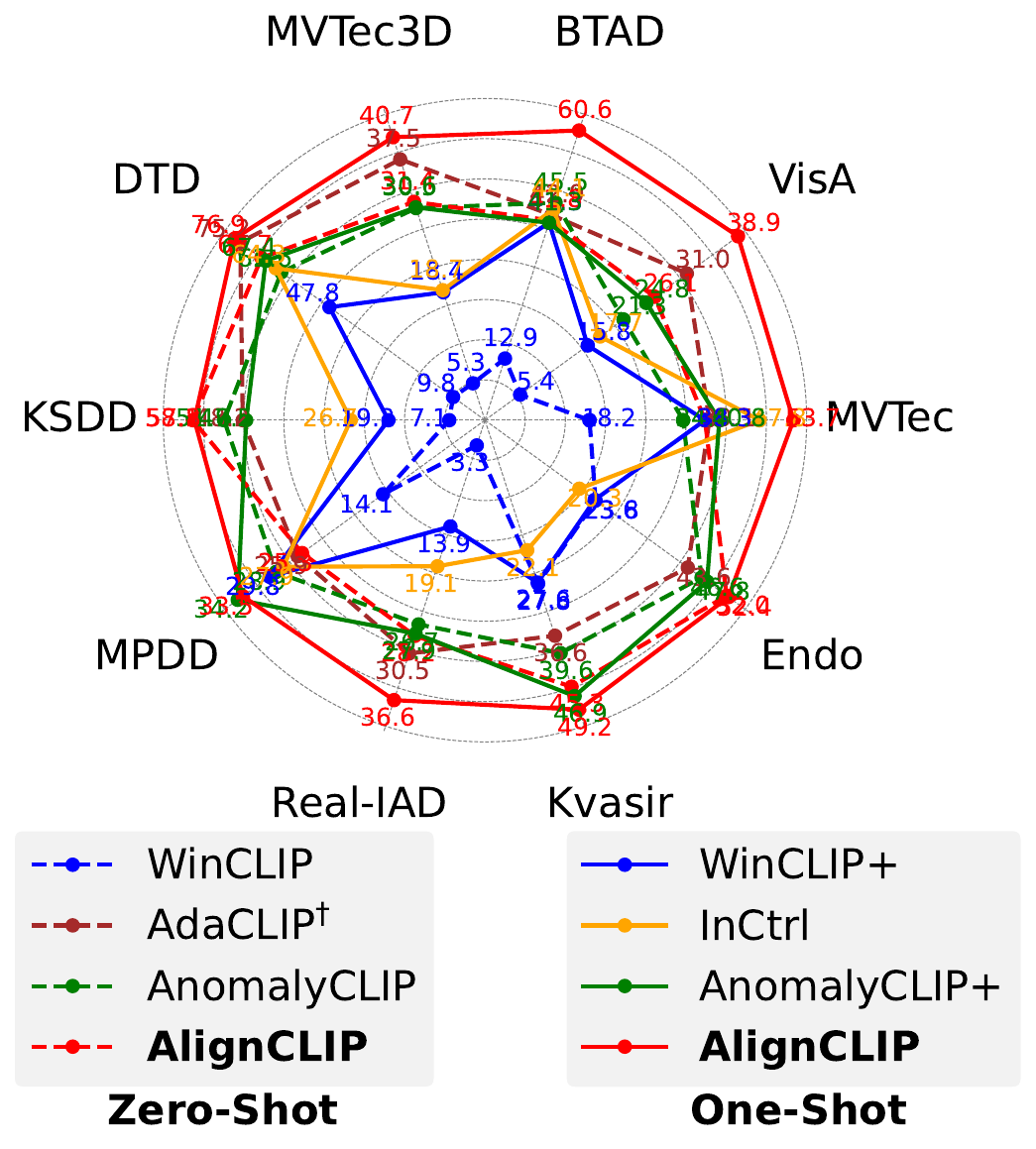} 
    \caption{AUPR.}
   \end{subfigure}
\begin{subfigure}{0.32\linewidth}
    \includegraphics[width= 1.00\columnwidth]{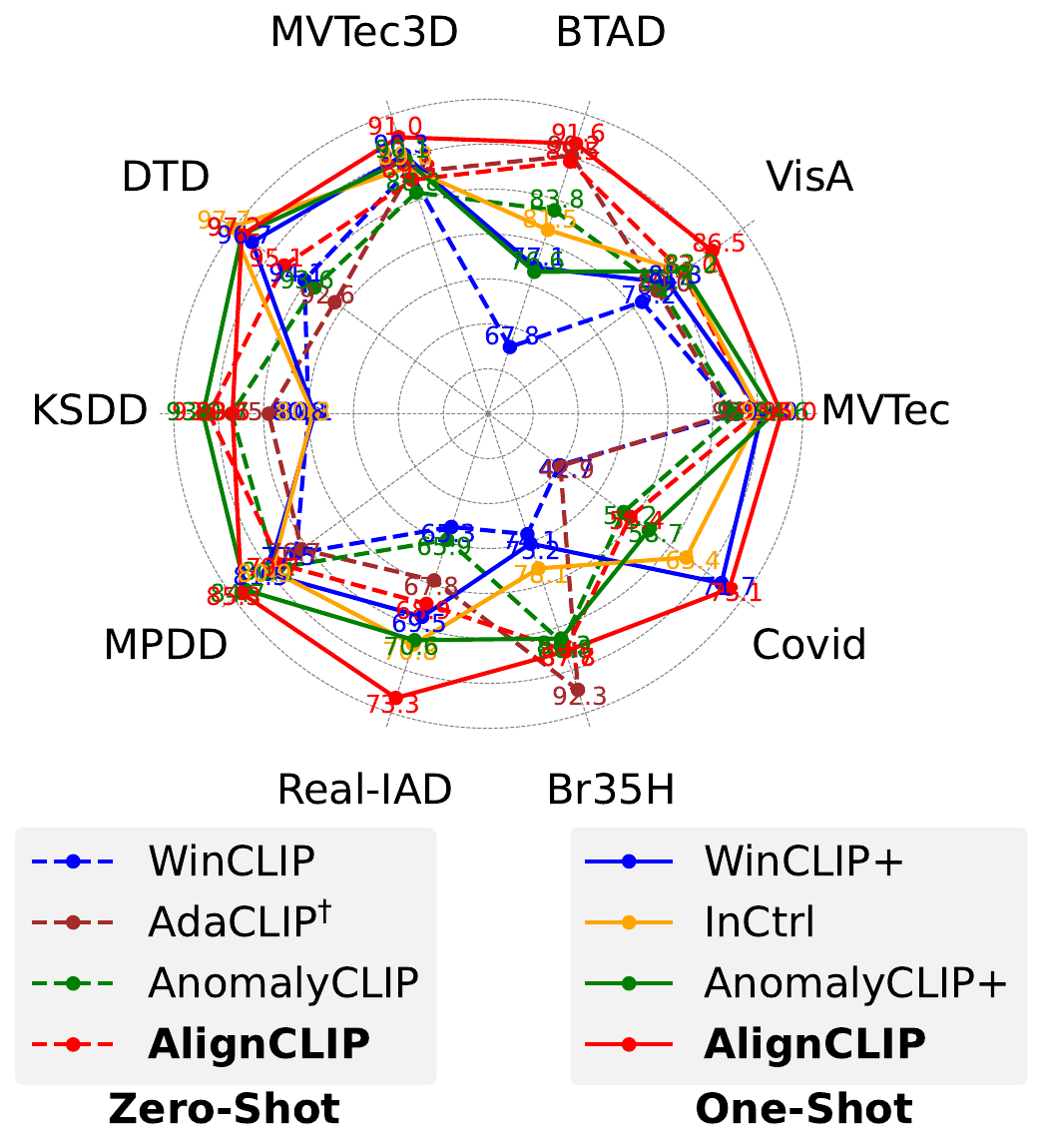} 
    \includegraphics[width= 1.00\columnwidth]{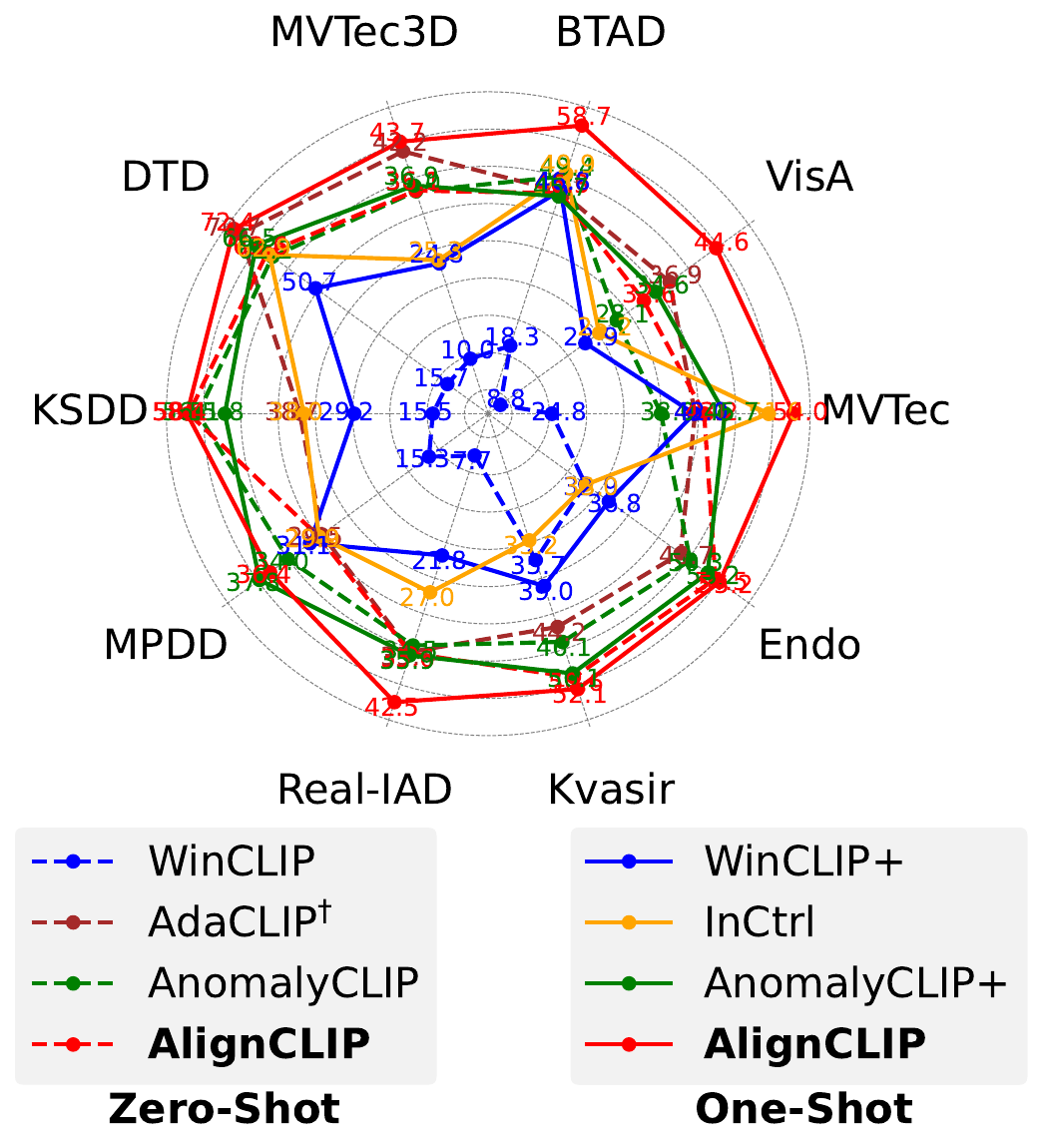} 
    \caption{F1$_\text{max}$.}
\end{subfigure}
\caption{Image-level anomaly classification and pixel-level anomaly segmentation comparisons of state-of-the-art zero-/one-shot methods and our \textbf{AdaptCLIP} with all three metrics, AUROC, AUPR and F1$_\text{max}$. The one-shot AdaptCLIP utilizes a training-free manner on target domains and achieves more accurate anomaly classification and segmentation on 8 industrial and 4 medical benchmarks.}\label{fig:comzerofewshot}
\vspace{-5pt}
\end{figure*}

In our main paper, we compare state-of-the-art methods with our AdaptCLIP using AUROC for image-level anomaly classification and AUPR for pixel-level anomaly segmentation. Here, we provide more comprehensive comparisons, including image-level anomaly classification in AUPR and F1$_\text{max}$ in Tabs.~\ref{tab:sota_iaupr} and~\ref{tab:sota_if1max}, and pixel-level anomaly segmentation in AUROC and F1$_\text{max}$ in Tabs.~\ref{tab:sota_pauroc} and~\ref{tab:sota_pf1max}, respectively. To more intuitively show the performance trends between zero-shot and few-shot methods on different datasets, we show comprehensive comparisons using all three metrics (AUROC, AUPR and F1$_\text{max}$) for image-level anomaly classification and pixel-level anomaly segmentation, as shown in Fig.~\ref{fig:comzerofewshot}.
In addition, we only report the averaged results of all categories for each dataset in our main paper. Here, we also provide more detailed reports in Tabs.~\ref{tab:mvtec},~\ref{tab:visa},~\ref{tab:mvtec3d},~\ref{tab:dtd},~\ref{tab:mpdd},~\ref{tab:realiad01},~\ref{tab:realiad24},~\ref{tab:btad},~\ref{tab:ksdd}, and~\ref{tab:medical} for each category on MVTec, VisA, MVTec3d, DTD, MPDD, Real-IAD, BTAD, KSDD, Br35H, Covid, Kvasir and Endo, respectively.

\begin{figure*}[t]
    \centering
    \begin{minipage}{0.48\linewidth}
        \centering
        \includegraphics[width=\linewidth]{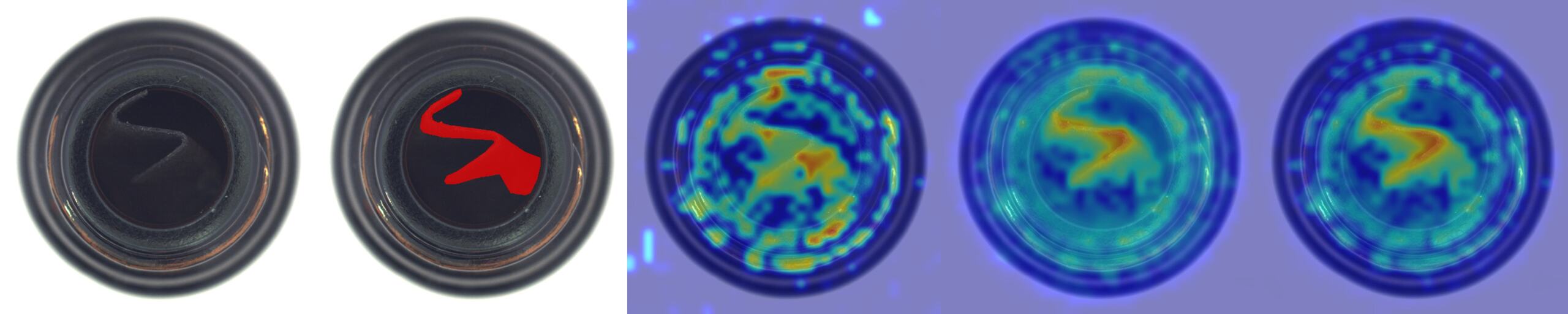}
        \includegraphics[width=\linewidth]{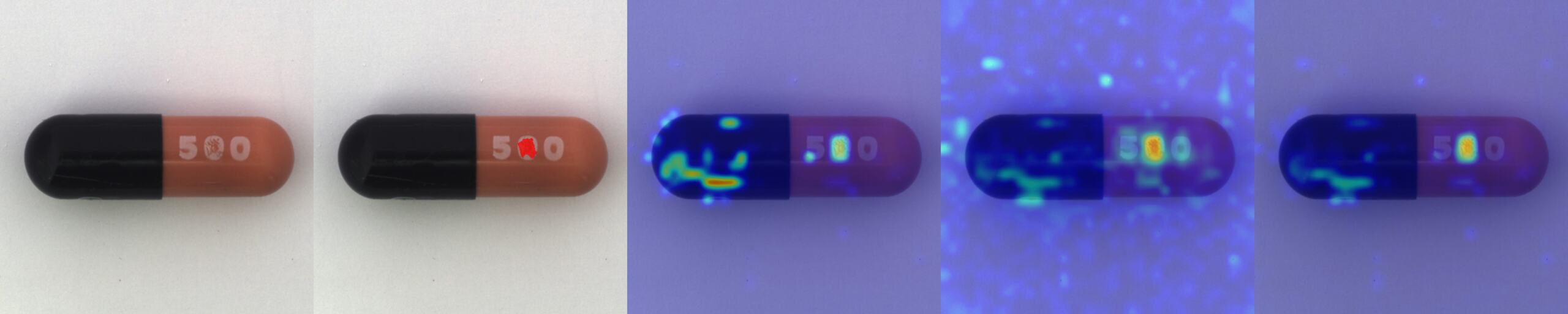}
        \includegraphics[width=\linewidth]{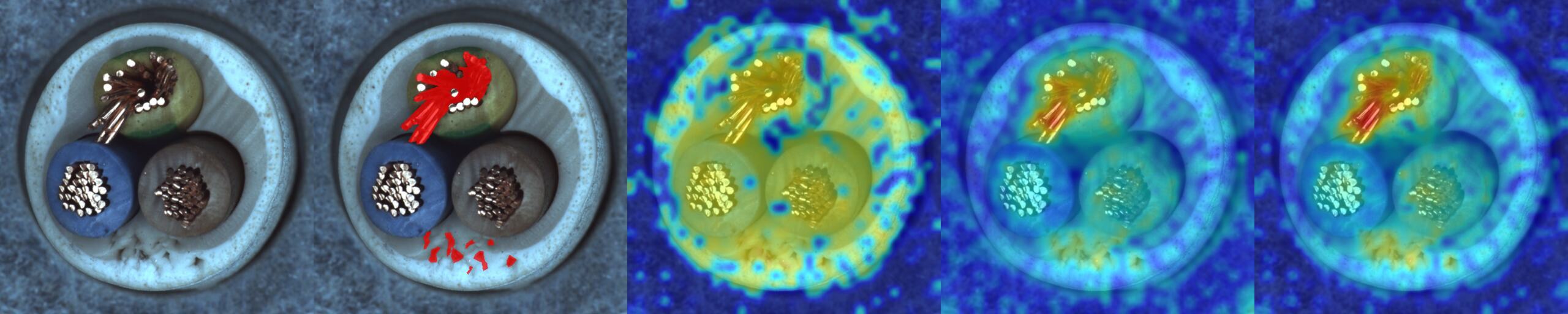}
        \includegraphics[width=\linewidth]{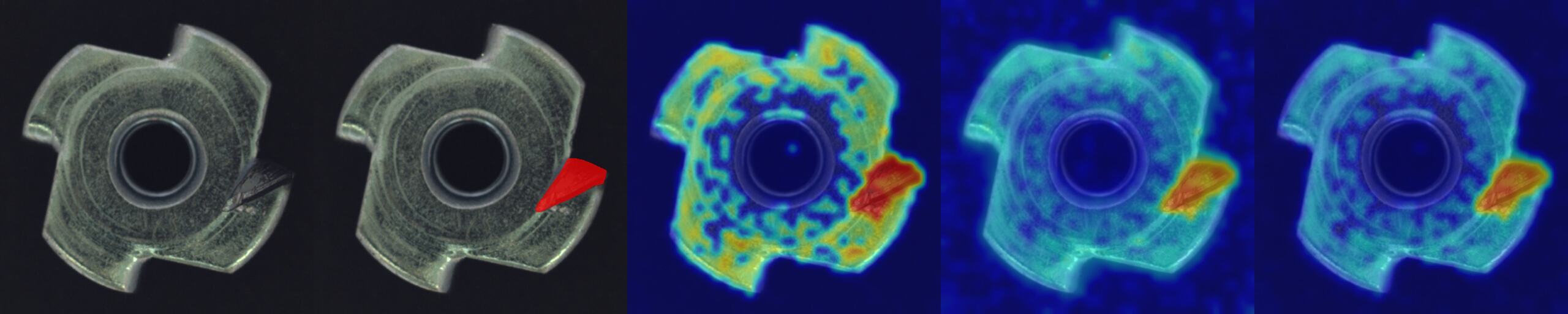}
        \includegraphics[width=\linewidth]{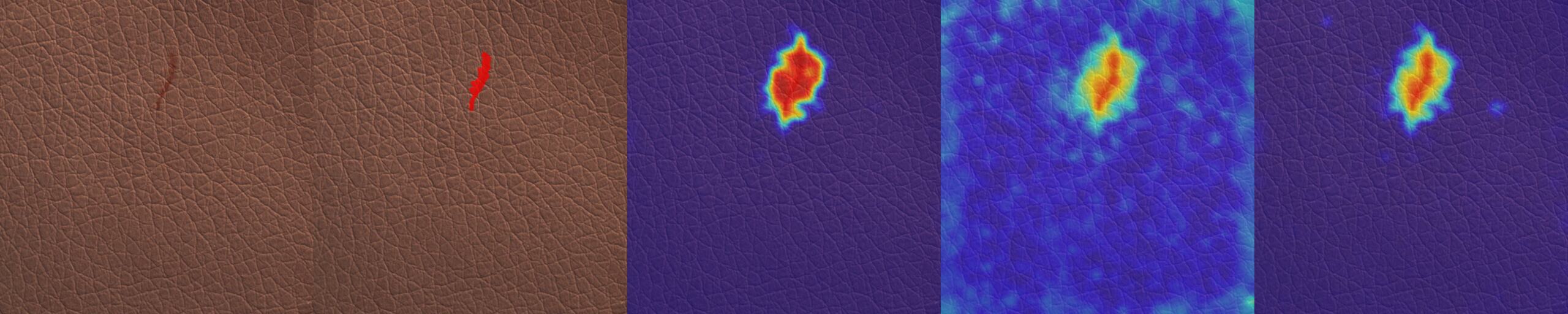}
        \includegraphics[width=\linewidth]{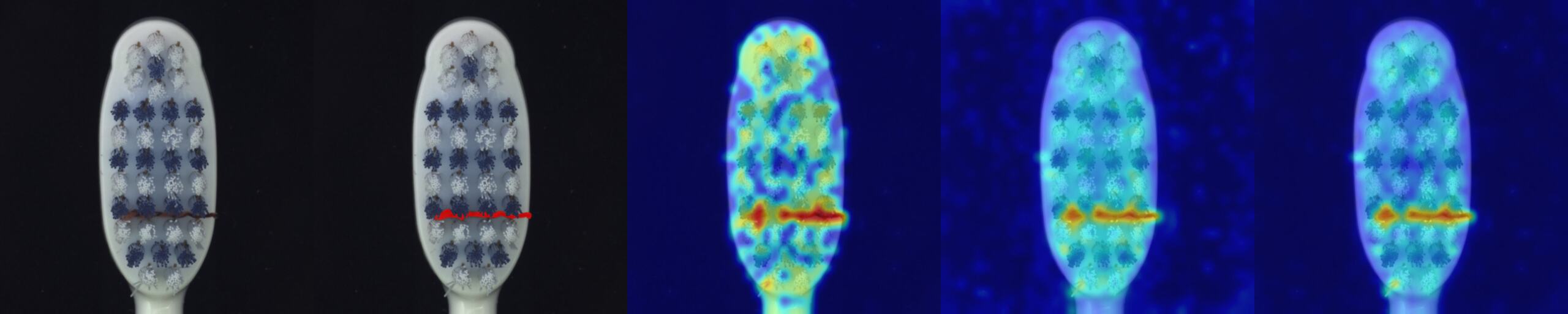}
        \includegraphics[width=\linewidth]{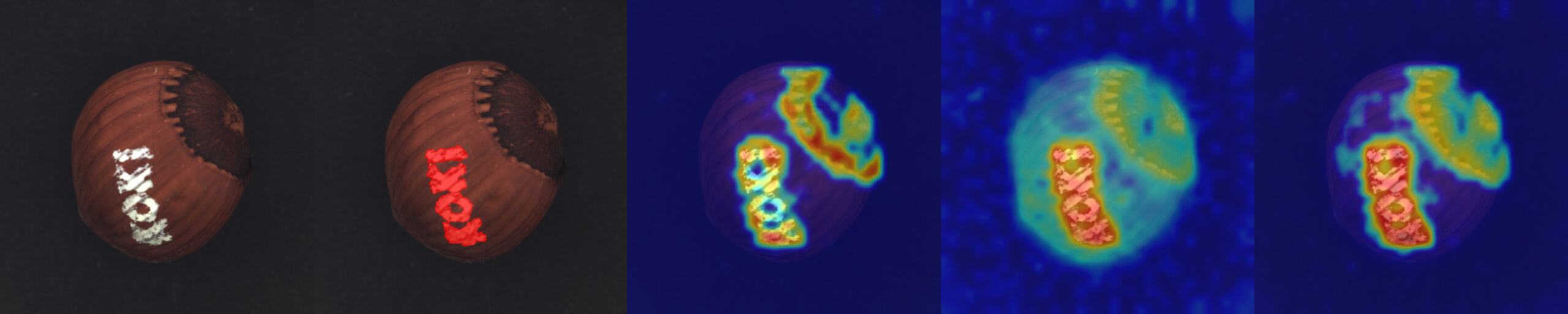}
        \includegraphics[width=\linewidth]{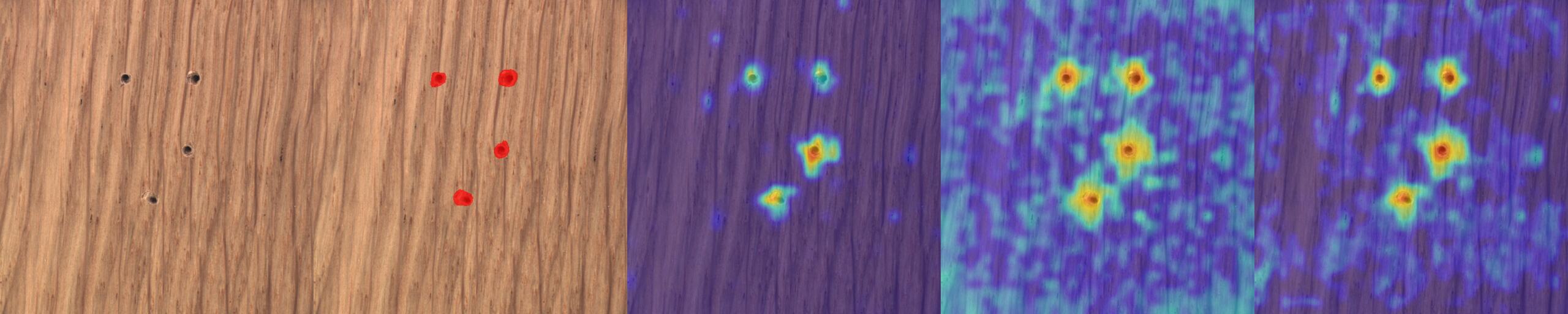 }
        \vspace{-0.5cm} 
        
        \begin{tabular}{*{5}{>{\centering\arraybackslash}p{0.15\linewidth}}}
        \scriptsize{Query} &\scriptsize{GT Mask} &\scriptsize{0-shot}  &\scriptsize{1-shot} &\scriptsize{4-shot}  \\
        \end{tabular}
    \end{minipage}
    \hspace{0.0001\linewidth} 
    \begin{minipage}{0.48\linewidth}
        \centering
        \includegraphics[width=\linewidth]{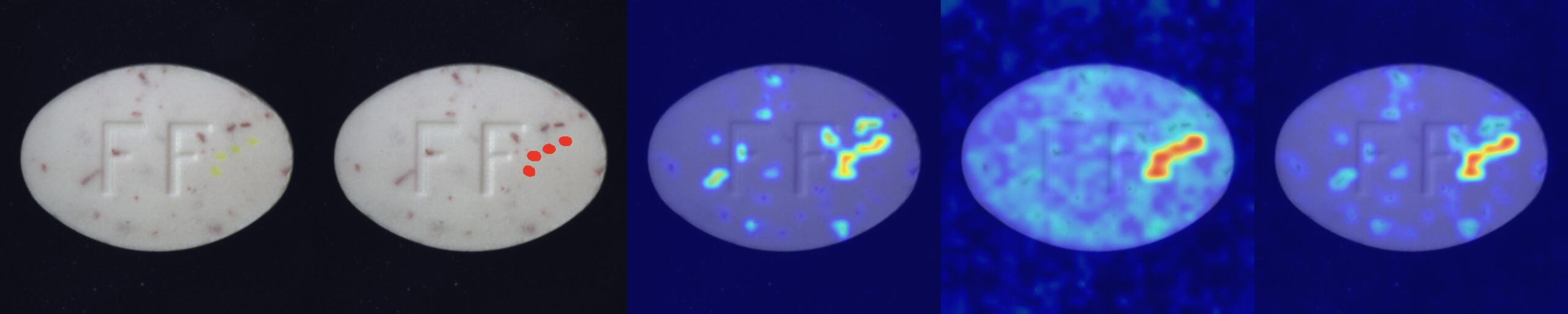}
        \includegraphics[width=\linewidth]{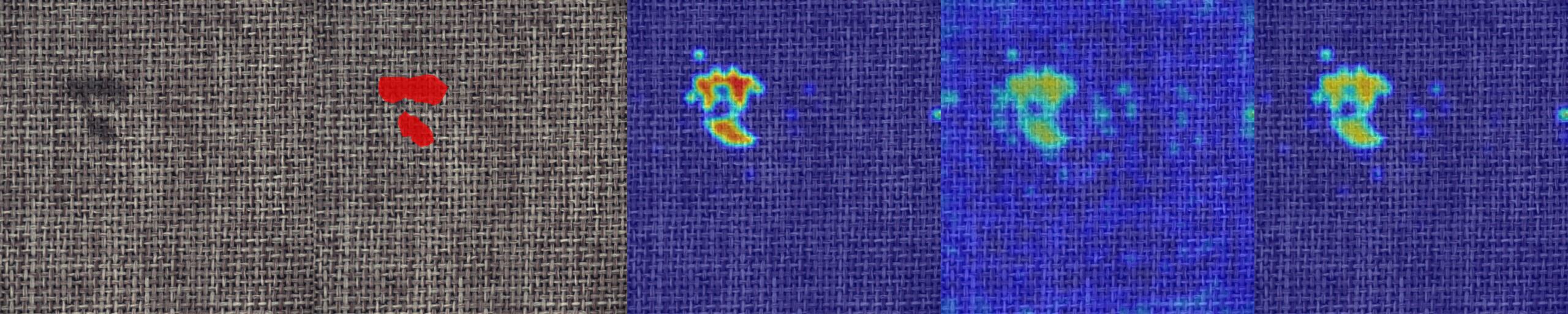}
        \includegraphics[width=\linewidth]{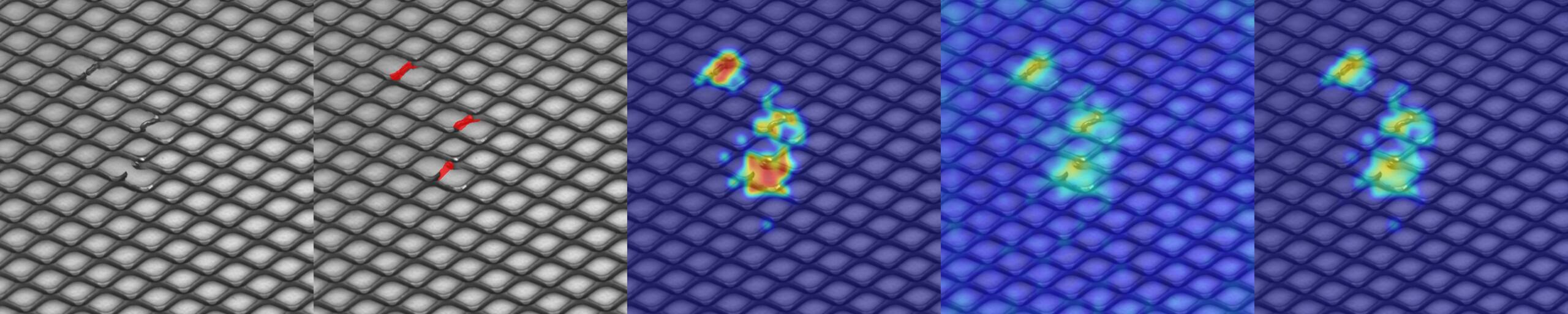}
        \includegraphics[width=\linewidth]{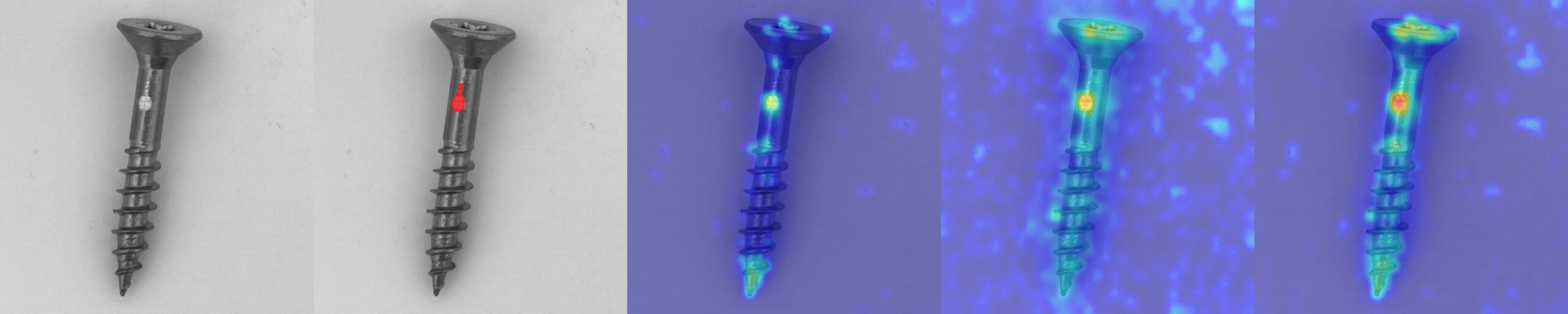}
        \includegraphics[width=\linewidth]{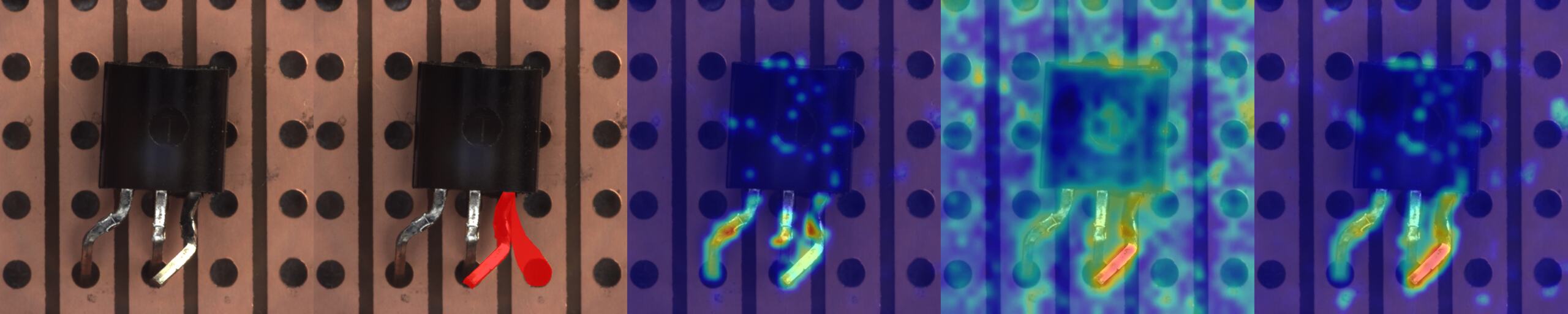}
        \includegraphics[width=\linewidth]{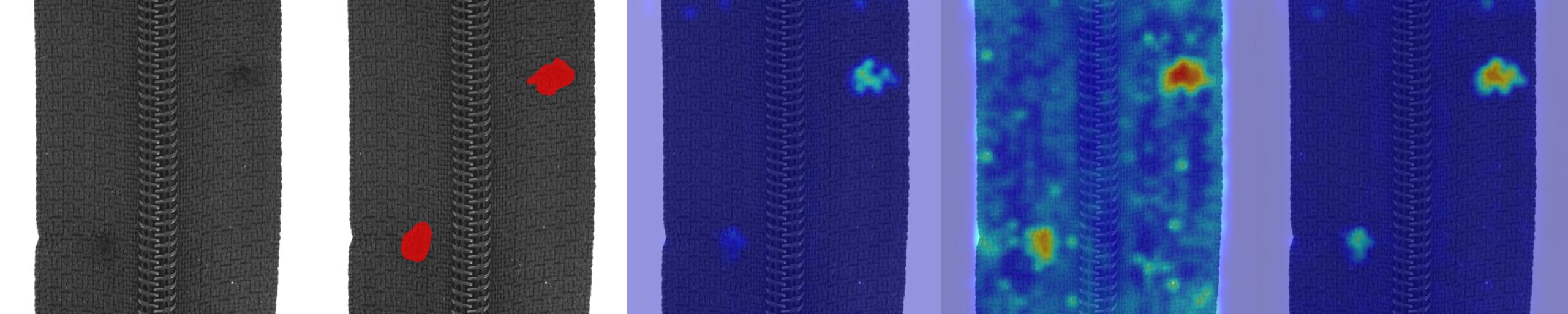}
        \includegraphics[width=\linewidth]{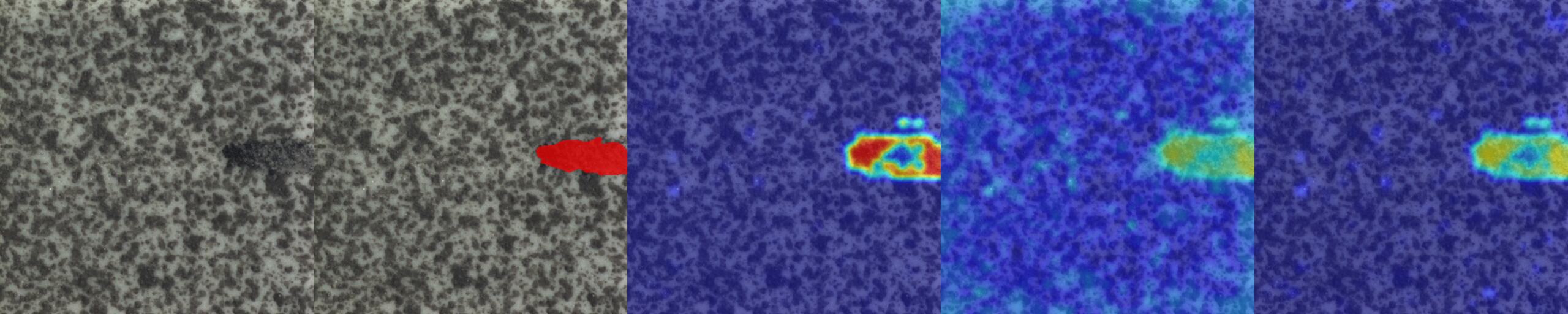}
        \includegraphics[width=\linewidth]{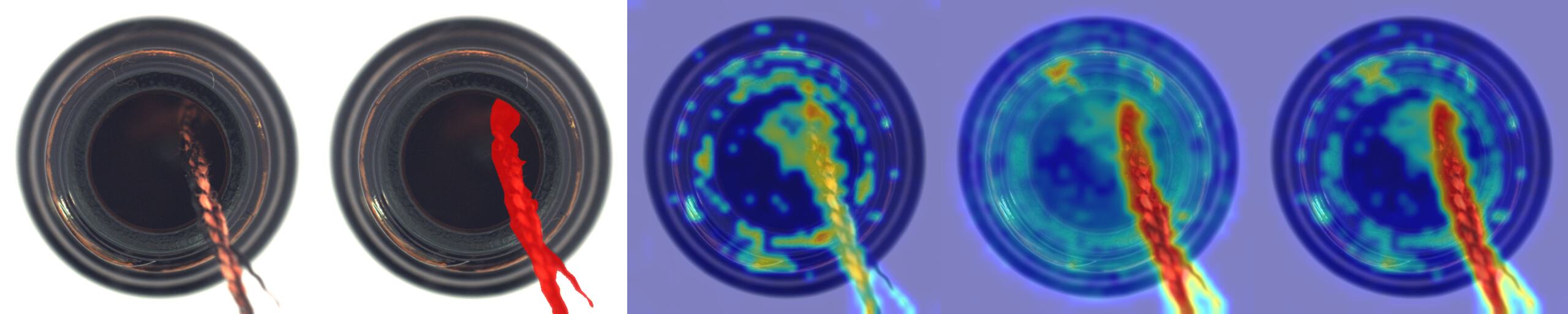}
        \vspace{-0.5cm} 
        
        \begin{tabular}{*{5}{>{\centering\arraybackslash}p{0.15\linewidth}}}
        \scriptsize{Query} &\scriptsize{GT Mask} &\scriptsize{0-shot}  &\scriptsize{1-shot} &\scriptsize{4-shot}  \\
        \end{tabular}
    \end{minipage}
 \vspace{-10pt}   
\caption{\small{Qualitative comparisons of our AdaptCLIP with different prompt numbers on \textbf{MVTec}.}}\label{fig:mvtec}
\vspace{-5pt}
\end{figure*}

\section{More Visualizations}

\begin{figure*}[t]
    \centering
    \begin{minipage}{0.48\linewidth}
        \centering
        \includegraphics[width=\linewidth]{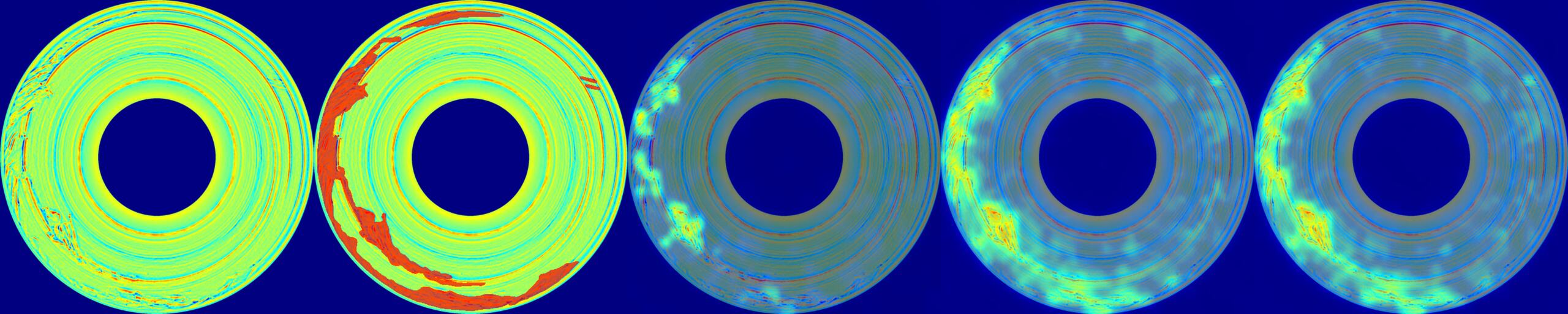}
        \includegraphics[width=\linewidth]{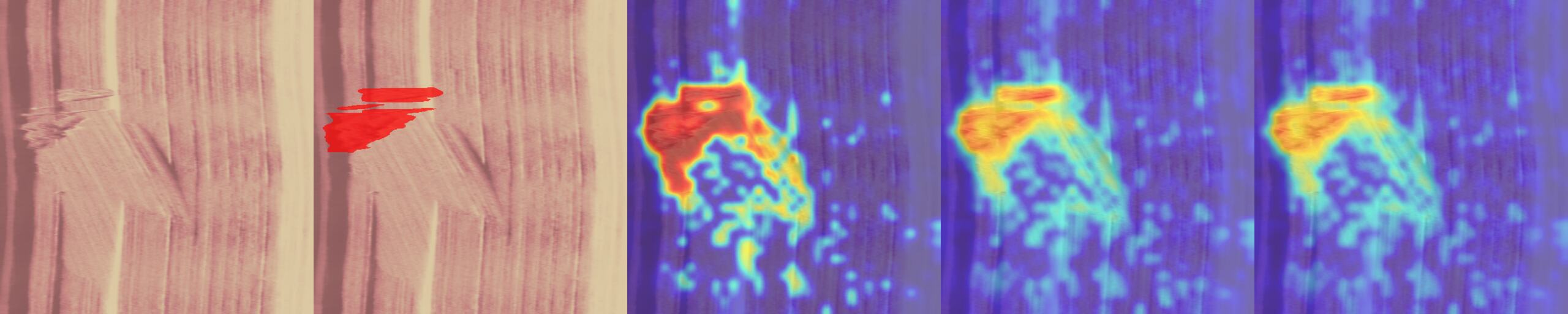}
        \includegraphics[width=\linewidth]{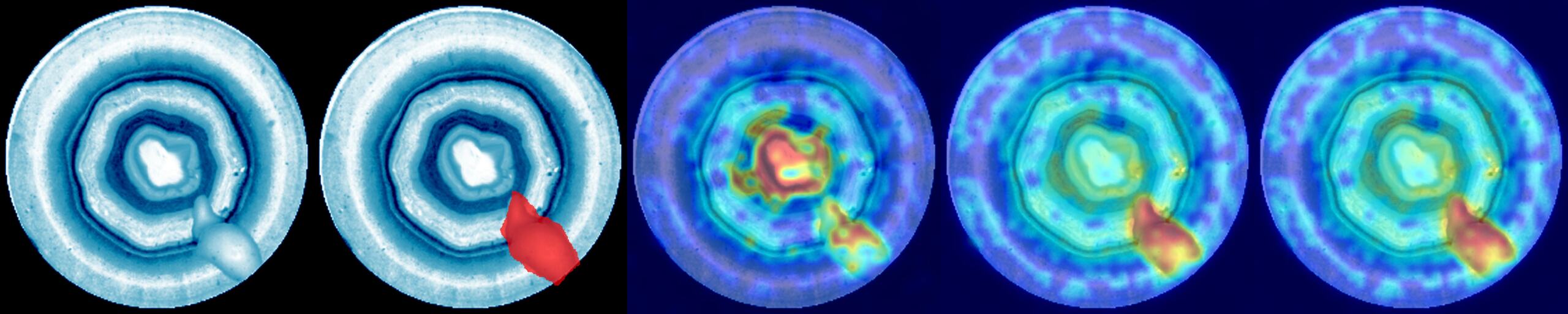}
        \vspace{-0.5cm} 
        
        \begin{tabular}{*{5}{>{\centering\arraybackslash}p{0.15\linewidth}}}
        \scriptsize{Query} &\scriptsize{GT Mask} &\scriptsize{0-shot}  &\scriptsize{1-shot} &\scriptsize{4-shot}  \\
        \end{tabular}
    \end{minipage}
    \hspace{0.0001\linewidth} 
    \begin{minipage}{0.48\linewidth}
        \centering
         \includegraphics[width=\linewidth]{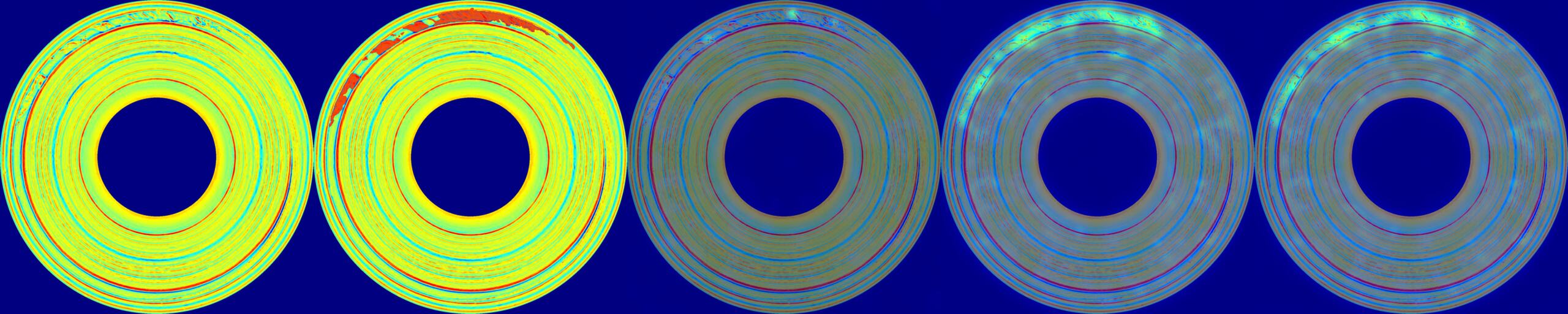}
         \includegraphics[width=\linewidth]{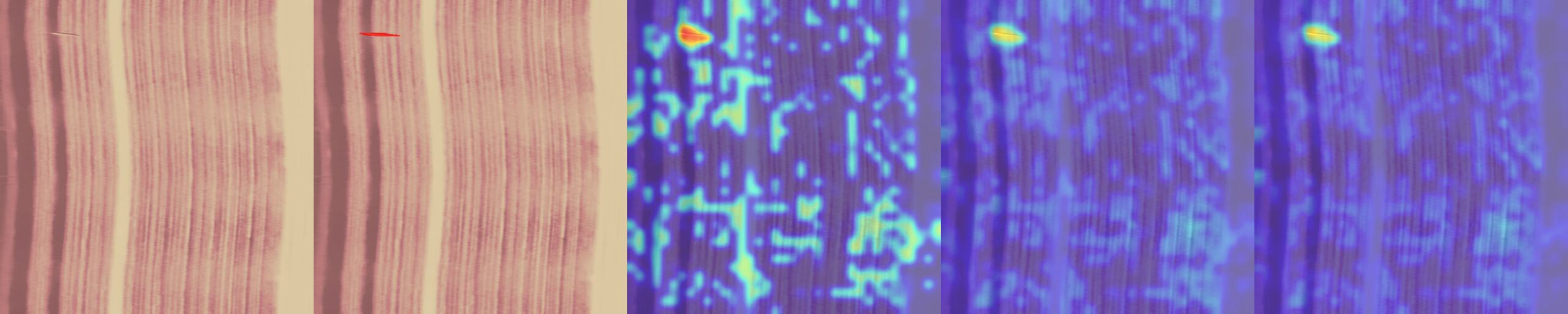}
        \includegraphics[width=\linewidth]{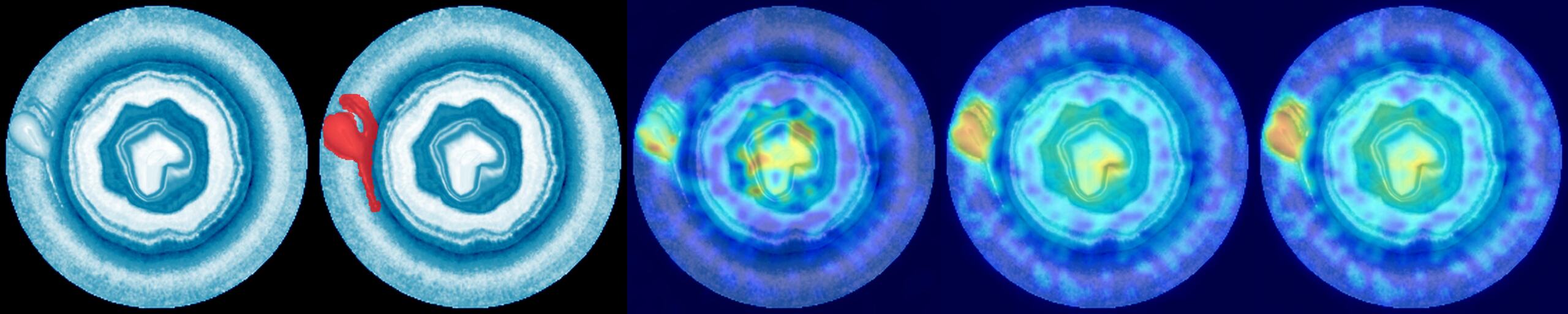}
        \vspace{-0.5cm} 
        
        \begin{tabular}{*{5}{>{\centering\arraybackslash}p{0.15\linewidth}}}
        \scriptsize{Query} &\scriptsize{GT Mask} &\scriptsize{0-shot}  &\scriptsize{1-shot} &\scriptsize{4-shot}  \\
        \end{tabular}
    \end{minipage}
 \vspace{-5pt}   
\caption{\small{Qualitative comparisons of our AdaptCLIP with different prompt numbers on \textbf{BTAD}.}}\label{fig:btad}
\vspace{-10pt}
\end{figure*}

\begin{figure*}[t]
    \centering
    \begin{minipage}{0.48\linewidth}
        \centering
        \includegraphics[width=\linewidth]{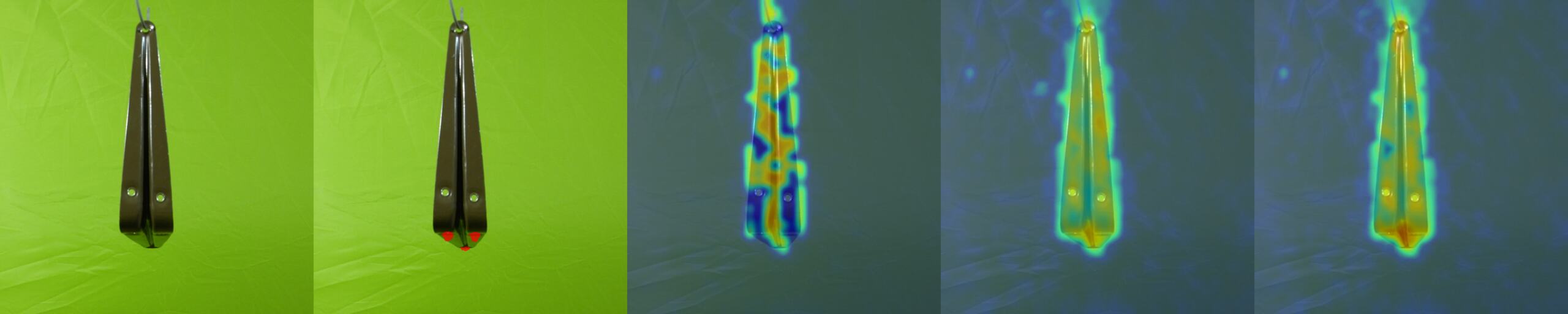}
        \includegraphics[width=\linewidth]{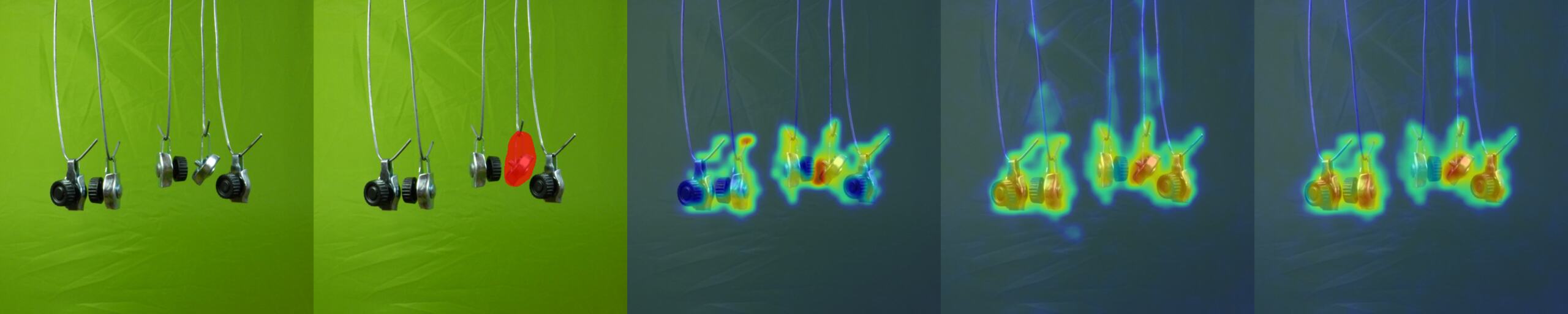}
        \includegraphics[width=\linewidth]{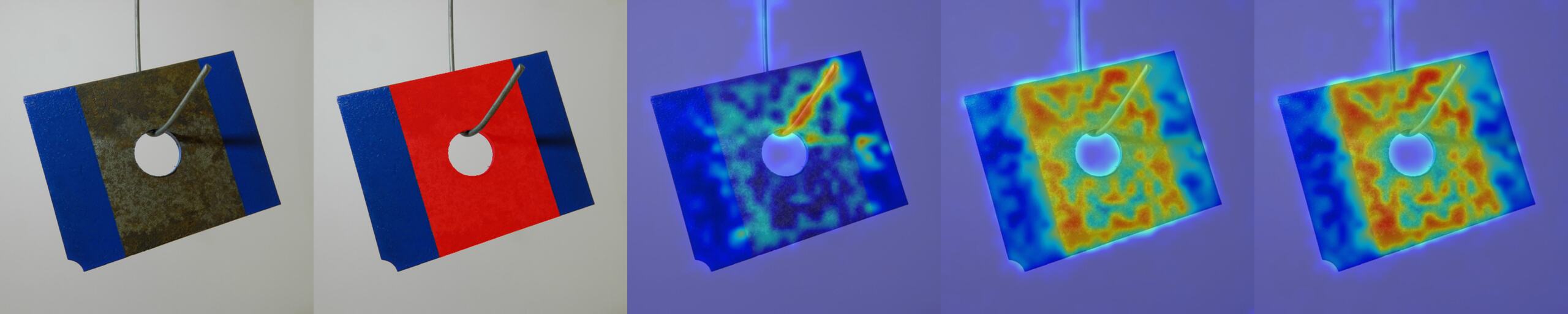}
        \vspace{-0.5cm} 
        
        \begin{tabular}{*{5}{>{\centering\arraybackslash}p{0.15\linewidth}}}
        \scriptsize{Query} &\scriptsize{GT Mask} &\scriptsize{0-shot}  &\scriptsize{1-shot} &\scriptsize{4-shot}  \\
        \end{tabular}
    \end{minipage}
    \hspace{0.0001\linewidth} 
    \begin{minipage}{0.48\linewidth}
        \centering
         \includegraphics[width=\linewidth]{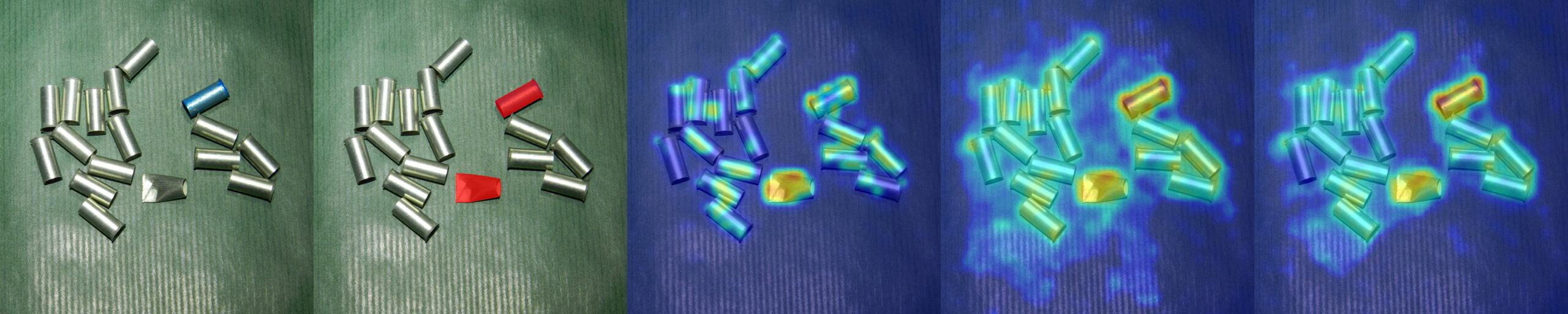}
         \includegraphics[width=\linewidth]{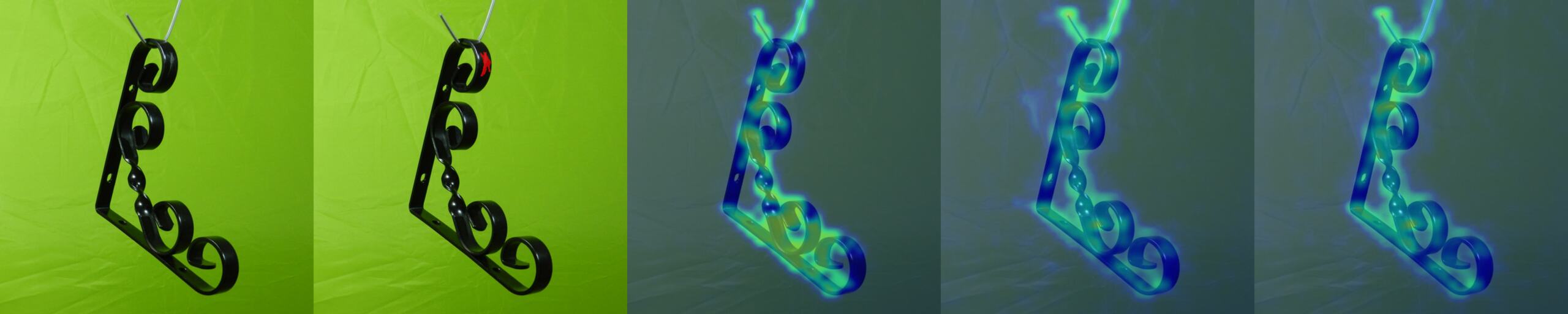}
        \includegraphics[width=\linewidth]{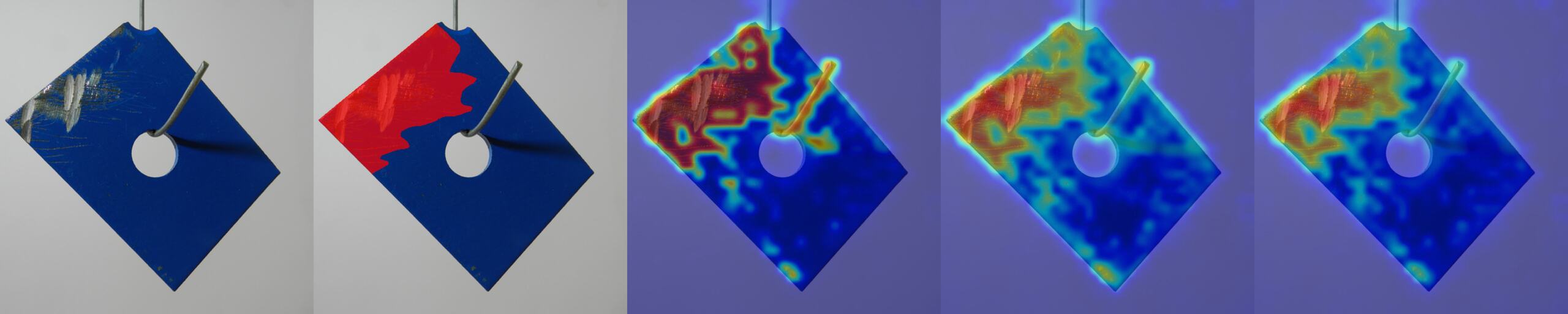}
        \vspace{-0.5cm} 
        
        \begin{tabular}{*{5}{>{\centering\arraybackslash}p{0.15\linewidth}}}
        \scriptsize{Query} &\scriptsize{GT Mask} &\scriptsize{0-shot}  &\scriptsize{1-shot} &\scriptsize{4-shot}  \\
        \end{tabular}
    \end{minipage}
 \vspace{-10pt}   
\caption{\small{Qualitative comparisons of our AdaptCLIP with different prompt numbers on \textbf{MPDD}.}}\label{fig:mpdd}
\vspace{-10pt}
\end{figure*}

\begin{figure*}[t]
    \centering
    \begin{minipage}{0.48\linewidth}
        \centering
        \includegraphics[width=\linewidth]{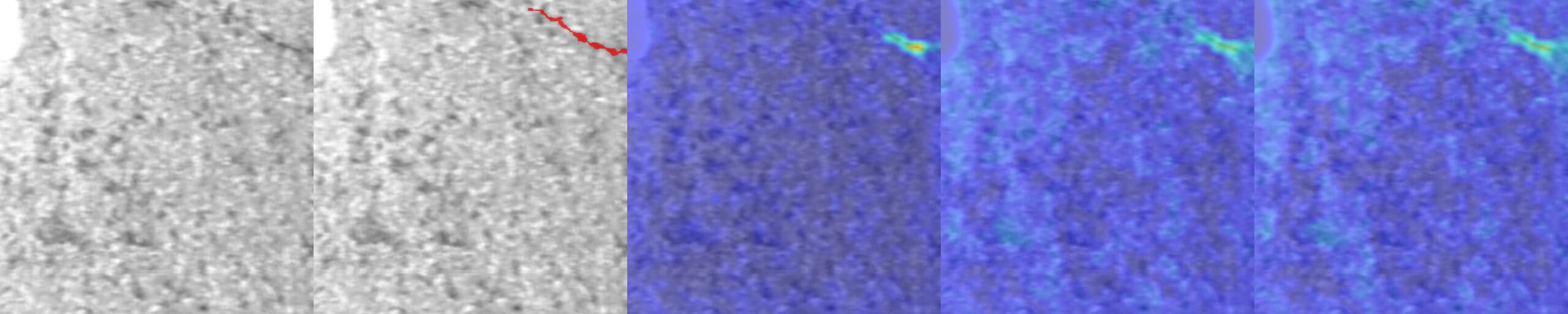}
        \includegraphics[width=\linewidth]{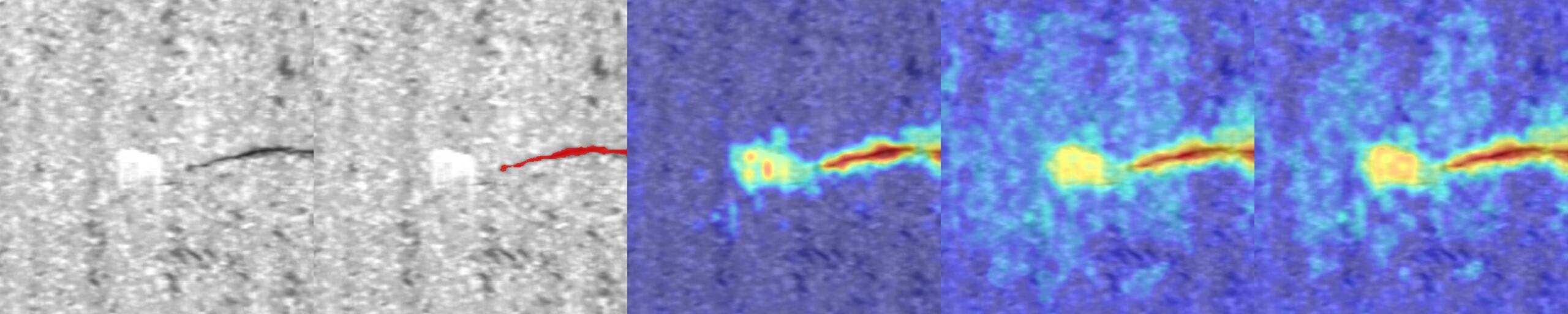}
        \vspace{-0.5cm} 

        \begin{tabular}{*{5}{>{\centering\arraybackslash}p{0.15\linewidth}}}
        \scriptsize{Query} &\scriptsize{GT Mask} &\scriptsize{0-shot}  &\scriptsize{1-shot} &\scriptsize{4-shot}  \\
        \end{tabular}
    \end{minipage}
    \hspace{0.0001\linewidth} 
    \begin{minipage}{0.48\linewidth}
        \centering
         \includegraphics[width=\linewidth]{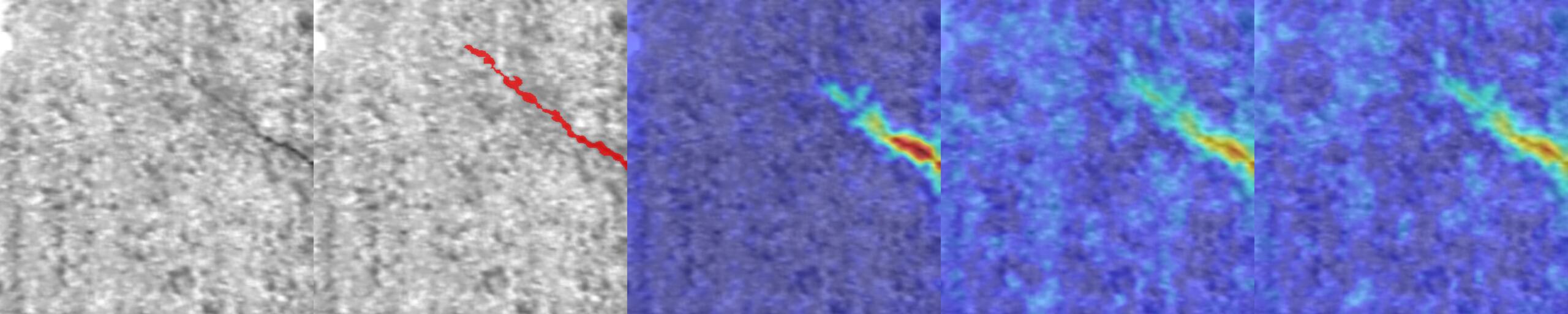}
         \includegraphics[width=\linewidth]{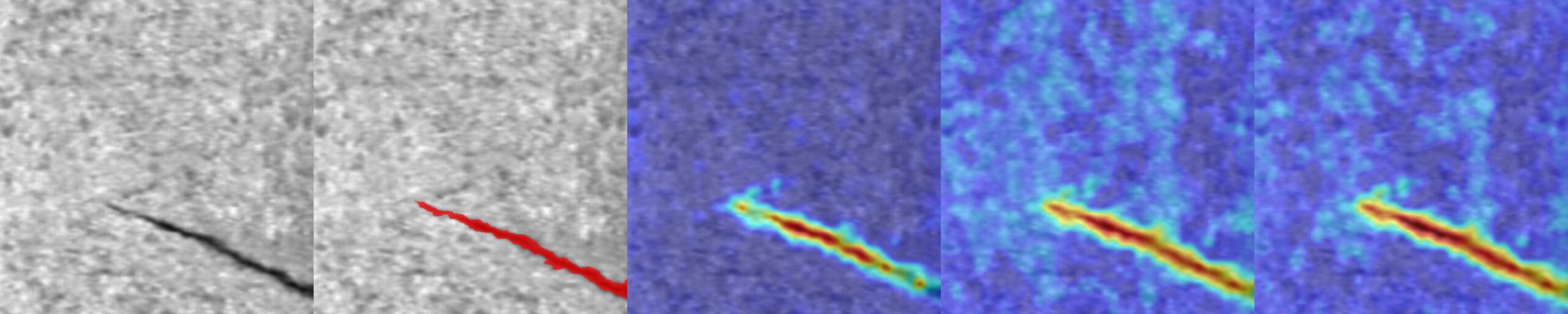}
        \vspace{-0.5cm} 
        
        \begin{tabular}{*{5}{>{\centering\arraybackslash}p{0.15\linewidth}}}
        \scriptsize{Query} &\scriptsize{GT Mask} &\scriptsize{0-shot}  &\scriptsize{1-shot} &\scriptsize{4-shot}  \\
        \end{tabular}
    \end{minipage}
 \vspace{-10pt}   
\caption{\small{Qualitative comparisons of our AdaptCLIP with different prompt numbers on \textbf{KSDD}.}}\label{fig:ksdd}
\vspace{-10pt}
\end{figure*}

\begin{figure*}[t]
    \centering
    \begin{minipage}{0.48\linewidth}
        \centering
        \includegraphics[width=\linewidth]{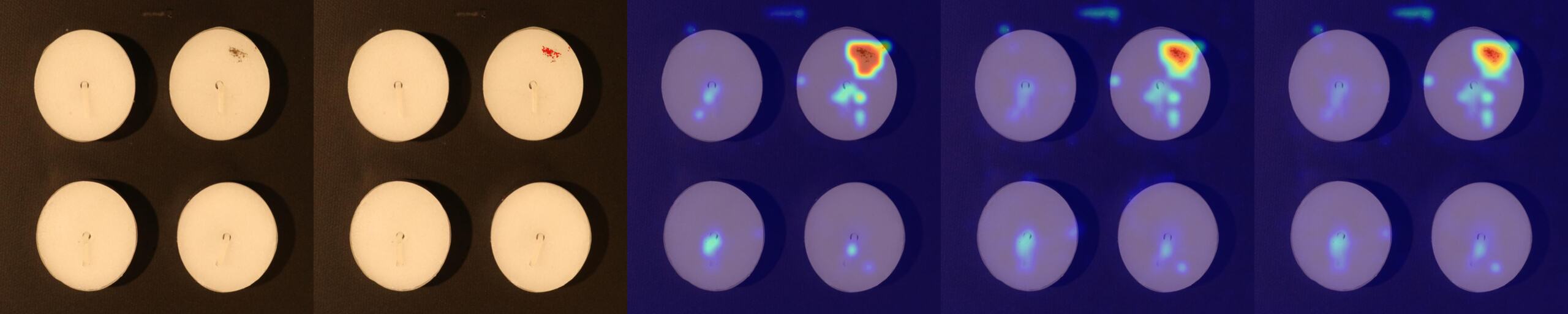}
        \includegraphics[width=\linewidth]{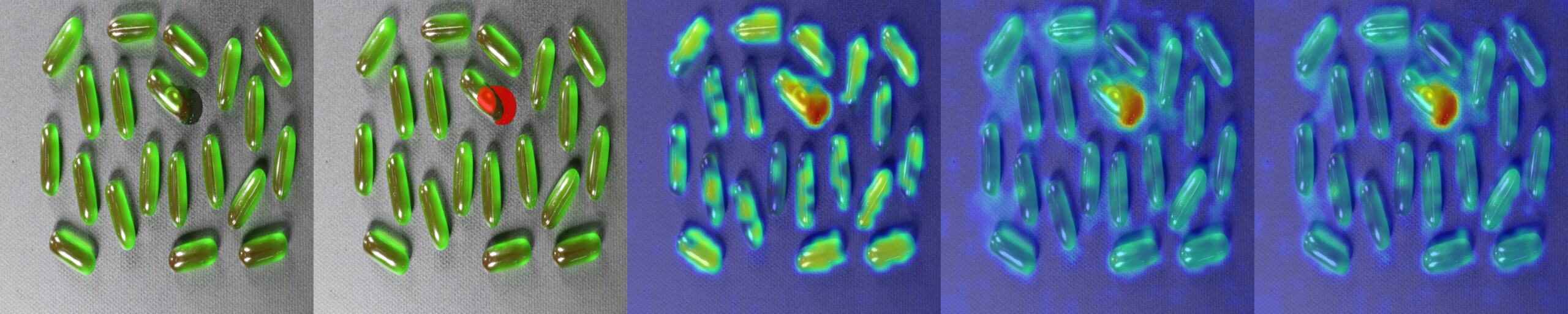}
        \includegraphics[width=\linewidth]{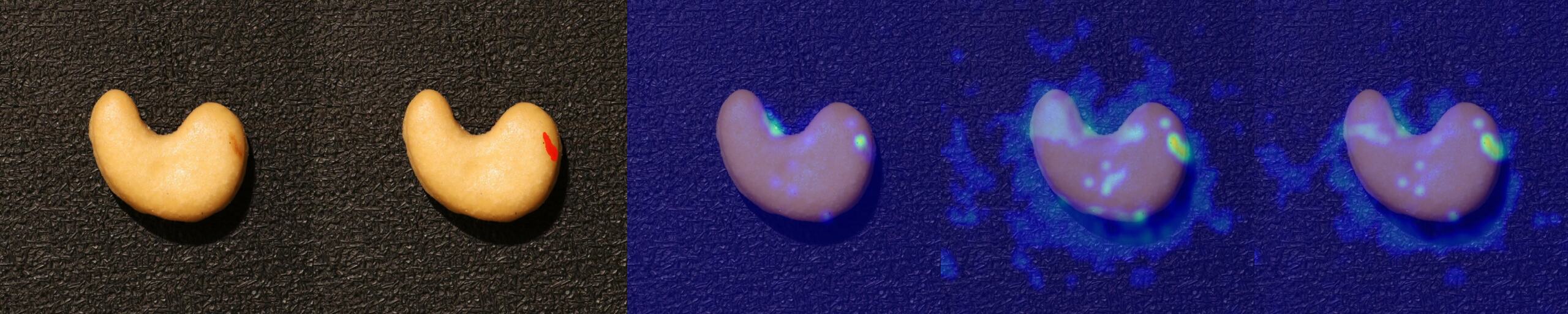}
        \includegraphics[width=\linewidth]{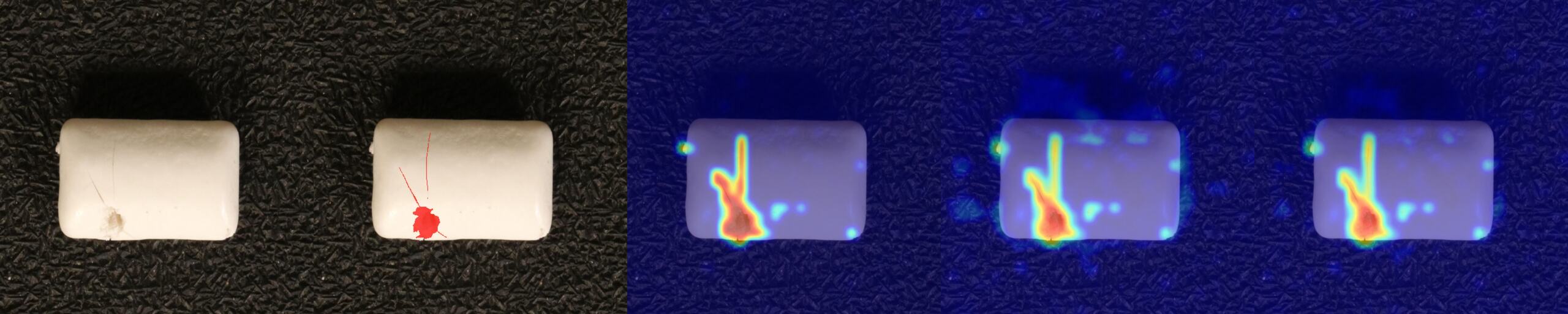}
        \includegraphics[width=\linewidth]{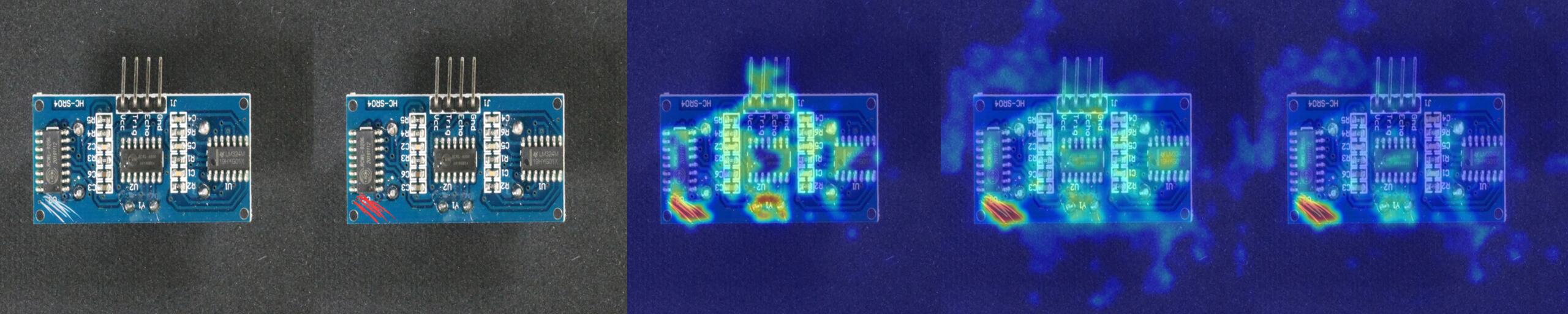}
        \includegraphics[width=\linewidth]{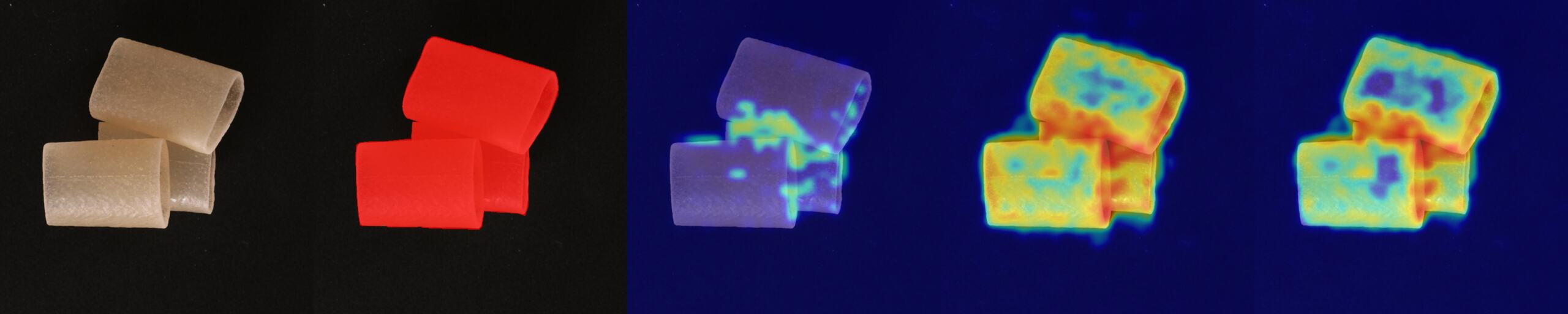}
        \vspace{-0.5cm} 
        
        \begin{tabular}{*{5}{>{\centering\arraybackslash}p{0.15\linewidth}}}
        \scriptsize{Query} &\scriptsize{GT Mask} &\scriptsize{0-shot}  &\scriptsize{1-shot} &\scriptsize{4-shot}  \\
        \end{tabular}
    \end{minipage}
    \hspace{0.0001\linewidth} 
    \begin{minipage}{0.48\linewidth}
        \centering
         \includegraphics[width=\linewidth]{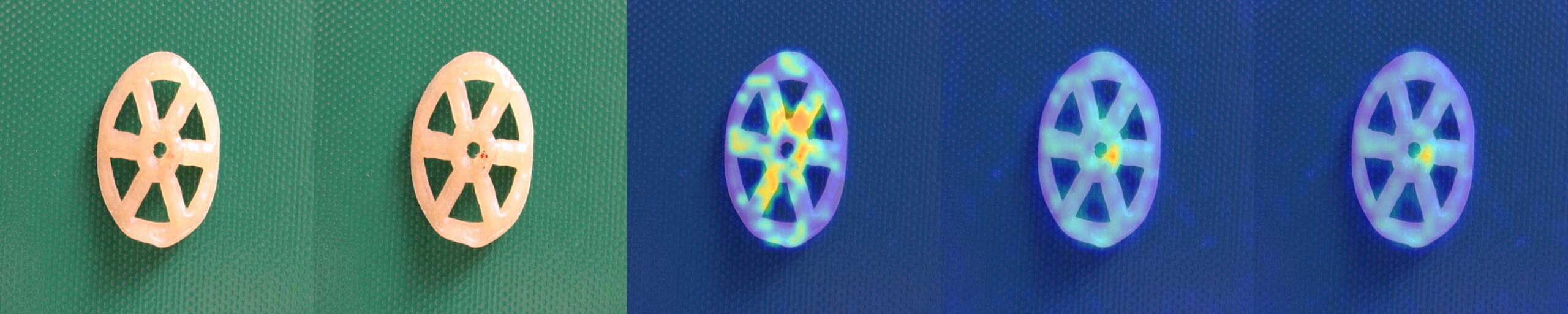}
         \includegraphics[width=\linewidth]{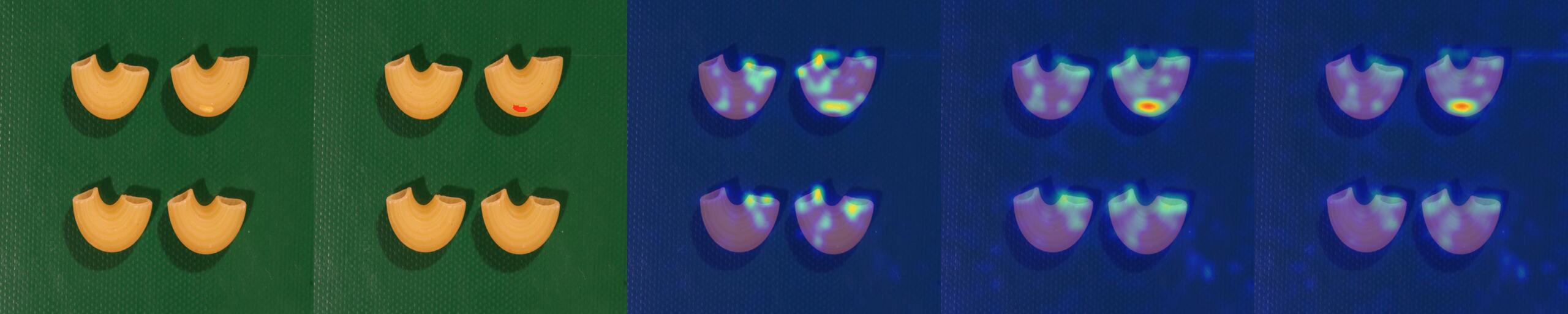}
        \includegraphics[width=\linewidth]{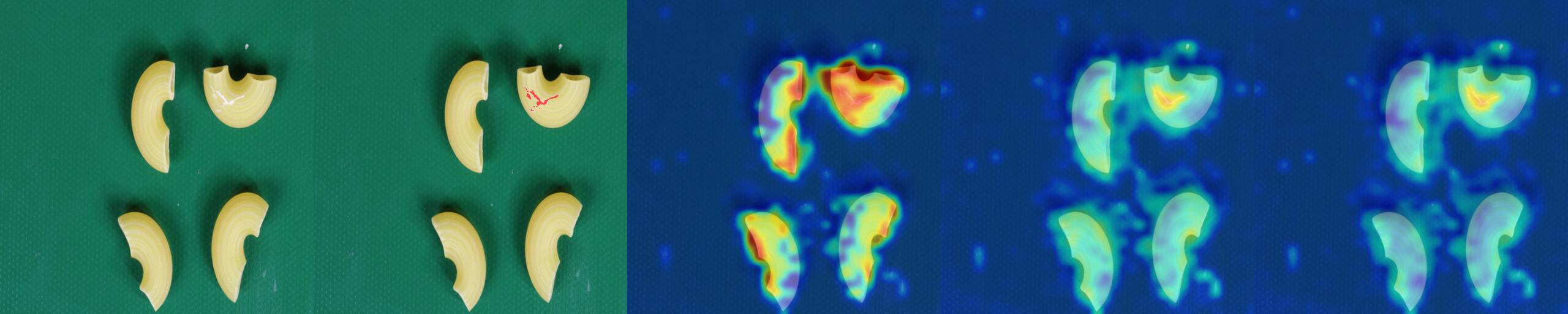}
        \includegraphics[width=\linewidth]{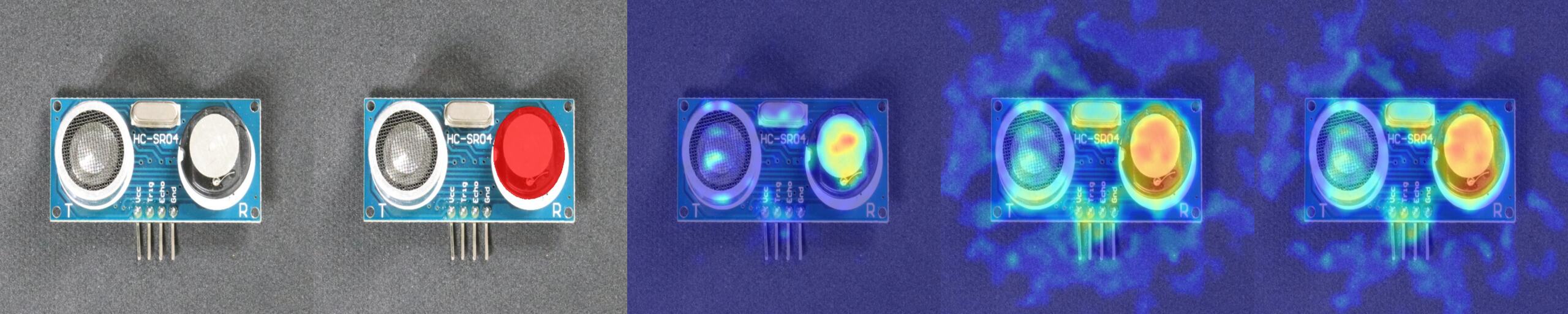}
        
        \includegraphics[width=\linewidth]{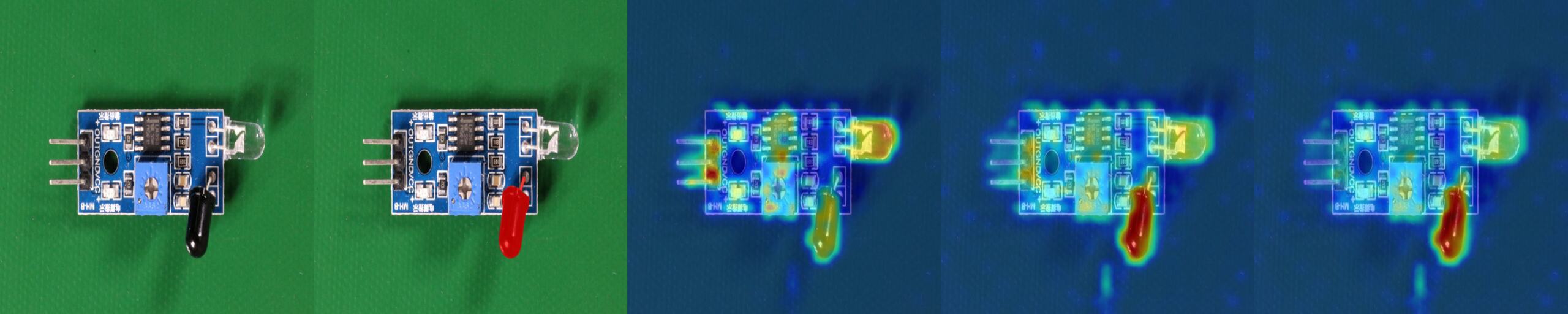}
        \includegraphics[width=\linewidth]{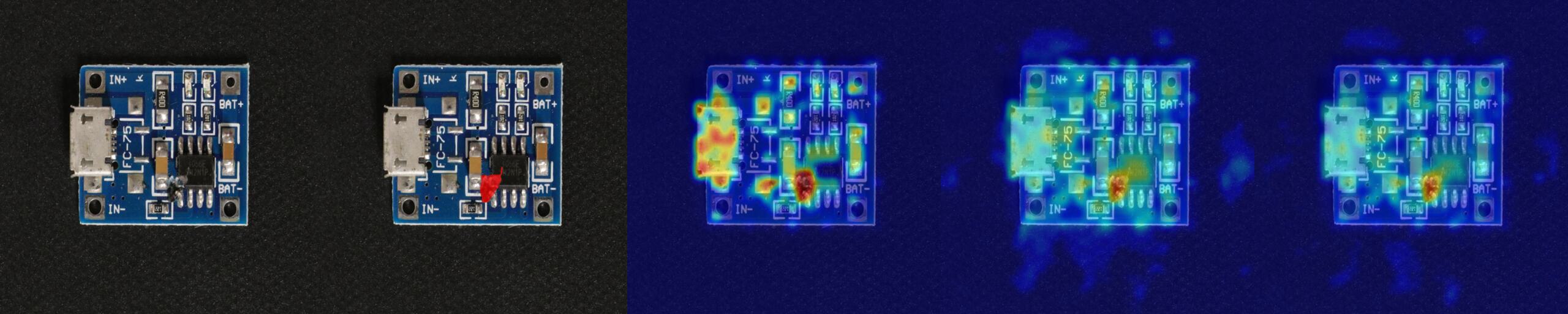}
        \vspace{-0.5cm} 
        
        \begin{tabular}{*{5}{>{\centering\arraybackslash}p{0.15\linewidth}}}
        \scriptsize{Query} &\scriptsize{GT Mask} &\scriptsize{0-shot}  &\scriptsize{1-shot} &\scriptsize{4-shot}  \\
        \end{tabular}
    \end{minipage}
 \vspace{-5pt}   
\caption{\small{Qualitative comparisons of our AdaptCLIP with different prompt numbers on \textbf{VisA}.}}\label{fig:visa}
\vspace{-10pt}
\end{figure*}

\begin{figure*}[t]
    \centering
    \begin{minipage}{0.48\linewidth}
        \centering
        \includegraphics[width=\linewidth]{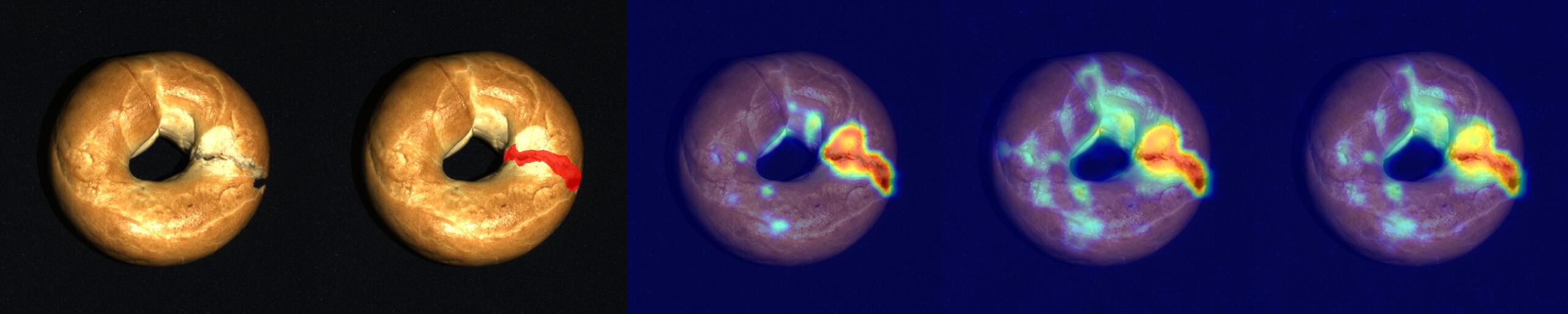}
        \includegraphics[width=\linewidth]{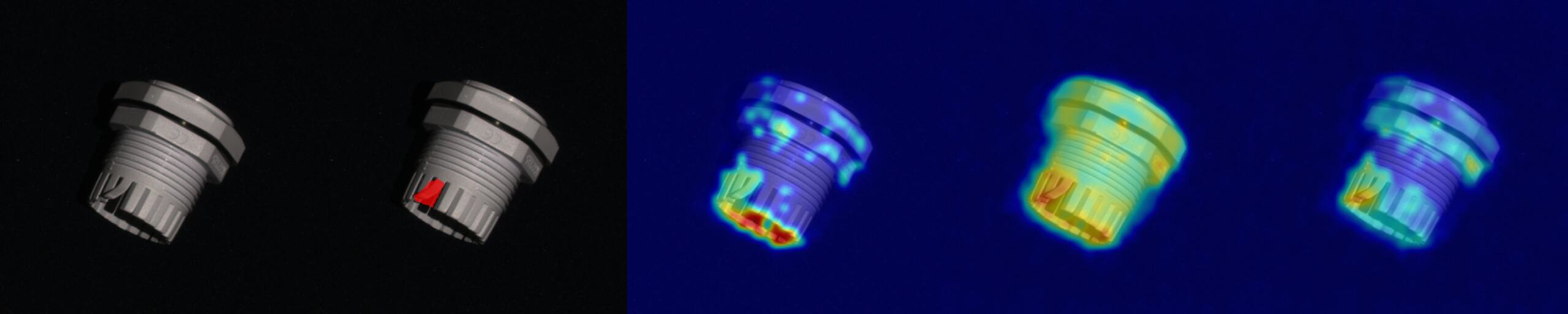}
        \includegraphics[width=\linewidth]{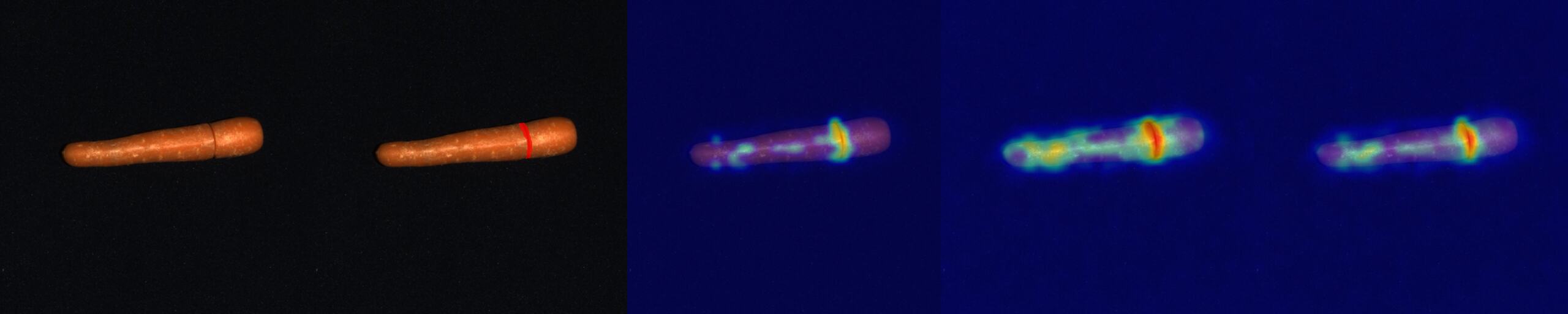}
        \includegraphics[width=\linewidth]{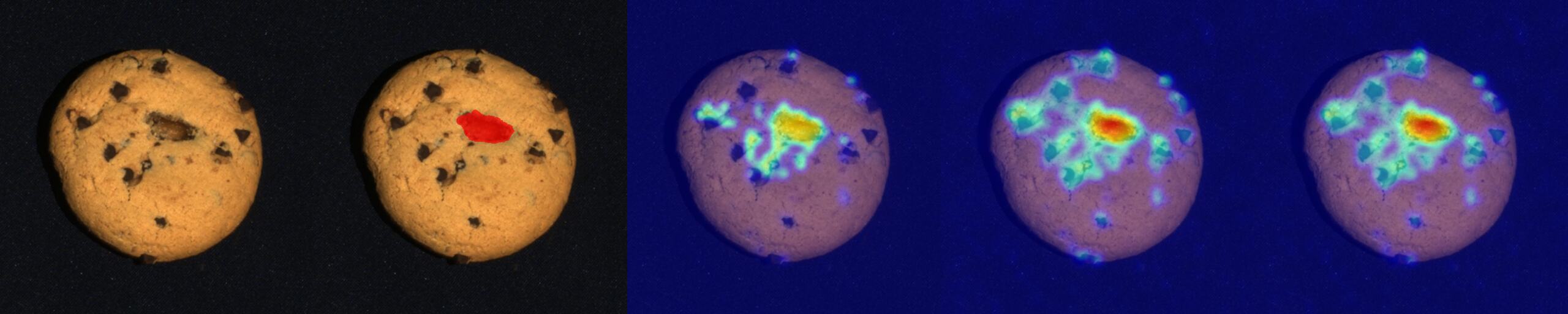}
        \includegraphics[width=\linewidth]{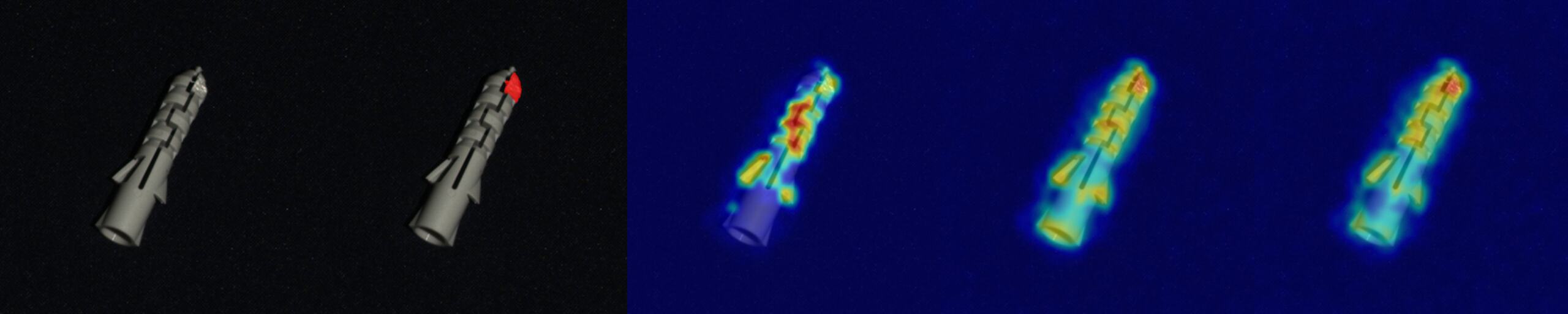}
        \vspace{-0.5cm} 
        
        \begin{tabular}{*{5}{>{\centering\arraybackslash}p{0.15\linewidth}}}
        \scriptsize{Query} &\scriptsize{GT Mask} &\scriptsize{0-shot}  &\scriptsize{1-shot} &\scriptsize{4-shot}  \\
        \end{tabular}
    \end{minipage}
    \hspace{0.0001\linewidth} 
    \begin{minipage}{0.48\linewidth}
        \centering
         \includegraphics[width=\linewidth]{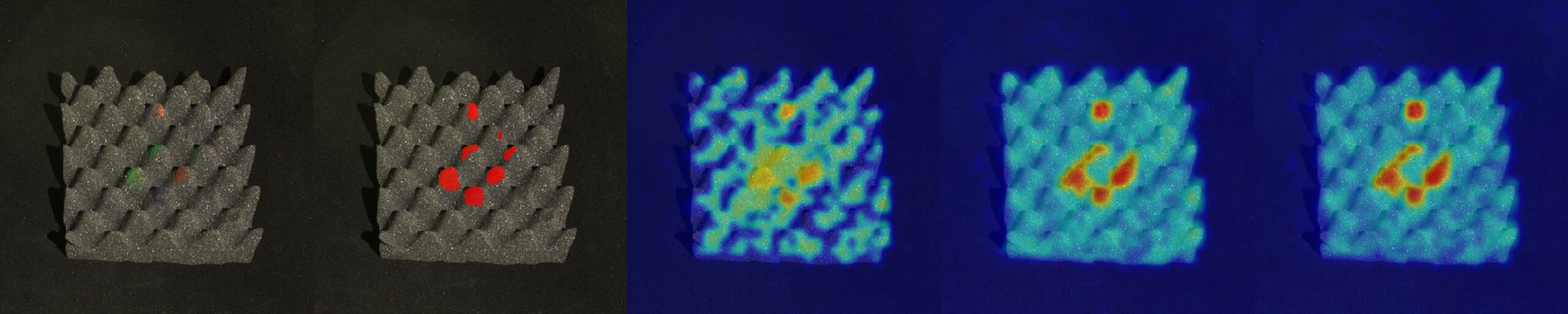}
         \includegraphics[width=\linewidth]{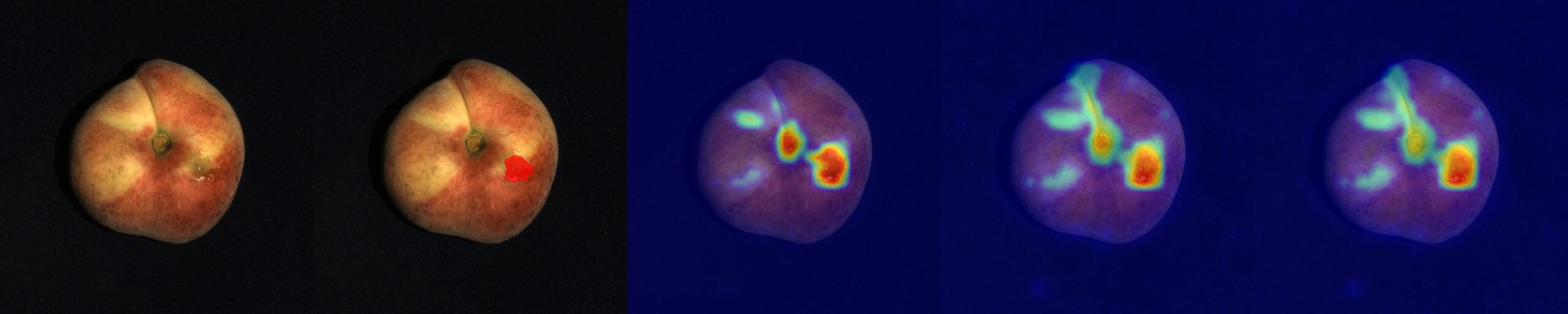}
        \includegraphics[width=\linewidth]{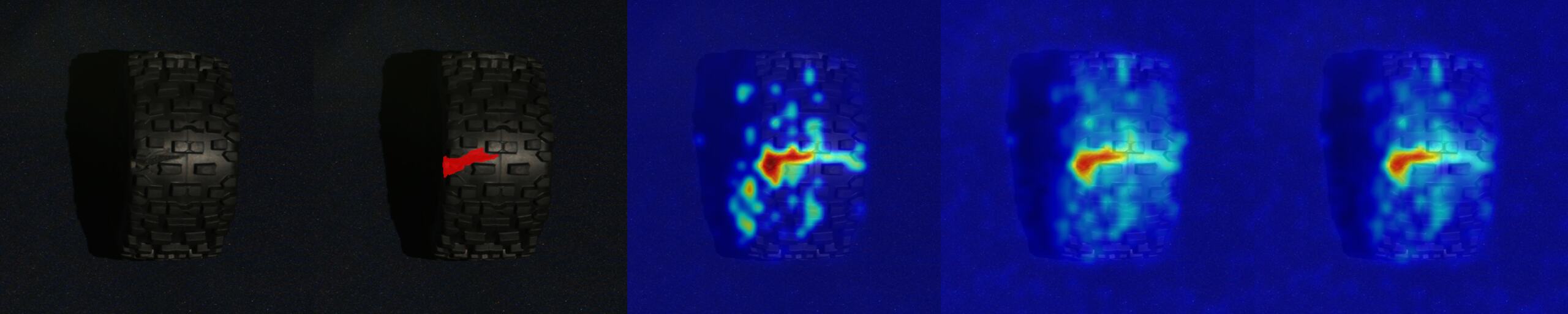}
        \includegraphics[width=\linewidth]{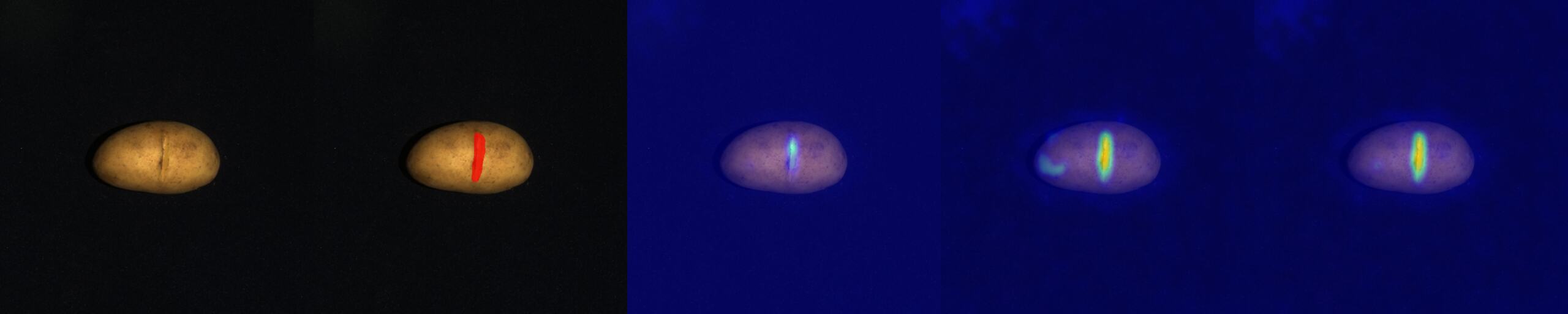}
        \includegraphics[width=\linewidth]{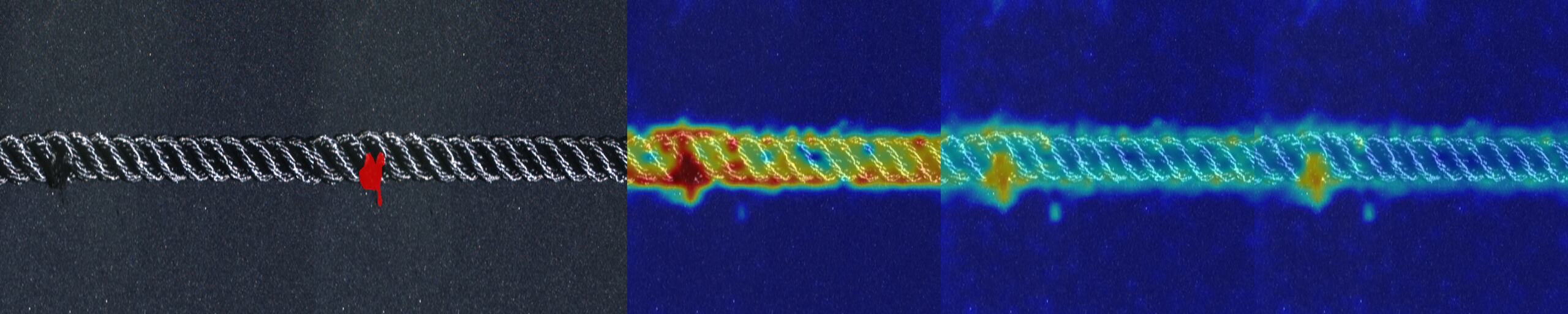}
        \vspace{-0.5cm} 
        
        \begin{tabular}{*{5}{>{\centering\arraybackslash}p{0.15\linewidth}}}
        \scriptsize{Query} &\scriptsize{GT Mask} &\scriptsize{0-shot}  &\scriptsize{1-shot} &\scriptsize{4-shot}  \\
        \end{tabular}
    \end{minipage}
 \vspace{-5pt}   
\caption{\small{Qualitative comparisons of our AdaptCLIP with different prompt numbers on \textbf{MVTec 3D}.}}\label{fig:mvtec3d}
\vspace{-10pt}
\end{figure*}

\begin{figure*}[t]
    \centering
    \begin{minipage}{0.48\linewidth}
        \centering
        \includegraphics[width=\linewidth]{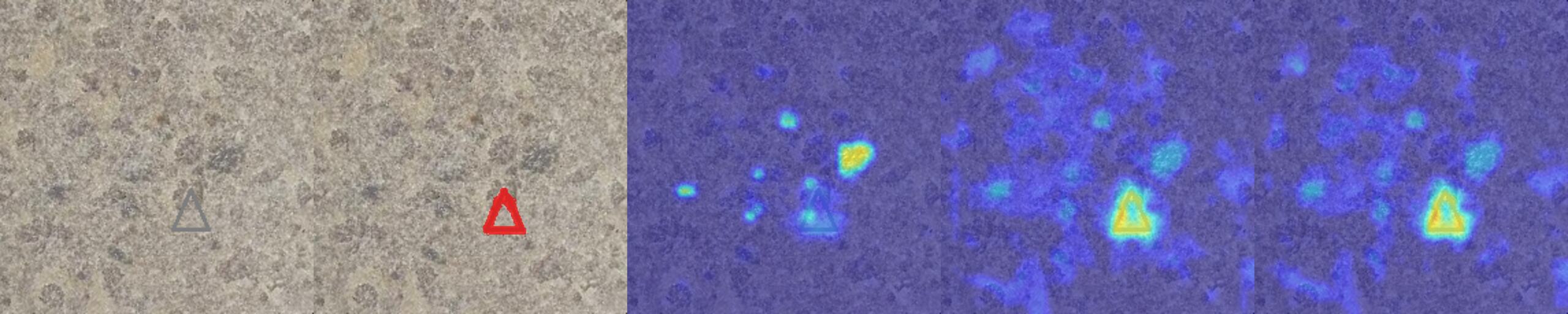}
        \includegraphics[width=\linewidth]{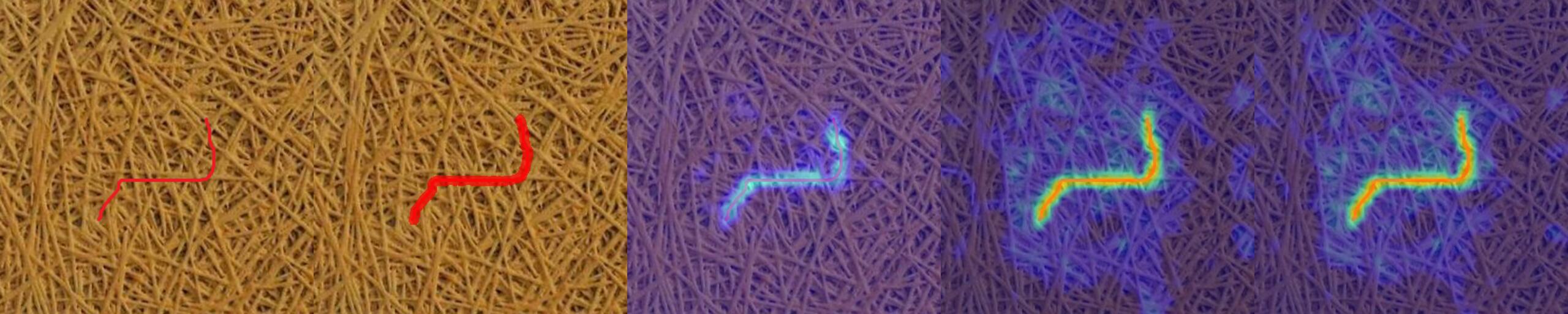}
        \includegraphics[width=\linewidth]{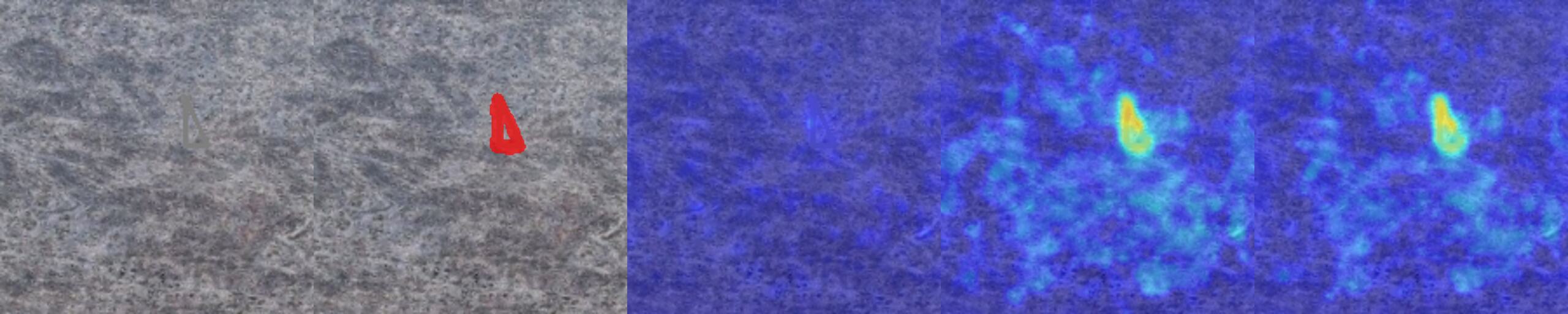}
        \includegraphics[width=\linewidth]{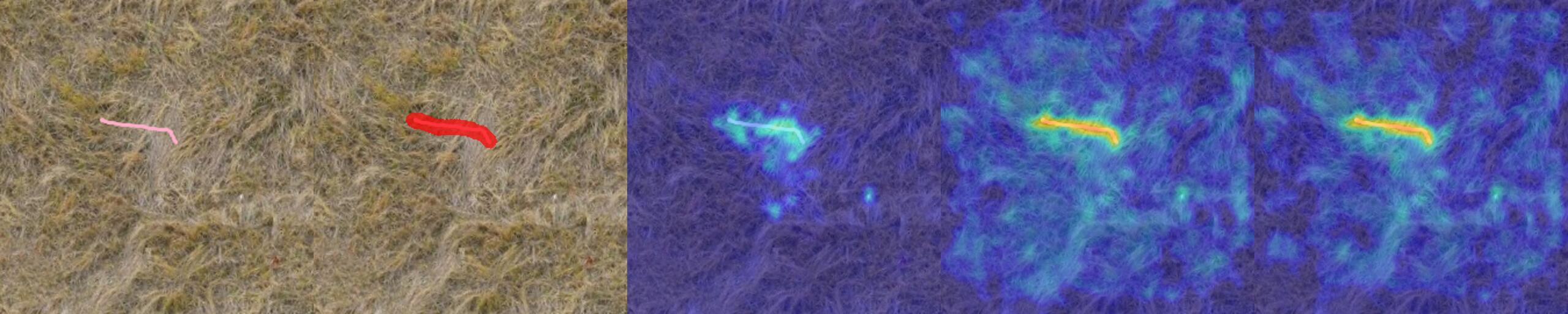}
        \includegraphics[width=\linewidth]{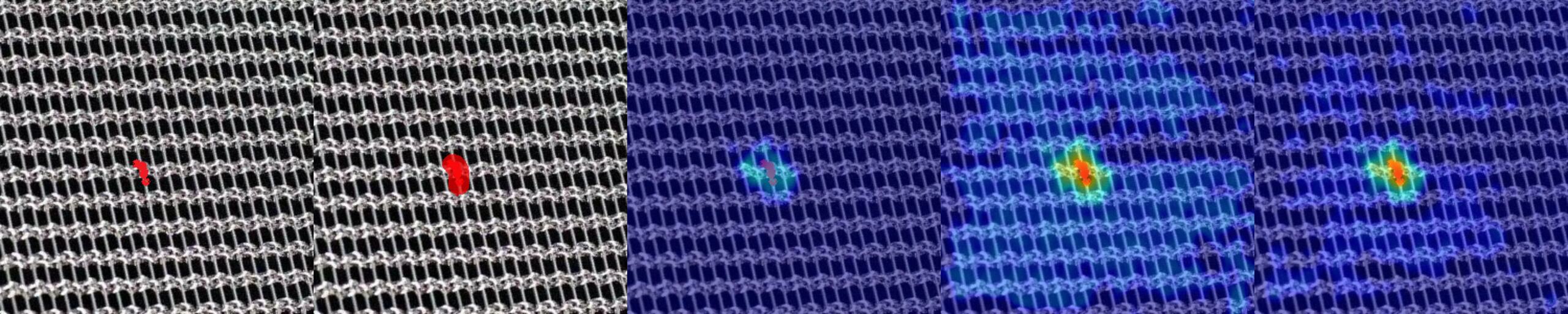}
        \includegraphics[width=\linewidth]{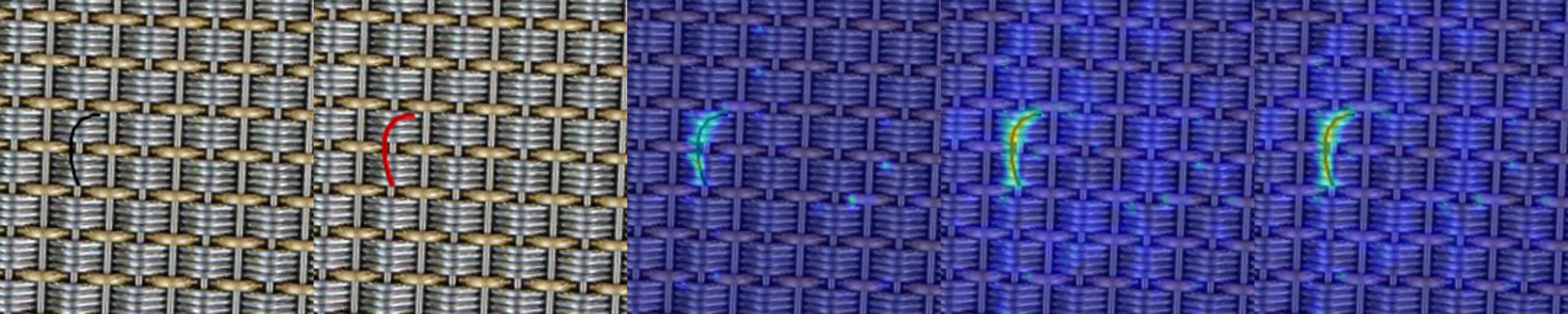}
        \vspace{-0.5cm} 
        
        \begin{tabular}{*{5}{>{\centering\arraybackslash}p{0.15\linewidth}}}
        \scriptsize{Query} &\scriptsize{GT Mask} &\scriptsize{0-shot}  &\scriptsize{1-shot} &\scriptsize{4-shot}  \\
        \end{tabular}
    \end{minipage}
    \hspace{0.0001\linewidth} 
    \begin{minipage}{0.48\linewidth}
        \centering
         \includegraphics[width=\linewidth]{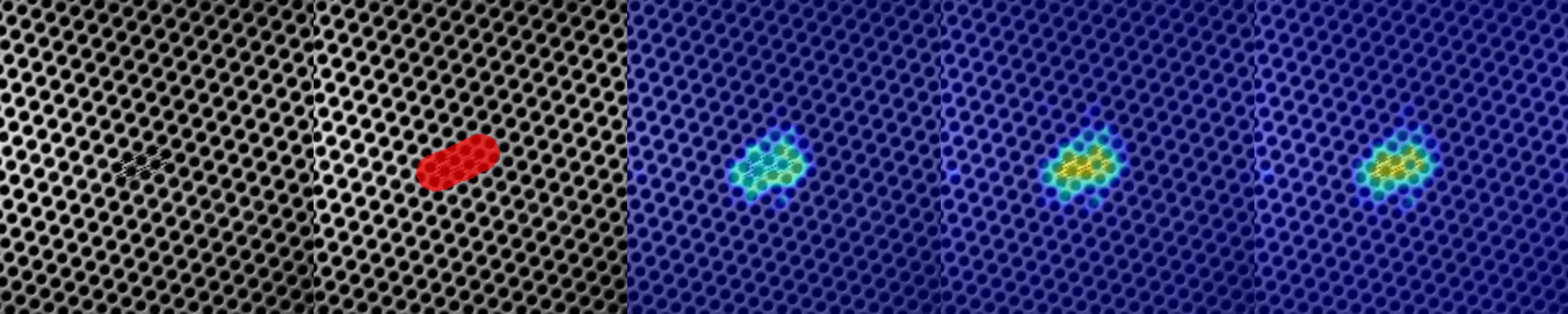}
         \includegraphics[width=\linewidth]{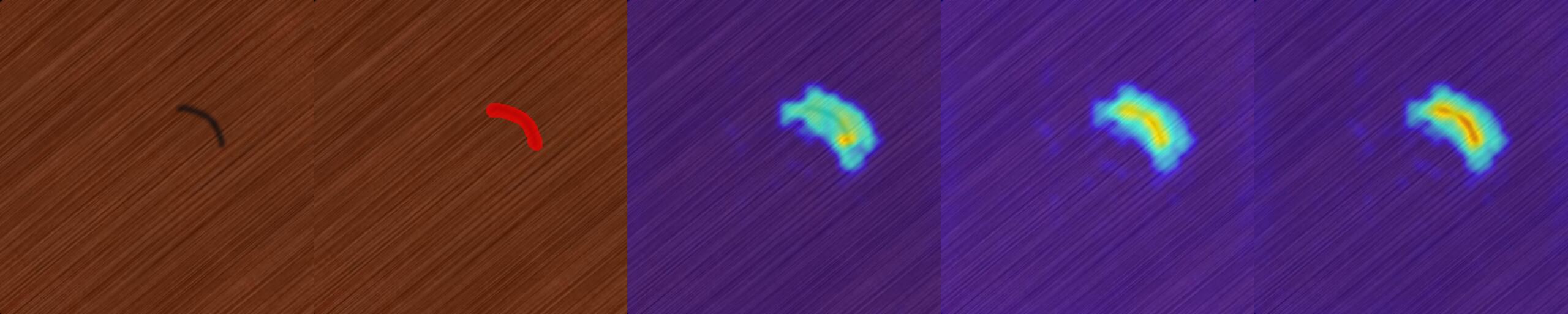}
        \includegraphics[width=\linewidth]{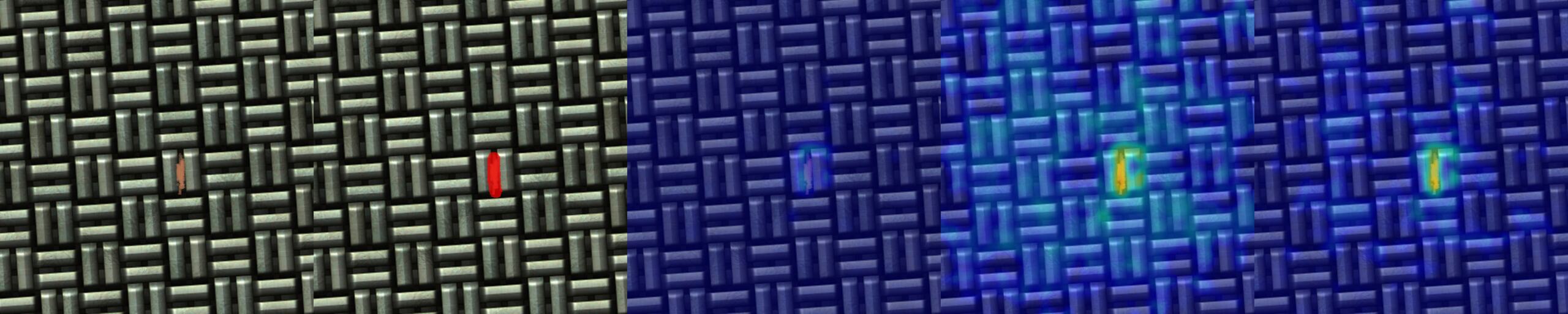}
        \includegraphics[width=\linewidth]{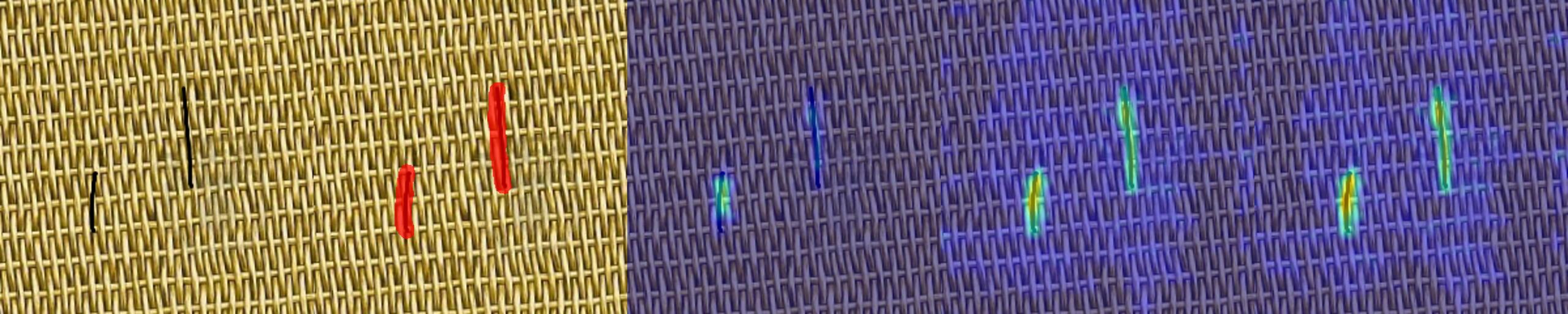}
        \includegraphics[width=\linewidth]{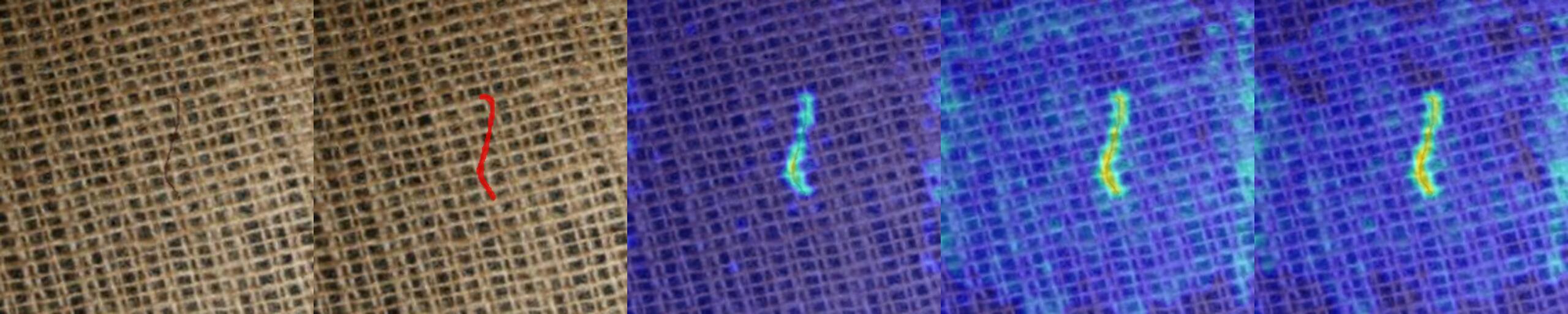}
        \includegraphics[width=\linewidth]{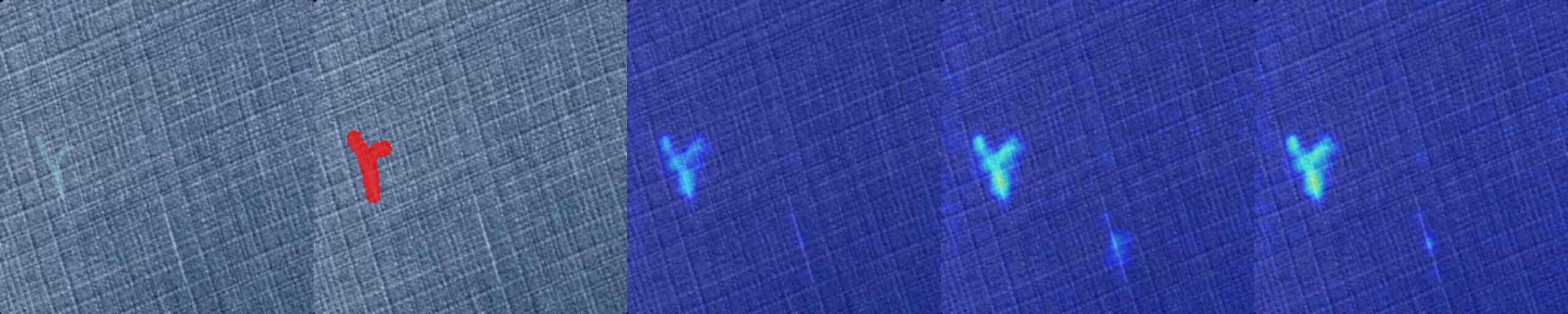}
        \vspace{-0.5cm} 
        
        \begin{tabular}{*{5}{>{\centering\arraybackslash}p{0.15\linewidth}}}
        \scriptsize{Query} &\scriptsize{GT Mask} &\scriptsize{0-shot}  &\scriptsize{1-shot} &\scriptsize{4-shot}  \\
        \end{tabular}
    \end{minipage}
 \vspace{-5pt}   
\caption{\small{Qualitative comparisons of our AdaptCLIP with different prompt numbers on \textbf{DTD}.}}\label{fig:dtd}
\vspace{-10pt}
\end{figure*}

\begin{figure*}[t]
    \centering
    \begin{minipage}{0.48\linewidth}
        \centering
        \includegraphics[width=\linewidth]{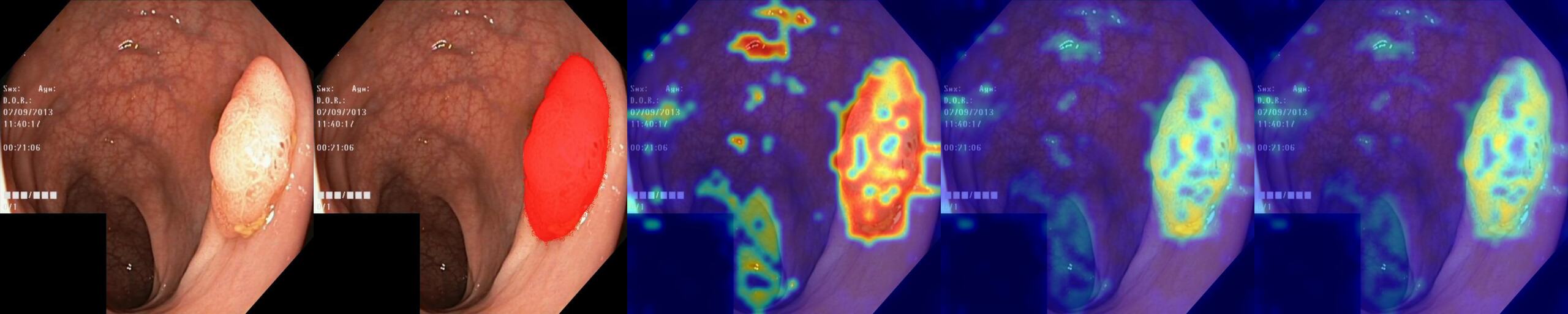}
        \includegraphics[width=\linewidth]{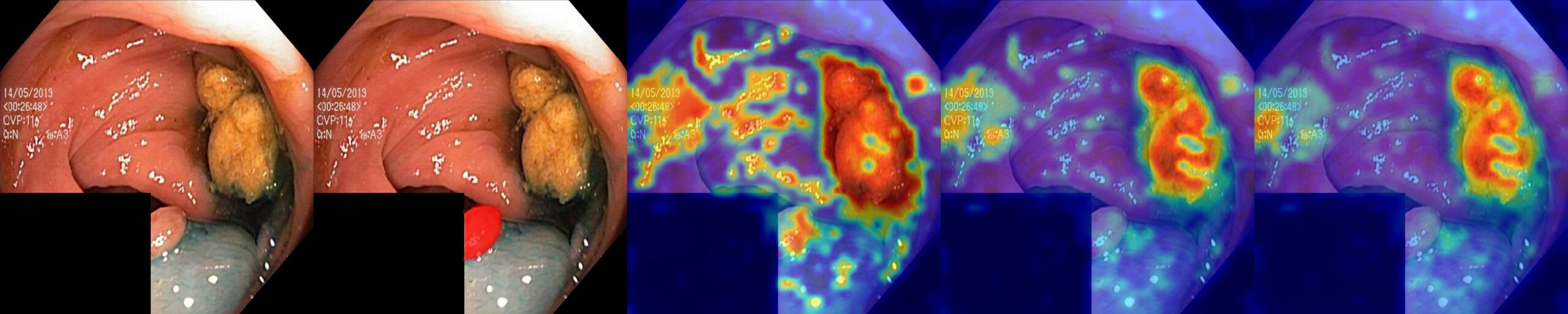}
        \vspace{-0.5cm} 
        
        \begin{tabular}{*{5}{>{\centering\arraybackslash}p{0.15\linewidth}}}
        \scriptsize{Query} &\scriptsize{GT Mask} &\scriptsize{0-shot}  &\scriptsize{1-shot} &\scriptsize{4-shot}  \\
        \end{tabular}
    \end{minipage}
    \hspace{0.0001\linewidth} 
    \begin{minipage}{0.48\linewidth}
        \centering
         \includegraphics[width=\linewidth]{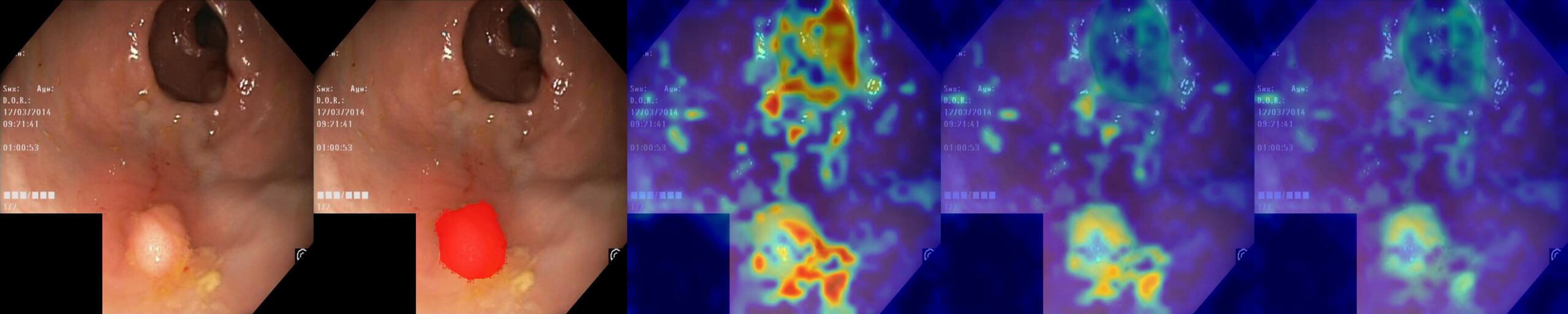}
         \includegraphics[width=\linewidth]{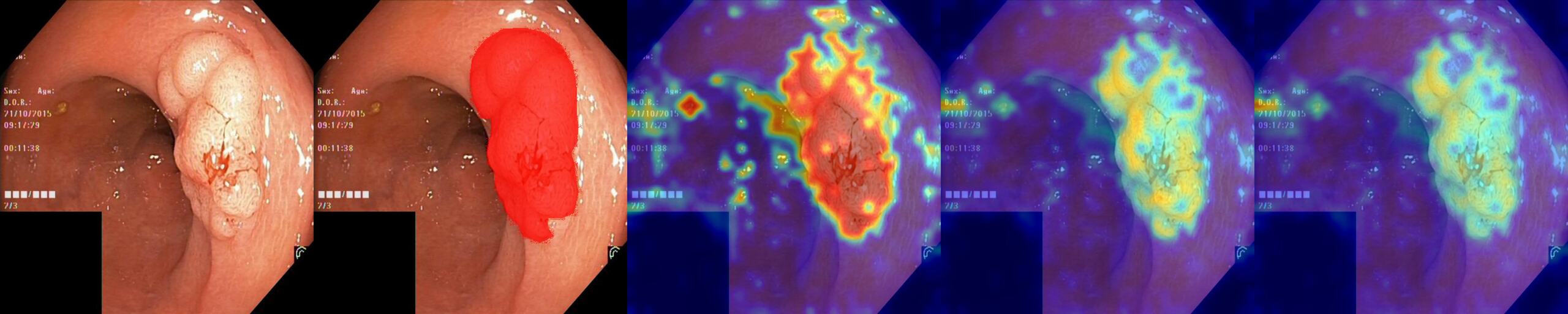}
        \vspace{-0.5cm} 
        
        \begin{tabular}{*{5}{>{\centering\arraybackslash}p{0.15\linewidth}}}
        \scriptsize{Query} &\scriptsize{GT Mask} &\scriptsize{0-shot}  &\scriptsize{1-shot} &\scriptsize{4-shot}  \\
        \end{tabular}
    \end{minipage}
 \vspace{-5pt}   
\caption{\small{Qualitative comparisons of our AdaptCLIP with different prompt numbers on \textbf{Kvasir}.}}\label{fig:kvasir}
\vspace{-10pt}
\end{figure*}

\begin{figure*}[t]
    \centering
    \begin{minipage}{0.48\linewidth}
        \centering
        \includegraphics[width=\linewidth]{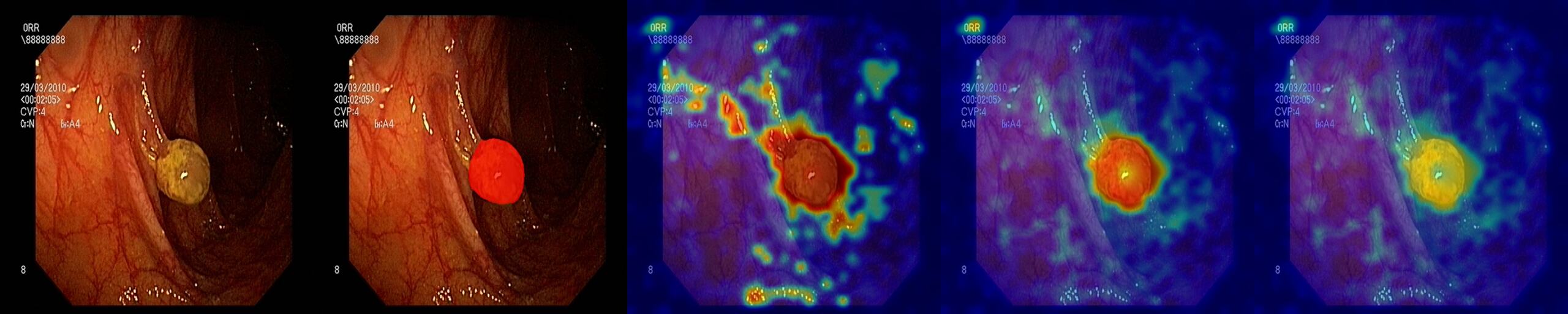}
        \includegraphics[width=\linewidth]{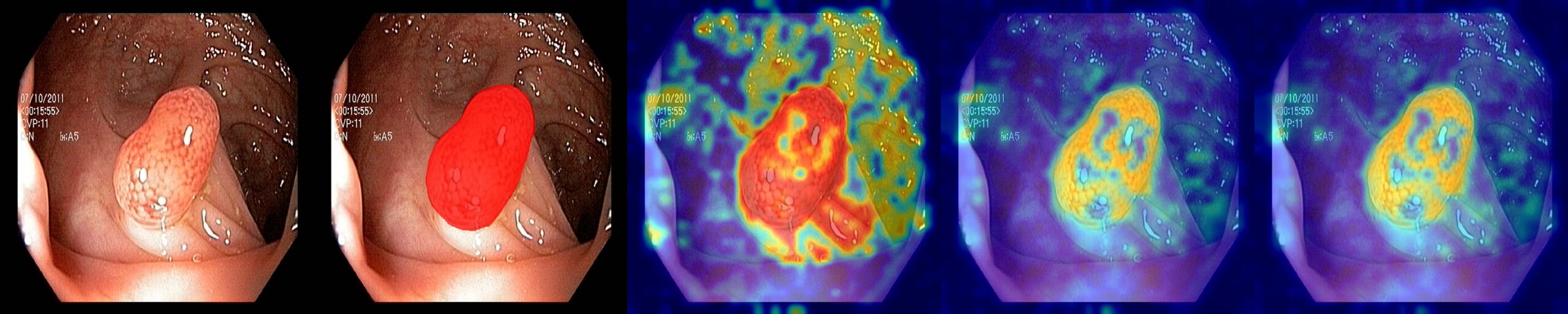}
        \vspace{-0.5cm} 
        
        \begin{tabular}{*{5}{>{\centering\arraybackslash}p{0.15\linewidth}}}
        \scriptsize{Query} &\scriptsize{GT Mask} &\scriptsize{0-shot}  &\scriptsize{1-shot} &\scriptsize{4-shot}  \\
        \end{tabular}
    \end{minipage}
    \hspace{0.0001\linewidth} 
    \begin{minipage}{0.48\linewidth}
        \centering
         \includegraphics[width=\linewidth]{./visualization/MedicalDataset_Endo_images_2f3249f3-47e5-4e2a-8273-6df59aa63d77.jpg_111.jpg}
         \includegraphics[width=\linewidth]{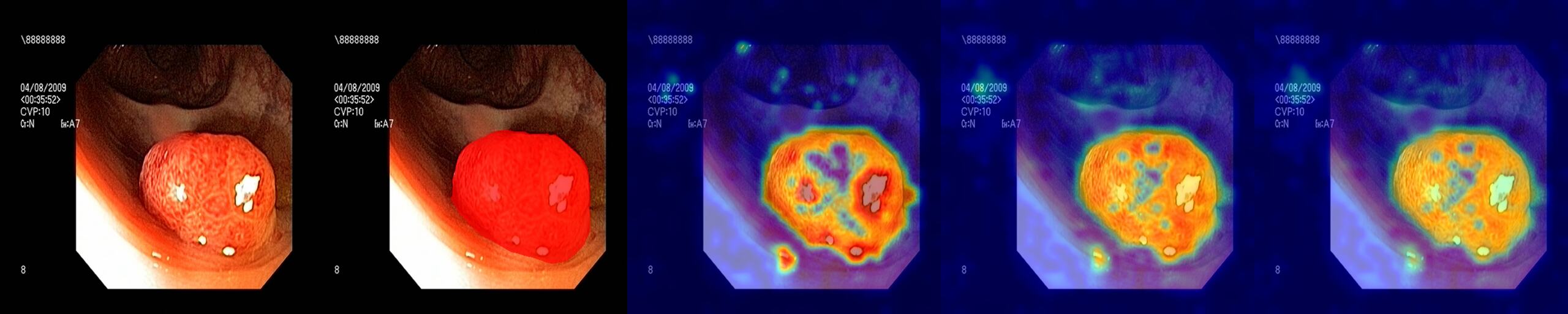}
        \vspace{-0.5cm} 
        
        \begin{tabular}{*{5}{>{\centering\arraybackslash}p{0.15\linewidth}}}
        \scriptsize{Query} &\scriptsize{GT Mask} &\scriptsize{0-shot}  &\scriptsize{1-shot} &\scriptsize{4-shot}  \\
        \end{tabular}
    \end{minipage}
 \vspace{-5pt}   
\caption{\small{Qualitative comparisons of our AdaptCLIP with different prompt numbers on \textbf{Endo}.}}\label{fig:endo}
\vspace{-10pt}
\end{figure*}

\begin{figure*}[t]
    \centering
    \vspace{-20pt}
    \begin{minipage}{0.48\linewidth}
        \centering
        \includegraphics[width=\linewidth]{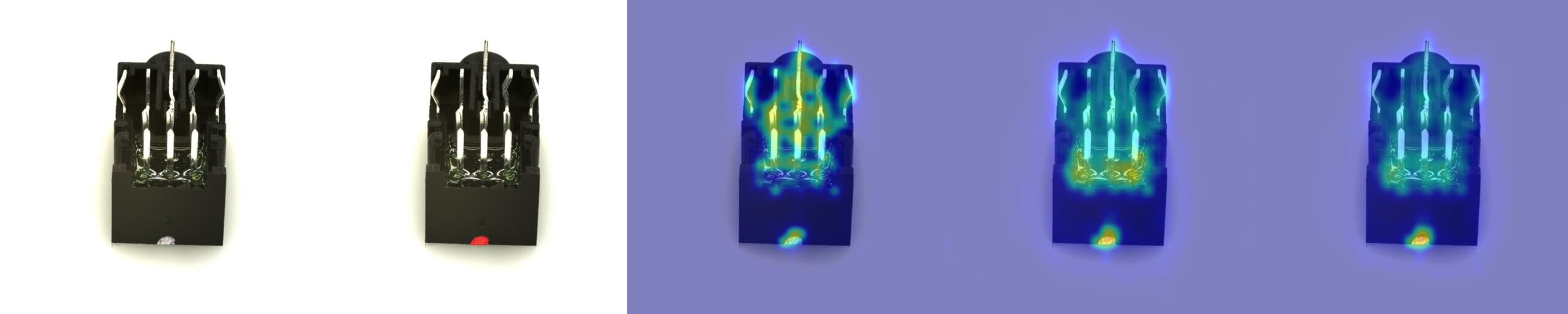}
        \vspace{-3.5pt}
        \includegraphics[width=\linewidth]{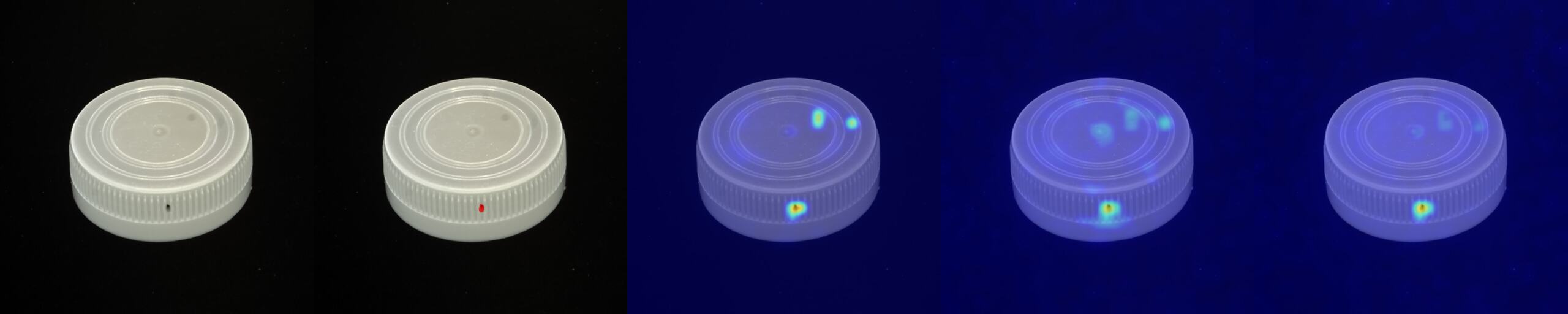}
        \vspace{-3.5pt}
        \includegraphics[width=\linewidth]{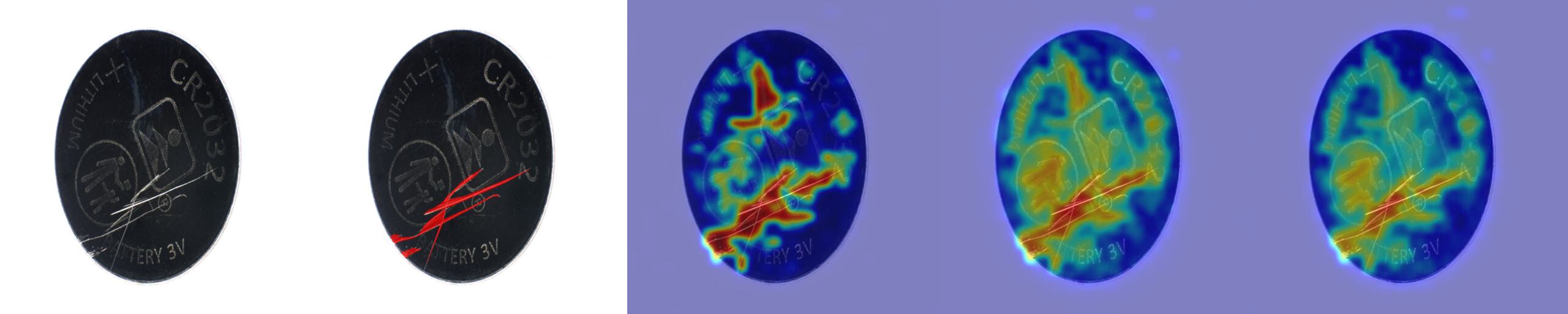}
        \vspace{-3.5pt}
        \includegraphics[width=\linewidth]{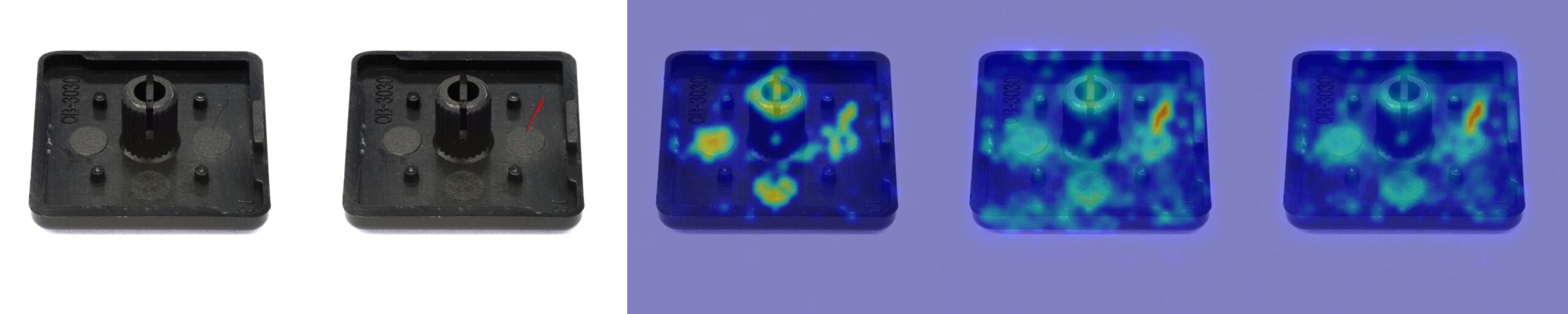}
        \vspace{-3.5pt}
        \includegraphics[width=\linewidth]{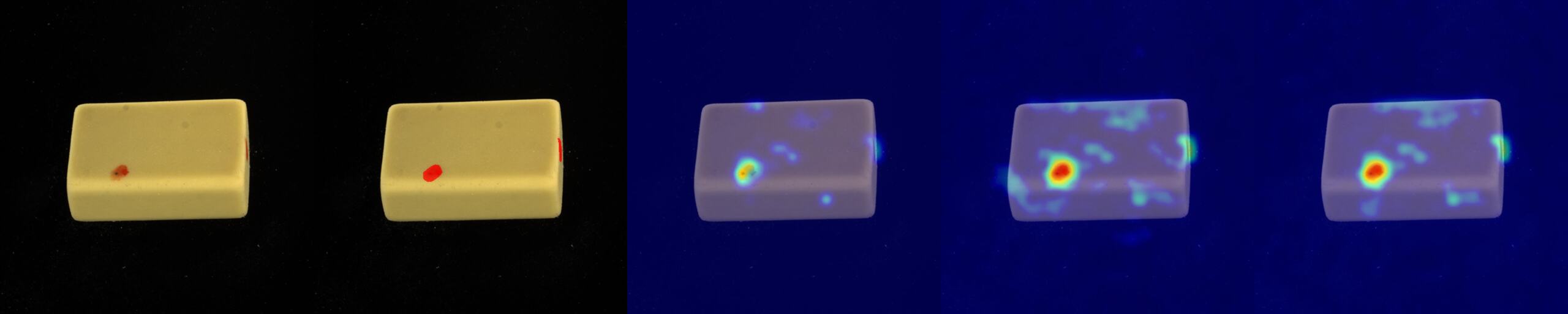}
        \vspace{-3.5pt}
        \includegraphics[width=\linewidth]{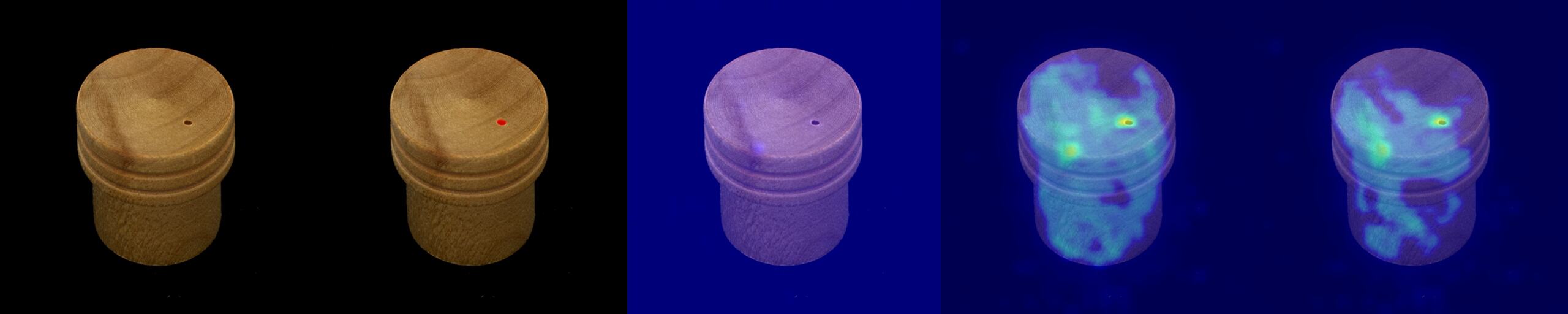}
        \vspace{-3.5pt}
        \includegraphics[width=\linewidth]{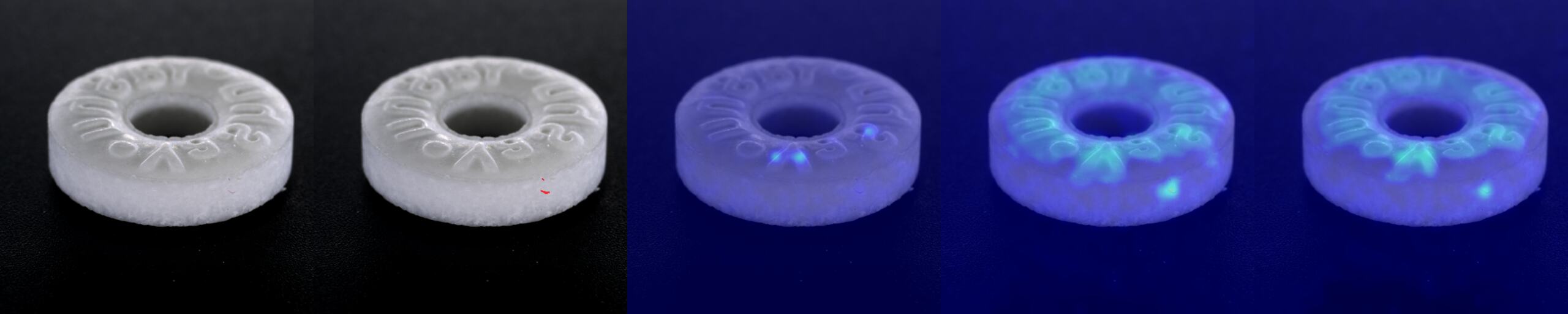}
        \vspace{-3.5pt}
        \includegraphics[width=\linewidth]{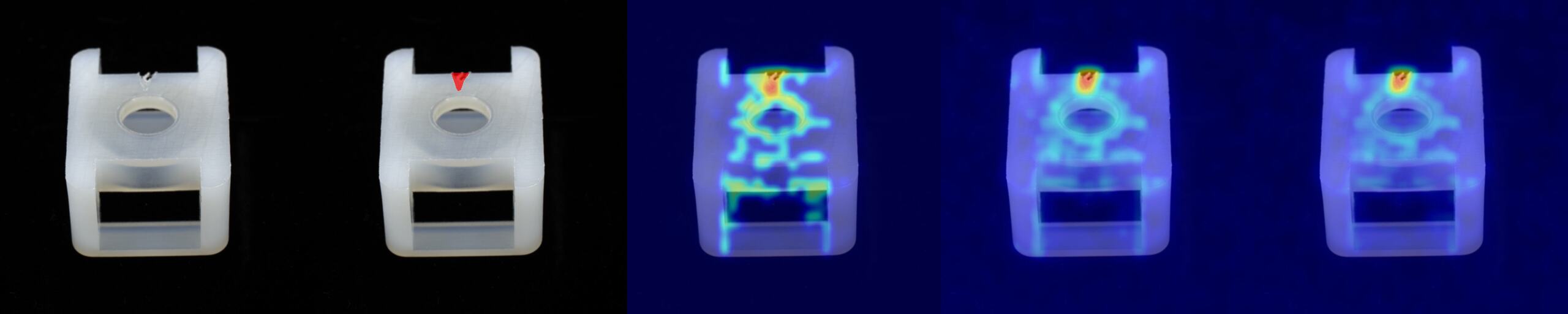}
        \vspace{-3.5pt}
        \includegraphics[width=\linewidth]{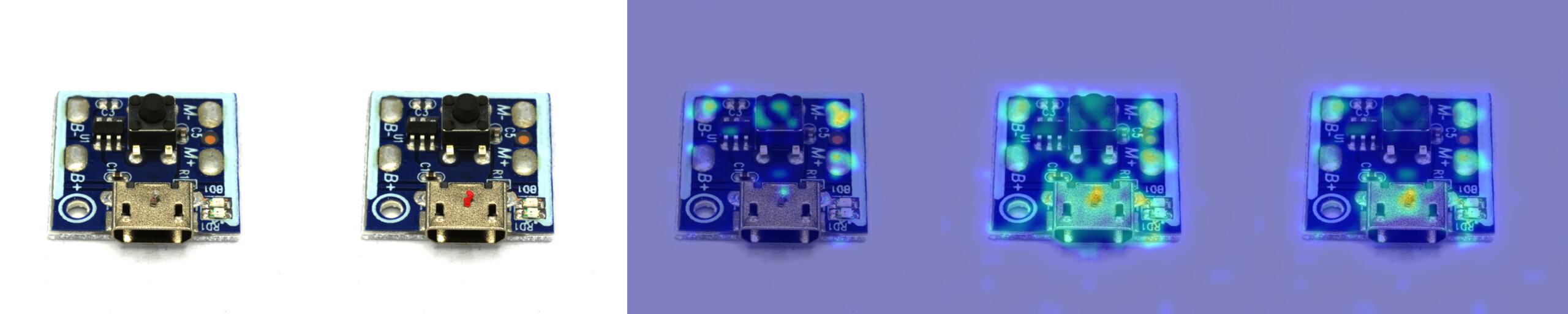}
        \vspace{-3.5pt}
        \includegraphics[width=\linewidth]{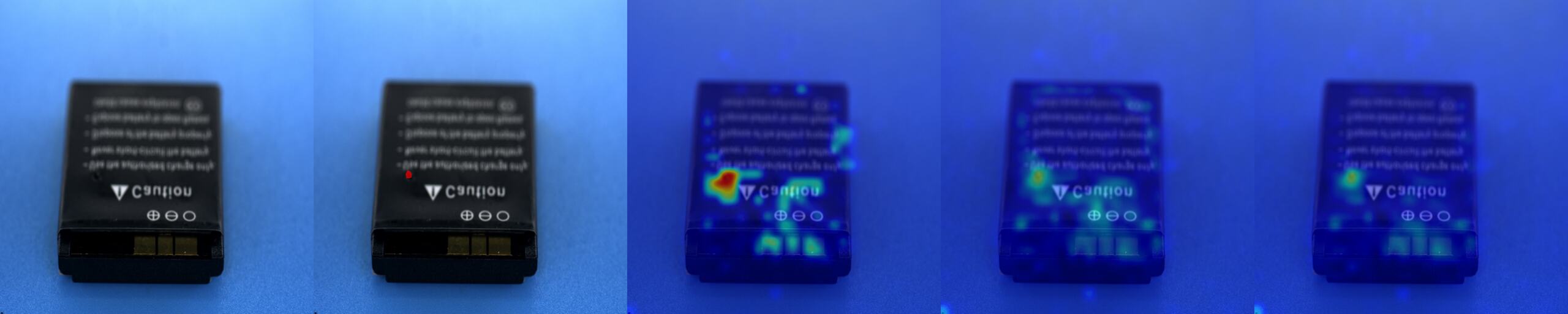}
        \vspace{-3.5pt}
        \includegraphics[width=\linewidth]{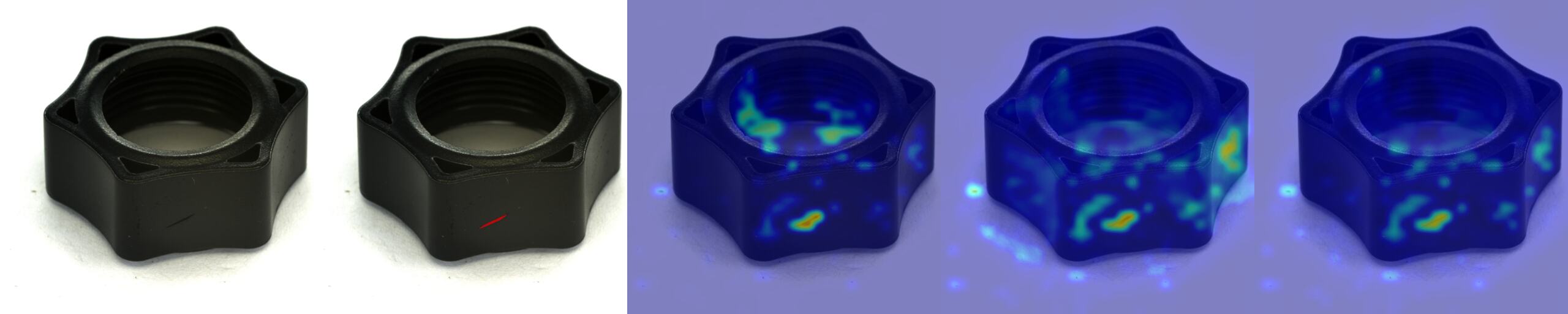}
        \vspace{-3.5pt}
        \includegraphics[width=\linewidth]{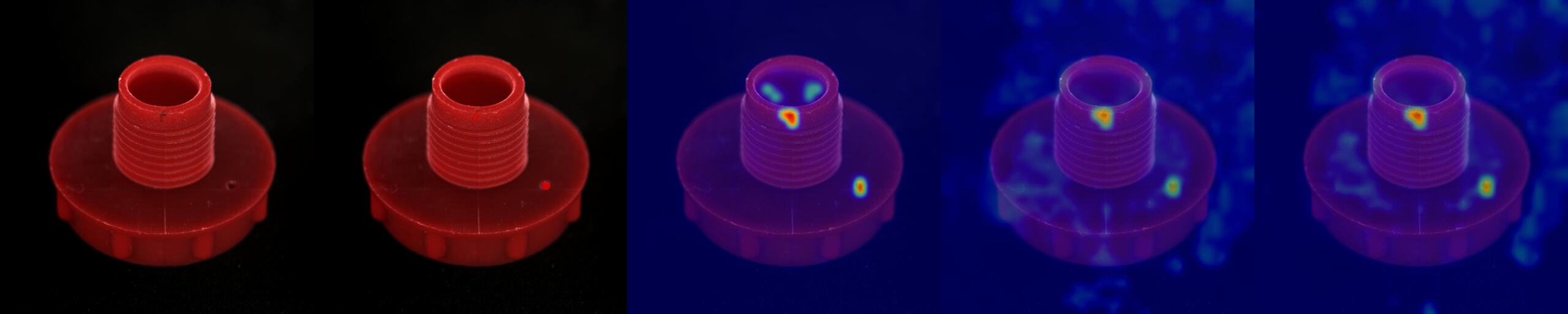}
        \vspace{-3.5pt}
        \includegraphics[width=\linewidth]{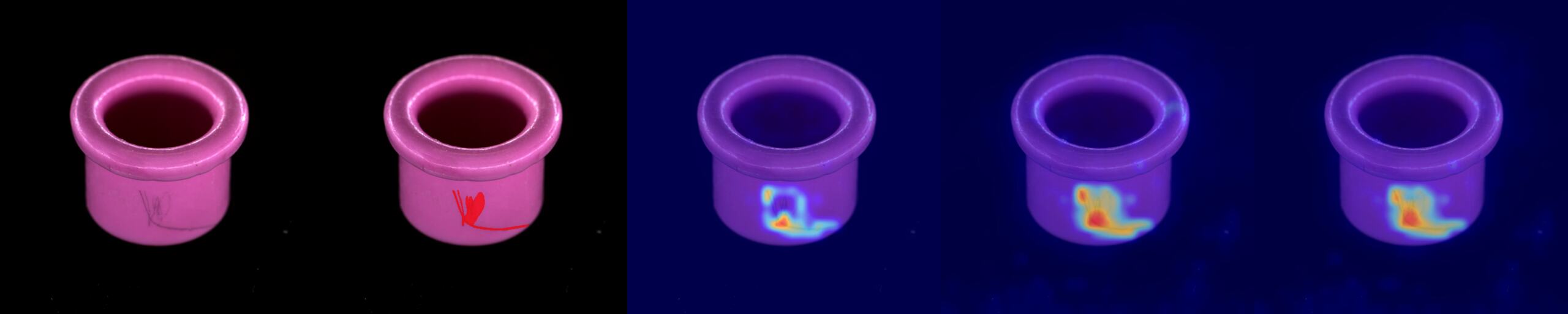}
        \vspace{-3.5pt}
        \includegraphics[width=\linewidth]{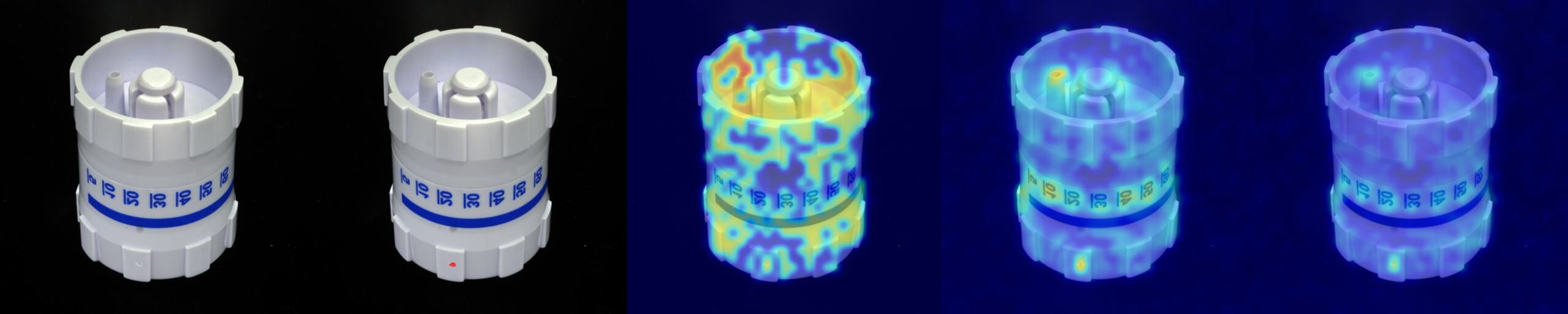}
        \vspace{-3.5pt}
        \includegraphics[width=\linewidth]{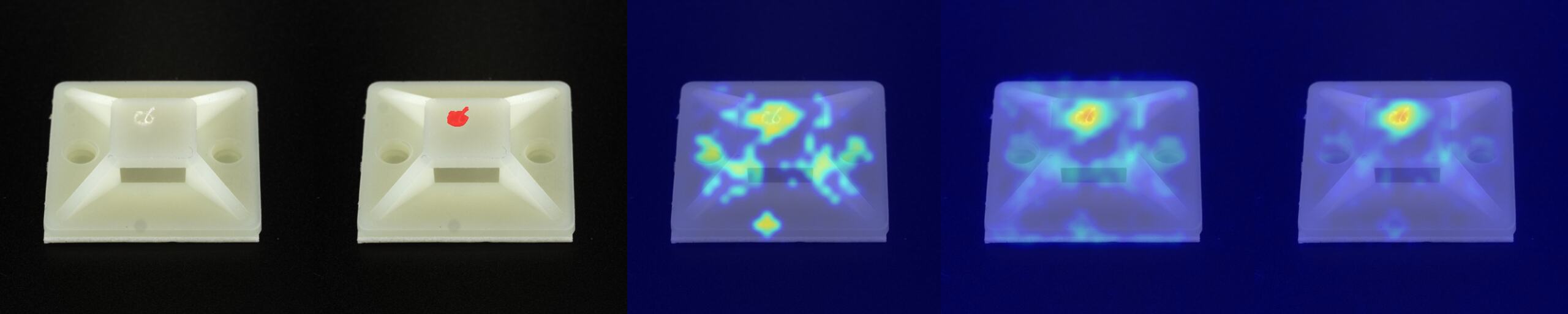}
        \vspace{-0.5cm} 
        
        \begin{tabular}{*{5}{>{\centering\arraybackslash}p{0.15\linewidth}}}
        \scriptsize{Query} &\scriptsize{GT Mask} &\scriptsize{0-shot}  &\scriptsize{1-shot} &\scriptsize{4-shot}  \\
        \end{tabular}
    \end{minipage}
    \hspace{0.0001\linewidth} 
    \begin{minipage}{0.48\linewidth}
        \centering
        \includegraphics[width=\linewidth]{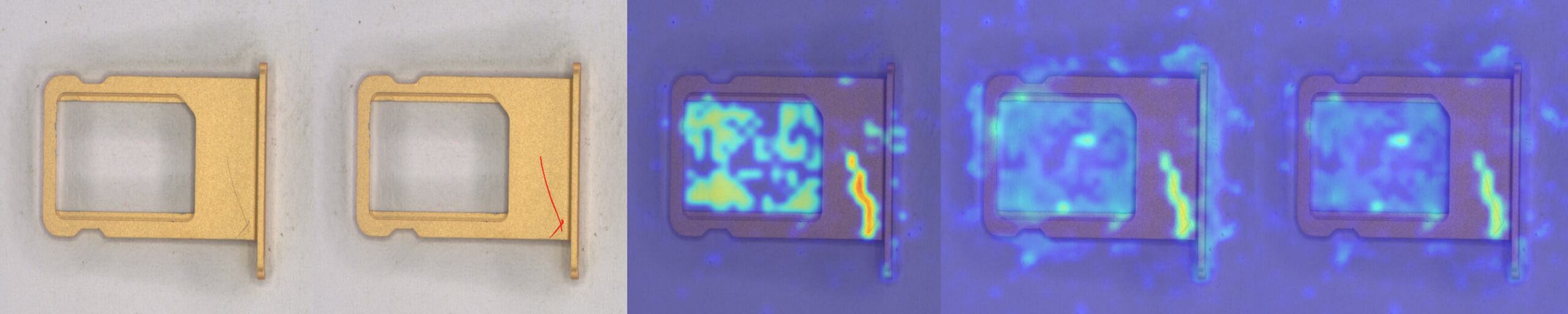}
        \vspace{-3.5pt}
        \includegraphics[width=\linewidth]{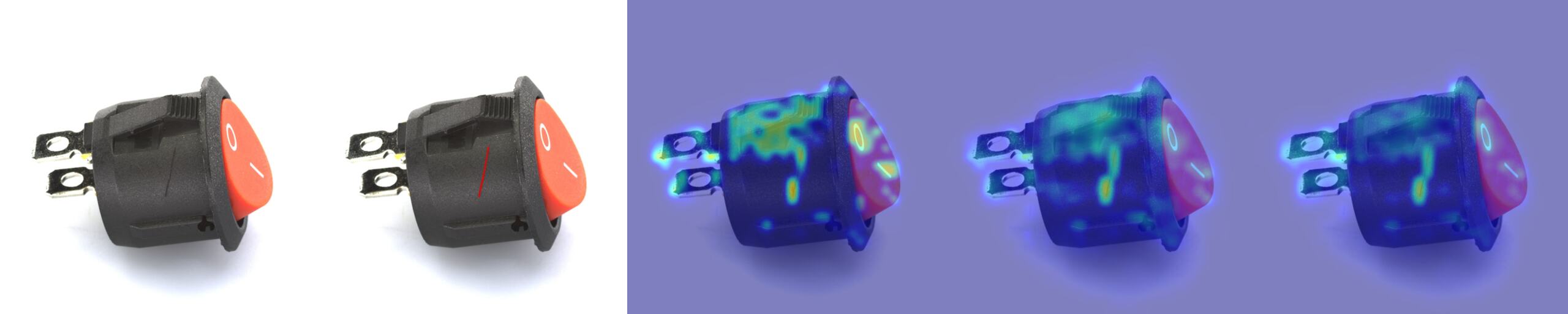}
        \vspace{-3.5pt}
        \includegraphics[width=\linewidth]{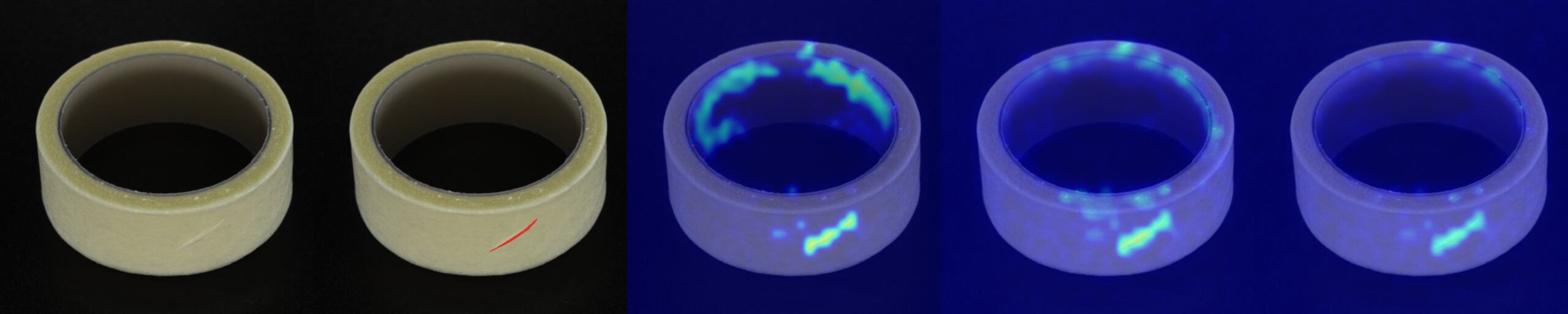}
        \vspace{-3.5pt}
        \includegraphics[width=\linewidth]{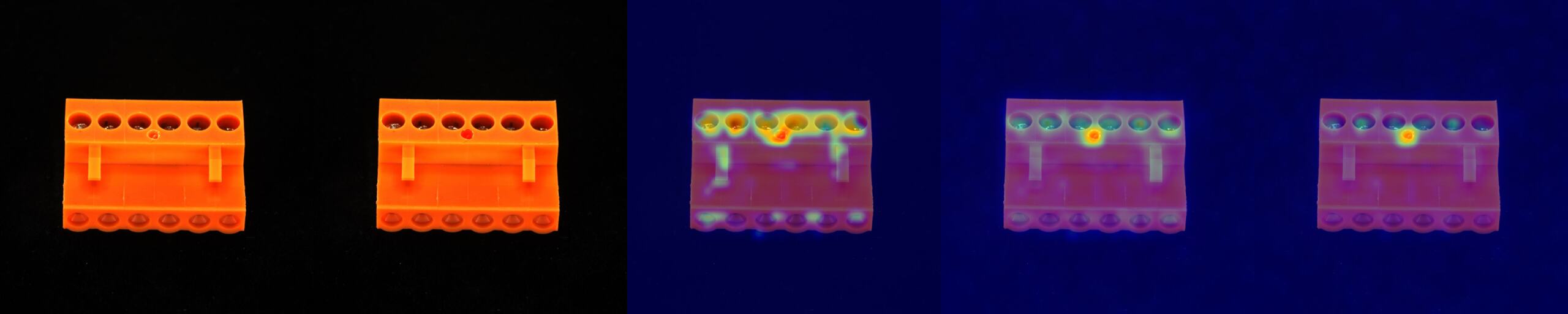}
        \vspace{-3.5pt}
        \includegraphics[width=\linewidth]{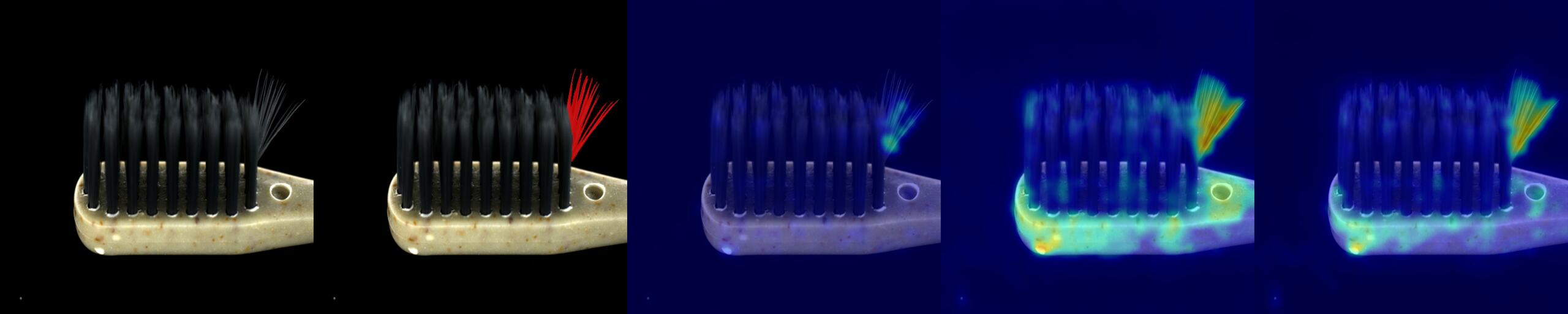}
        \vspace{-3.5pt}
        \includegraphics[width=\linewidth]{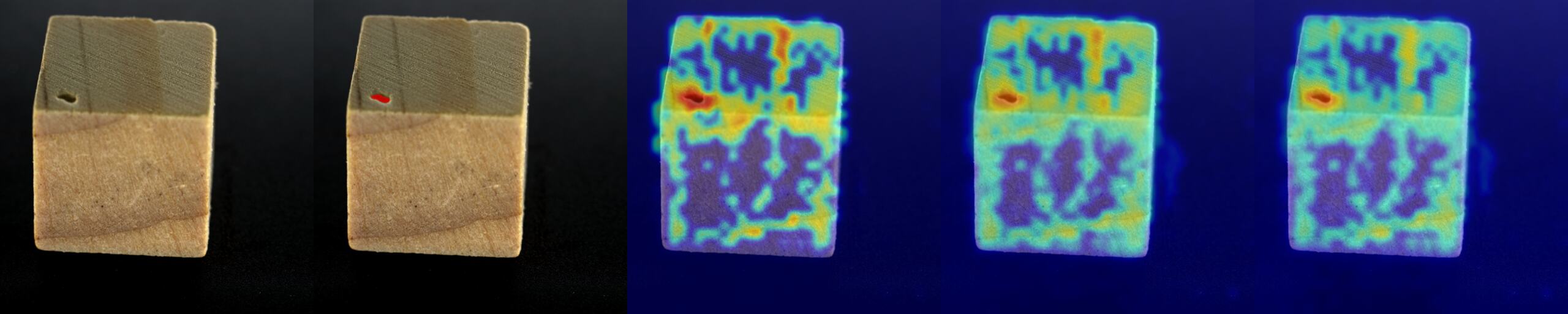}
        \vspace{-3.5pt}
        \includegraphics[width=\linewidth]{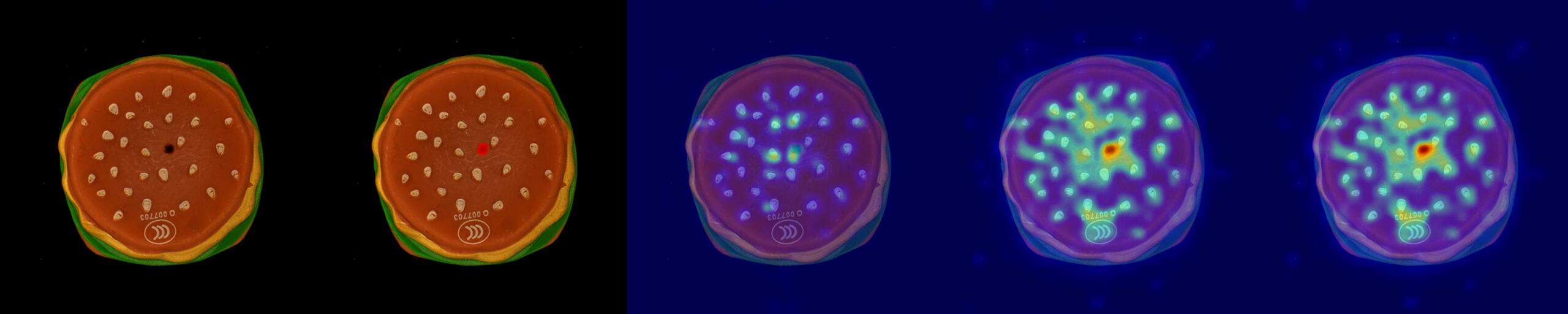}
        \vspace{-3.5pt}
        \includegraphics[width=\linewidth]{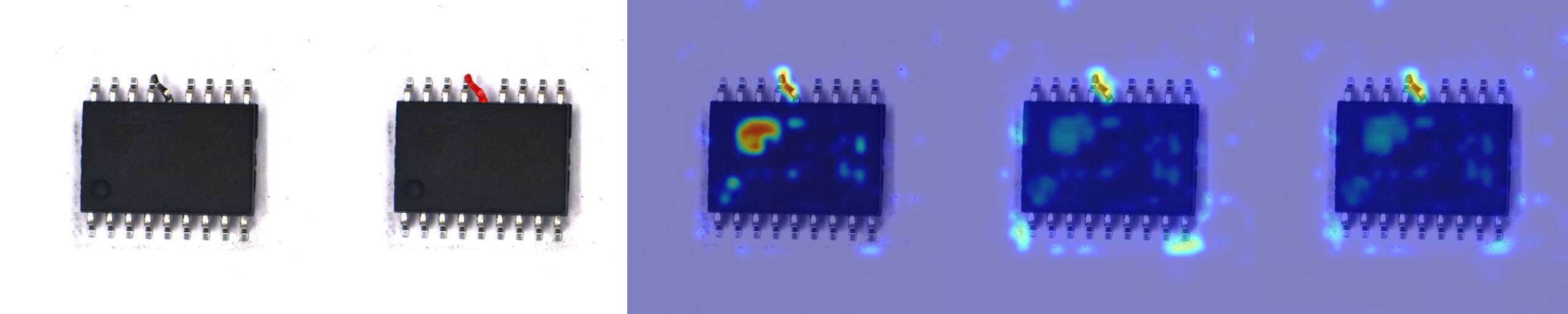}
        \vspace{-3.5pt}
        \includegraphics[width=\linewidth]{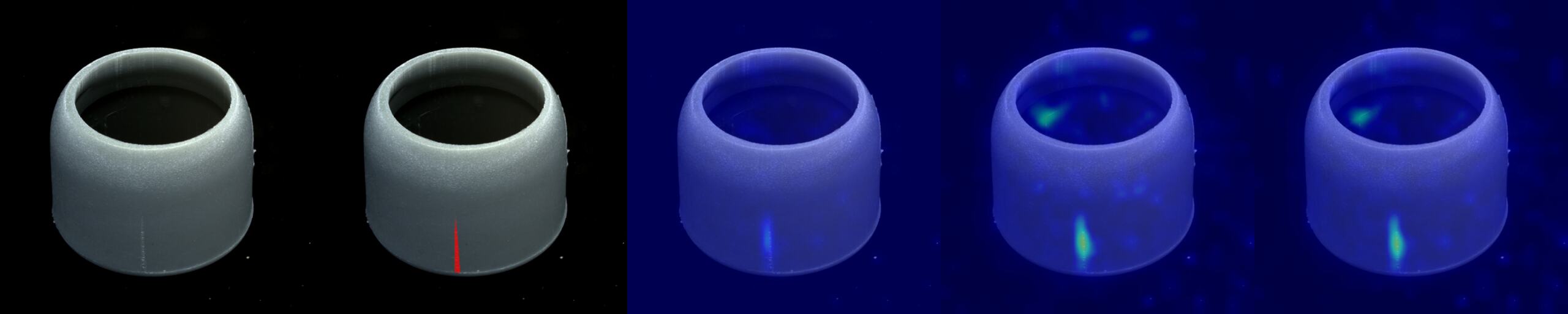}
        \vspace{-3.5pt}
        \includegraphics[width=\linewidth]{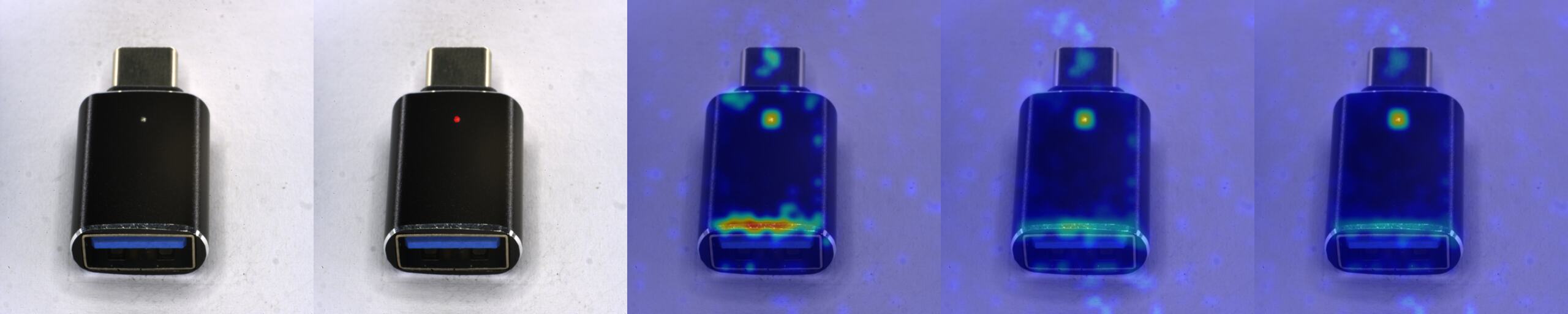}
        \vspace{-3.5pt}
        \includegraphics[width=\linewidth]{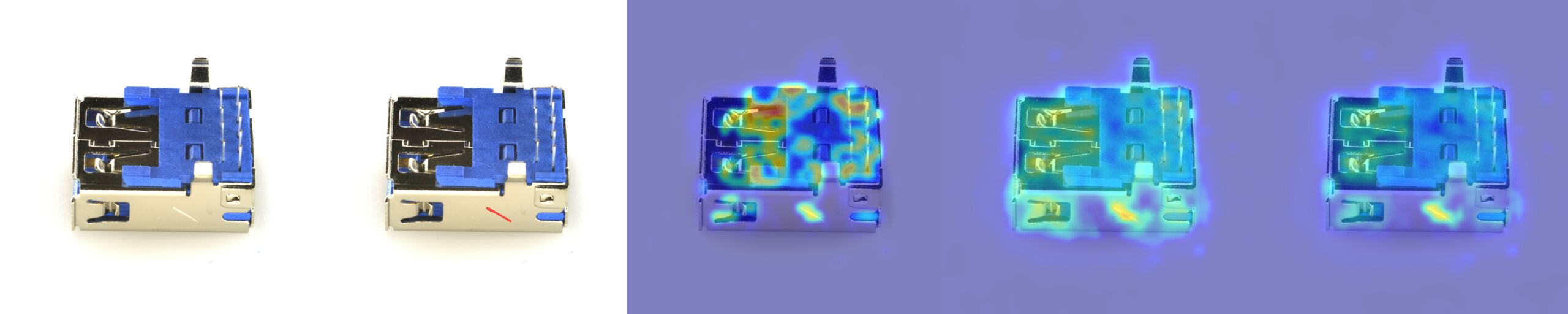}
        \vspace{-3.5pt}
        \includegraphics[width=\linewidth]{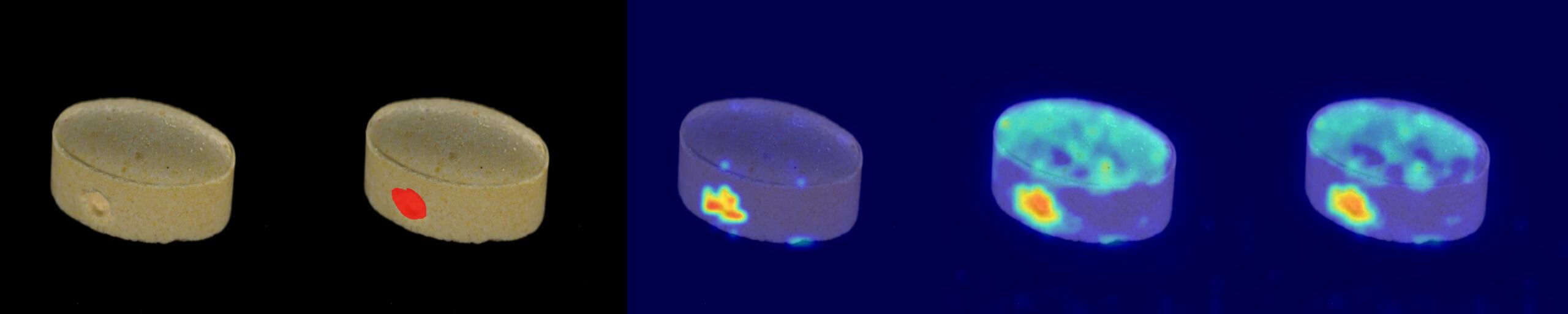}
        \vspace{-3.5pt}
        \includegraphics[width=\linewidth]{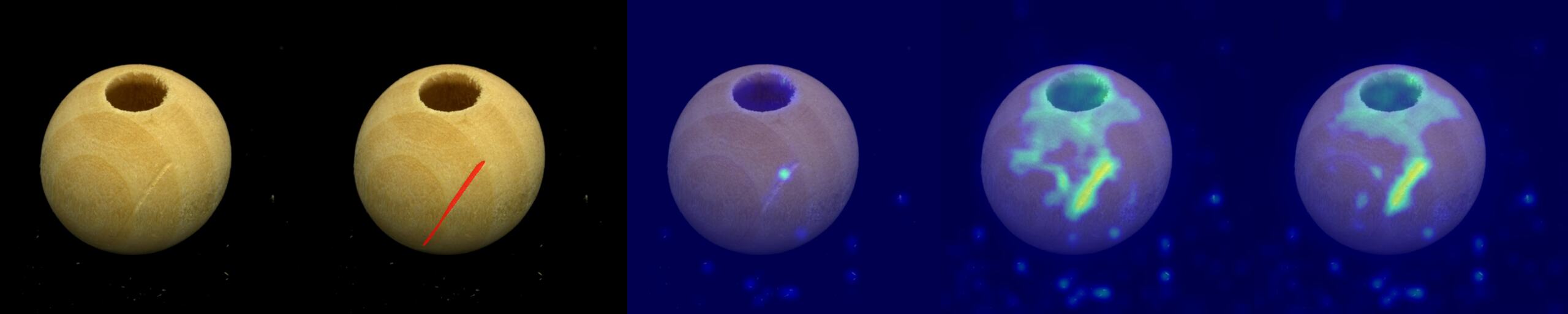}
        \vspace{-3.5pt}
        \includegraphics[width=\linewidth]{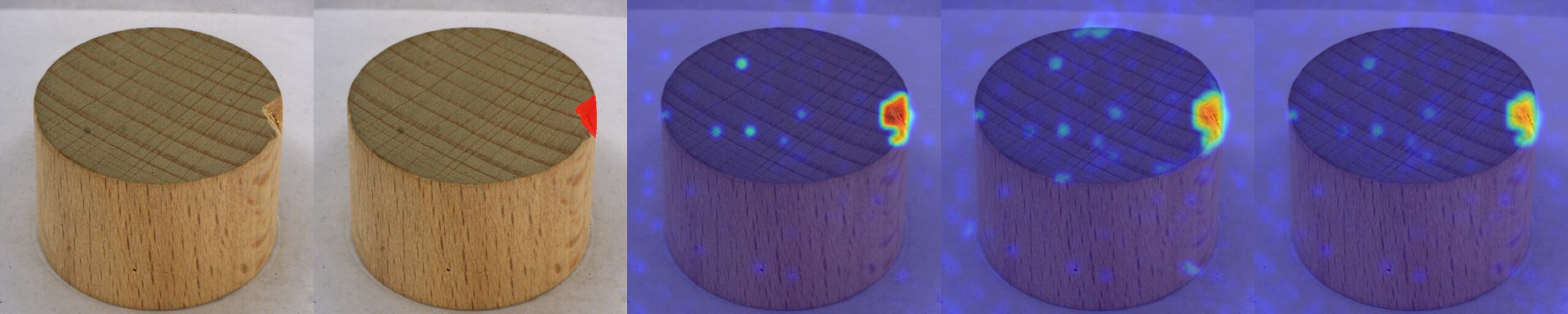}
        \vspace{-3.5pt}
        \includegraphics[width=\linewidth]{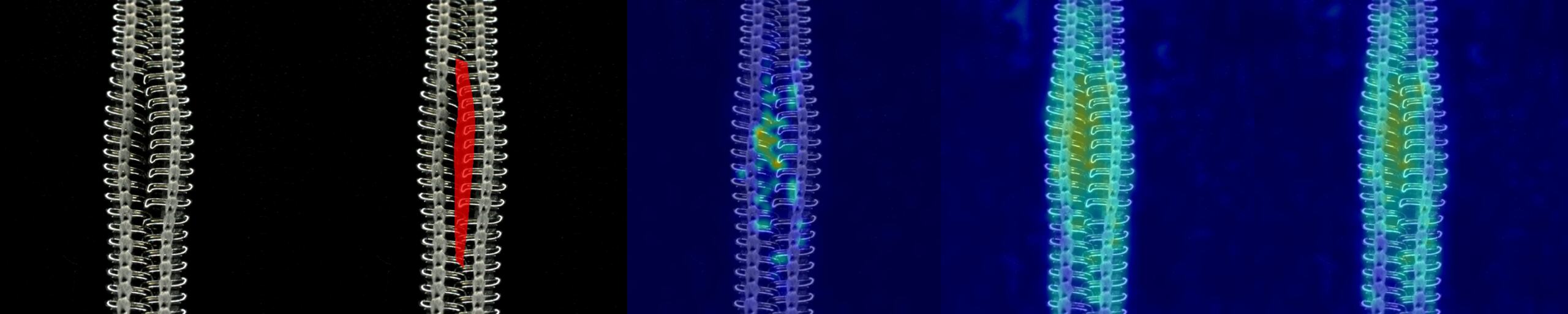}
        \vspace{-0.5cm} 
        
        \begin{tabular}{*{5}{>{\centering\arraybackslash}p{0.15\linewidth}}}
        \scriptsize{Query} &\scriptsize{GT Mask} &\scriptsize{0-shot}  &\scriptsize{1-shot} &\scriptsize{4-shot}  \\
        \end{tabular}
    \end{minipage}
 \vspace{-5pt}   
\caption{\small{Qualitative comparisons of our AdaptCLIP with different prompt numbers on \textbf{Real-IAD}.}}\label{fig:real-iad}
\vspace{-10pt}
\end{figure*}

In our main paper, we only visualize some selected examples from MVTec, VisA, Real-IAD, Kvasir and Endo to compare zero-shot and few-shot AdaptCLIP. Here, we show more visualizations for all 91 categories from 8 industrial and 2 medical datasets, as shown in Figs.~\ref{fig:mvtec},~\ref{fig:btad},~\ref{fig:mpdd},~\ref{fig:ksdd},~\ref{fig:visa},~\ref{fig:mvtec3d},~\ref{fig:dtd},~\ref{fig:kvasir},~\ref{fig:endo} and~\ref{fig:real-iad}.

\begin{table*}[t]
\setlength\tabcolsep{3pt}
\centering
\small
\caption{\small{Image-level anomaly classification comparisons with \textbf{AUPR} metric on industrial and medical domains.}}
\label{tab:sota_iaupr}
\vspace{-5pt}
\resizebox{1.0\textwidth}{!}{
\begin{tabular}{cc|cc cc cc cc |c| cc|c}
\toprule
  \multirow{2}{*}{\textbf{Shots}}  &\multirow{2}{*}{\textbf{Methods}}    &\multicolumn{9}{c|}{\textbf{Industrial}}    & \multicolumn{3}{c}{\textbf{Medical}}     \\ 
    \cmidrule{3-11}\cmidrule{12-14}
  &
  &{MVTec}	
  &{VisA}  	
  &{BTAD} 	
  &{MVTec3D}
  &{DTD} 
  &{KSDD}
  &{MPDD} 
  &{Real-IAD}	
  &{AVG}
  &{Br35H} 	&{Covid}
  &{AVG}\\
\midrule
\multirow{4}{*}{\textbf{0-shot}}		&WinCLIP~\cite{cvpr2023winclip}	&95.6	&78.7	&70.9	&89.6	&97.7	&84.9	&69.2	&62.9	&81.2	&82.2	&42.9	&62.6\\
	&AdaCLIP$^\dag$~\cite{eccv24adaclip} 	&95.6	&84.3	&95.5	&92.7	&96.4	&89.3	&70.8	&70.2	&86.9	&96.7	&43.6	&70.2\\
	&AnomalyCLIP~\cite{iclr2024anomalyclip} 	&96.4	&85.3	&88.2	&91.8	&97.2	&94.2	&82.5	&64.3	&\blue{87.5}	&94.2	&55.5	&\red{74.9}\\
\rowcolor[HTML]{EFEFEF}	&\textbf{AdaptCLIP}		&96.7	&87.6	&92.2	&93.4	&98.4	&95.7	&74.5	&70.8	&\red{88.8}	&95.1	&54.4	&\blue{74.8}\\
\midrule
\multirow{4}{*}{\textbf{1-shot}}		&WinCLIP+~\cite{cvpr2023winclip}	&\pmerror{96.8}{0.2}	&\pmerror{81.7}{1.5}	&\pmerror{80.5}{3.3}	&\pmerror{91.5}{0.2}	&\pmerror{99.0}{0.1}	&\pmerror{84.6}{0.9}	&\pmerror{73.5}{2.2}	&\pmerror{71.2}{0.3}	&84.9	&\pmerror{79.4}{1.9}	&\pmerror{76.9}{6.4}	&\blue{78.2}\\
	&InCtrl~\cite{cvpr2024inctrl} 	&\pmerror{95.2}{0.3}	&\pmerror{84.1}{1.5}	&\pmerror{83.4}{9.3}	&\pmerror{91.9}{0.7}	&\pmerror{98.9}{0.2}	&\pmerror{81.5}{2.2}	&\pmerror{75.7}{0.8}	&\pmerror{69.9}{0.0}	&85.1	&\pmerror{82.6}{8.7}	&\pmerror{65.1}{5.6}	&73.9\\
\bottomrule
\end{tabular}
}
\vspace{-5pt}
\end{table*}

\begin{table*}[t]
\setlength\tabcolsep{3pt}
\centering
\small
\caption{\small{Image-level anomaly classification comparisons with \textbf{F1$_\text{max}$} metric on industrial and medical domains.}}
\label{tab:sota_if1max}
\vspace{-5pt}
\resizebox{1.0\textwidth}{!}{
\begin{tabular}{cc|cc cc cc cc |c| cc|c}
\toprule
  \multirow{2}{*}{\textbf{Shots}}  &\multirow{2}{*}{\textbf{Methods}}    &\multicolumn{9}{c|}{\textbf{Industrial}}    & \multicolumn{3}{c}{\textbf{Medical}}     \\ 
    \cmidrule{3-11}\cmidrule{12-14}
    &
  &{MVTec}	
  &{VisA}  	
  &{BTAD} 	
  &{MVTec3D}
  &{DTD} 
  &{KSDD}
  &{MPDD} 
  &{Real-IAD}	
  &{AVG}
 &{Br35H} 	&{Covid}
  &{AVG}\\
\midrule
\multirow{4}{*}{\textbf{0-shot}}		&WinCLIP~\cite{cvpr2023winclip}	&92.7	&78.2	&67.8	&89.7	&94.1	&80.8	&77.5	&65.3	&80.8	&74.1	&42.7	&58.4\\
	&AdaCLIP$^\dag$~\cite{eccv24adaclip} 	&92.4	&80.0	&90.2	&89.6	&92.6	&85.4	&76.7	&67.8	&84.3	&92.3	&42.9	&67.6\\
	&AnomalyCLIP~\cite{iclr2024anomalyclip} 	&92.7	&80.4	&83.8	&88.8	&93.6	&89.7	&80.4	&65.9	&\blue{84.4}	&86.8	&54.2	&\blue{70.5}\\
\rowcolor[HTML]{EFEFEF}	&\textbf{AdaptCLIP}		&93.7	&83.0	&89.5	&89.3	&95.1	&92.3	&79.7	&68.9	&\red{86.4}	&87.7	&55.4	&\red{71.6}\\
    \midrule
\multirow{4}{*}{\textbf{1-shot}}		&WinCLIP+~\cite{cvpr2023winclip}	&\pmerror{94.0}{0.4}	&\pmerror{81.3}{0.8}	&\pmerror{77.1}{3.5}	&\pmerror{90.3}{0.2}	&\pmerror{96.7}{0.3}	&\pmerror{80.1}{1.6}	&\pmerror{81.5}{0.9}	&\pmerror{69.5}{0.1}	&83.8	&\pmerror{75.2}{1.8}	&\pmerror{71.7}{5.5}	&\blue{73.5}\\
	&InCtrl~\cite{cvpr2024inctrl} 	&\pmerror{93.9}{0.4}	&\pmerror{83.1}{1.4}	&\pmerror{81.5}{5.6}	&\pmerror{89.8}{0.3}	&\pmerror{97.7}{0.1}	&\pmerror{80.3}{1.2}	&\pmerror{80.9}{2.2}	&\pmerror{70.8}{0.0}	&84.8	&\pmerror{78.1}{2.7}	&\pmerror{65.4}{3.0}	&71.8\\
\bottomrule
\end{tabular}
}
\vspace{-5pt}
\end{table*}

\begin{table*}[t]
\setlength\tabcolsep{3pt}
\centering
\small
\caption{\small{Pixel-level anomaly segmentation comparisons with \textbf{AUROC} metric on industrial and medical domains.}}
\label{tab:sota_pauroc}
\vspace{-5pt}
\resizebox{1.0\textwidth}{!}{
\begin{tabular}{cc|cc cc cc cc |c| cc|c}
\toprule
  \multirow{2}{*}{\textbf{Shots}}  &\multirow{2}{*}{\textbf{Methods}}    &\multicolumn{9}{c|}{\textbf{Industrial}}    & \multicolumn{3}{c}{\textbf{Medical}}     \\ 
    \cmidrule{3-11}\cmidrule{12-14}
    &
  &{MVTec}	
  &{VisA}  	
  &{BTAD} 	
  &{MVTec3D}
  &{DTD} 
  &{KSDD}
  &{MPDD} 
  &{Real-IAD}	
  &{AVG}
  &{Kvasir} 	
  &{Endo} 
  &{AVG}\\
\midrule
\multirow{4}{*}{\textbf{0-shot}}		&WinCLIP~\cite{cvpr2023winclip}	&82.3	&73.2	&72.7	&91.2	&79.5	&93.0	&71.2	&84.5	&81.0	&69.7	&68.2	&69.0\\
	&AdaCLIP$^\dag$~\cite{eccv24adaclip} 	&88.3	&95.7	&91.6	&97.1	&98.3	&97.6	&95.5	&96.1	&95.0	&77.8	&83.8	&80.8\\
	&AnomalyCLIP~\cite{iclr2024anomalyclip} 	&91.1	&95.5	&94.2	&96.2	&97.9	&98.1	&96.5	&95.1	&\red{95.6}	&79.0	&84.2	&\blue{81.6}\\
	\rowcolor[HTML]{EFEFEF}	&\textbf{AdaptCLIP}	&90.9	&95.7	&93.8	&97.2	&97.7	&98.1	&95.9	&94.9	&\blue{95.5}	&82.1	&86.5	&\red{84.3}\\
        \midrule
\multirow{4}{*}{\textbf{1-shot}}		&WinCLIP+~\cite{cvpr2023winclip}	&\pmerror{93.4}{0.2}	&\pmerror{94.7}{0.1}	&\pmerror{95.6}{0.2}	&\pmerror{96.8}{0.2}	&\pmerror{96.5}{0.1}	&\pmerror{97.6}{0.1}	&\pmerror{94.7}{1.3}	&\pmerror{95.0}{0.0}	&95.5	&\pmerror{73.1}{2.6}	&\pmerror{72.3}{0.3}	&72.7\\
	&InCtrl~\cite{cvpr2024inctrl} 	&\pmerror{94.6}{0.2}	&\pmerror{89.0}{0.2}	&\pmerror{96.6}{0.1}	&\pmerror{94.4}{0.2}	&\pmerror{98.6}{0.1}	&\pmerror{97.8}{0.2}	&\pmerror{94.4}{1.0}	&\pmerror{95.4}{0.0}	&95.1	&\pmerror{65.8}{1.3}	&\pmerror{66.5}{6.2}	&66.2\\
\bottomrule
\end{tabular}
}
\vspace{-5pt}
\end{table*}

\begin{table*}[t]
\setlength\tabcolsep{3pt}
\centering
\small
\caption{\small{Pixel-level anomaly segmentation comparisons with \textbf{F1$_\text{max}$} metric on industrial and medical domains.}}
\label{tab:sota_pf1max}
\vspace{-5pt}
\resizebox{1.0\textwidth}{!}{
\begin{tabular}{cc|cc cc cc cc |c| cc|c}
\toprule
  \multirow{2}{*}{\textbf{Shots}}  &\multirow{2}{*}{\textbf{Methods}}    &\multicolumn{9}{c|}{\textbf{Industrial}}    & \multicolumn{3}{c}{\textbf{Medical}}     \\ 
    \cmidrule{3-11}\cmidrule{12-14}
    &
  &{MVTec}	
  &{VisA}  	
  &{BTAD} 	
  &{MVTec3D}
  &{DTD} 
  &{KSDD}
  &{MPDD} 
  &{Real-IAD}	
  &{AVG}
  &{Kvasir} 	
  &{Endo} 
  &{AVG}\\
\midrule
\multirow{4}{*}{\textbf{0-shot}}		&WinCLIP~\cite{cvpr2023winclip}	&24.8	&8.8	&18.3	&10.0	&15.7	&15.5	&15.3	&7.7	&14.5	&35.7	&32.9	&34.3\\
	&AdaCLIP$^\dag$~\cite{eccv24adaclip} 	&42.1	&36.9	&46.6	&42.2	&70.7	&38.7	&29.5	&35.4	&\blue{42.8}	&44.2	&48.7	&46.5\\
	&AnomalyCLIP~\cite{iclr2024anomalyclip} 	&38.0	&28.1	&49.4	&36.0	&62.2	&56.5	&34.0	&34.5	&42.3	&46.1	&50.3	&\blue{48.2}\\
	\rowcolor[HTML]{EFEFEF}	&\textbf{AdaptCLIP}	&43.2	&32.6	&46.3	&36.1	&63.6	&58.1	&29.9	&35.6	&\red{43.2}	&50.6	&54.5	&\red{52.6}\\
    \midrule
\multirow{4}{*}{\textbf{1-shot}}		&WinCLIP+~\cite{cvpr2023winclip}	&\pmerror{42.0}{0.8}	&\pmerror{22.9}{0.2}	&\pmerror{46.8}{2.3}	&\pmerror{24.8}{0.7}	&\pmerror{50.7}{0.7}	&\pmerror{29.2}{0.2}	&\pmerror{31.1}{2.2}	&\pmerror{21.8}{0.2}	&33.7	&\pmerror{39.0}{1.9}	&\pmerror{36.8}{0.3}	&37.9\\
	&InCtrl~\cite{cvpr2024inctrl} 	&\pmerror{51.0}{1.6}	&\pmerror{25.2}{0.4}	&\pmerror{49.9}{1.3}	&\pmerror{25.3}{0.7}	&\pmerror{62.9}{0.5}	&\pmerror{38.0}{0.6}	&\pmerror{29.9}{1.7}	&\pmerror{27.0}{0.0}	&38.7	&\pmerror{33.2}{1.1}	&\pmerror{33.0}{3.7}	&33.1\\
\bottomrule
\end{tabular}
}
\vspace{-5pt}
\end{table*}

\begin{table*}
  \setlength\tabcolsep{21pt}
  \centering
  \small
  \caption{\small{Image-level anomaly classification and pixel-level anomaly segmentation results on \textbf{MVTec} with zero-shot and few-shot AdaptCLIP.}}\label{tab:mvtec}
  \vspace{-5pt}
  \resizebox{1.0\textwidth}{!}{
    \begin{tabular}{cc  ccc  c ccc}
    \toprule
     \multirow{2}{*}{\textbf{Shot}}  &\multirow{2}{*}{\textbf{Catergoies}} &\multicolumn{3}{c}{\textbf{Anomaly Classification}} & &\multicolumn{3}{c}{\textbf{Anomaly Segmentation}} \\
     \cline{3-5} \cline{7-9}\\
       & &\textbf{I-AUROC}	&\textbf{I-AUPR}& \textbf{I-$\text{F1}_{\text{max}}$}	&	&\textbf{P-AUROC}	&\textbf{P-AUPR}	&\textbf{P-$\text{F1}_{\text{max}}$}	\\
    \midrule
     \multirow{16}{*}{\textbf{0-shot}}
        & carpet & 100.0 & 99.9 & 100.0 && 99.2 & 68.2 & 67.2 \\
        & bottle & 90.7 & 97.1 & 90.8 && 92.4 & 60.3 & 56.4 \\
        & hazelnut & 94.8 & 97.0 & 92.4 && 97.6 & 51.0 & 53.7 \\
        & leather & 100.0 & 99.9 & 99.5 && 99.1 & 27.9 & 43.7 \\
        & cable & 83.6 & 89.9 & 82.5 && 76.6 & 15.6 & 24.7 \\
\bottomrule
  \end{tabular}
  }
  \vspace{-10pt}
\end{table*}

\begin{table*}
  \setlength\tabcolsep{15pt}
  \centering
  \small
  \caption{\small{Image-level anomaly classification and pixel-level anomaly segmentation results on \textbf{VisA} with zero-shot and few-shot AdaptCLIP.}}\label{tab:visa}
  \vspace{-5pt}
  \resizebox{1.0\textwidth}{!}{
    \begin{tabular}{cc  ccc  c ccc}
    \toprule
    \multirow{2}{*}{\textbf{Shot}}  &\multirow{2}{*}{\textbf{Catergoies}} &\multicolumn{3}{c}{\textbf{Anomaly Classification}} & &\multicolumn{3}{c}{\textbf{Anomaly Segmentation}} \\
     \cline{3-5} \cline{7-9}\\
       & &\textbf{I-AUROC}	&\textbf{I-AUPR}& \textbf{I-$\text{F1}_{\text{max}}$}	&	&\textbf{P-AUROC}	&\textbf{P-AUPR}	&\textbf{P-$\text{F1}_{\text{max}}$}	\\
    \midrule
     \multirow{13}{*}{\textbf{0-shot}}
        & candle & 87.4 & 90.5 & 81.4 && 98.7 & 21.6 & 35.2 \\
        & capsules & 93.9 & 96.8 & 91.6 && 94.2 & 33.4 & 44.1 \\
        & cashew & 86.1 & 94.0 & 84.0 && 93.4 & 22.1 & 27.2 \\
        & chewinggum & 96.7 & 98.7 & 96.4 && 99.5 & 82.7 & 77.8 \\
        & fryum & 91.5 & 95.9 & 88.5 && 95.3 & 26.5 & 33.6 \\
\bottomrule
  \end{tabular}
  }
  \vspace{-10pt}
\end{table*}

\begin{table*}
  \setlength\tabcolsep{12pt}
  \centering
  \small
  \caption{\small{Image-level anomaly classification and pixel-level anomaly segmentation results on \textbf{MVTec 3D} with zero-shot and few-shot AdaptCLIP.}}\label{tab:mvtec3d}
  \vspace{-5pt}
  \resizebox{1.0\textwidth}{!}{
    \begin{tabular}{cc  ccc  c ccc}
    \toprule
    \multirow{2}{*}{\textbf{Shot}}  &\multirow{2}{*}{\textbf{Catergoies}} &\multicolumn{3}{c}{\textbf{Anomaly Classification}} & &\multicolumn{3}{c}{\textbf{Anomaly Segmentation}} \\
     \cline{3-5} \cline{7-9}\\
      & &\textbf{I-AUROC}	&\textbf{I-AUPR}& \textbf{I-$\text{F1}_{\text{max}}$}	&	&\textbf{P-AUROC}	&\textbf{P-AUPR}	&\textbf{P-$\text{F1}_{\text{max}}$}	\\
    \midrule
     \multirow{11}{*}{\textbf{0-shot}}
        & bagel & 93.4 & 98.4 & 92.7 && 99.6 & 73.1 & 67.2 \\
        & cable\_gland & 73.6 & 92.4 & 90.2 && 98.2 & 30.1 & 33.6 \\
        & carrot & 78.7 & 94.7 & 91.3 && 98.1 & 24.5 & 31.4 \\
        & cookie & 72.7 & 92.0 & 88.0 && 96.3 & 42.9 & 44.2 \\
        & tire & 78.9 & 93.7 & 87.4 && 97.9 & 29.7 & 36.7 \\
\bottomrule
  \end{tabular}
  }
  \vspace{-10pt}
\end{table*}

\begin{table*}
  \setlength\tabcolsep{15pt}
  \centering
  \small
  \caption{\small{Image-level anomaly classification and pixel-level anomaly segmentation results on \textbf{DTD} with zero-shot and few-shot AdaptCLIP.}}\label{tab:dtd}
  \vspace{-5pt}
  \resizebox{1.0\textwidth}{!}{
    \begin{tabular}{cc  ccc  c ccc}
    \toprule
     \multirow{2}{*}{\textbf{Shot}}  &\multirow{2}{*}{\textbf{Catergoies}} &\multicolumn{3}{c}{\textbf{Anomaly Classification}} & &\multicolumn{3}{c}{\textbf{Anomaly Segmentation}} \\
     \cline{3-5} \cline{7-9}\\
       & &\textbf{I-AUROC}	&\textbf{I-AUPR}& \textbf{I-$\text{F1}_{\text{max}}$}	&	&\textbf{P-AUROC}	&\textbf{P-AUPR}	&\textbf{P-$\text{F1}_{\text{max}}$}	\\
    \midrule
    \multirow{13}{*}{\textbf{0-shot}}
        & Woven\_001 & 100.0 & 99.8 & 100.0 && 99.8 & 78.2 & 71.7 \\
        & Woven\_127 & 94.8 & 95.4 & 88.4 && 94.5 & 54.0 & 54.6 \\
        & Woven\_104 & 98.9 & 99.7 & 98.1 && 95.2 & 67.6 & 64.8 \\
        & Stratified\_154 & 98.1 & 99.5 & 96.4 && 99.6 & 77.6 & 68.9 \\
        & Blotchy\_099 & 98.3 & 99.4 & 99.4 && 99.5 & 79.7 & 69.4 \\
\bottomrule
  \end{tabular}
  }
  \vspace{-10pt}
\end{table*}

\begin{table*}
  \setlength\tabcolsep{15pt}
  \centering
  \small
  \caption{\small{Image-level anomaly classification and pixel-level anomaly segmentation results on \textbf{MPDD} with zero-shot and few-shot AdaptCLIP.}}\label{tab:mpdd}
  \vspace{-5pt}
  \resizebox{1.0\textwidth}{!}{
    \begin{tabular}{cc  ccc  c ccc}
    \toprule
    \multirow{2}{*}{\textbf{Shot}}  &\multirow{2}{*}{\textbf{Catergoies}} &\multicolumn{3}{c}{\textbf{Anomaly Classification}} & &\multicolumn{3}{c}{\textbf{Anomaly Segmentation}} \\
     \cline{3-5} \cline{7-9}\\
      & &\textbf{I-AUROC}	&\textbf{I-AUPR}& \textbf{I-$\text{F1}_{\text{max}}$}	&	&\textbf{P-AUROC}	&\textbf{P-AUPR}	&\textbf{P-$\text{F1}_{\text{max}}$}	\\
    \midrule
     \multirow{7}{*}{\textbf{0-shot}}
        & bracket\_black &   65.8    &  73.1  &   77.9    &&   96.4    &  15.8  &   26.4 \\
        & bracket\_brown &   56.3    &  72.1  &   80.3    &&   92.6    &  2.9   &    6.7 \\
        & bracket\_white &   61.4    &  54.7  &   71.4    &&   99.5    &  4.0   &    9.1 \\
        &   connector   &   71.4    &  53.3  &   61.9    &&   96.1    &  14.4  &   20.5 \\
        &  metal\_plate  &   88.6    &  94.8  &   91.8    &&   93.3    &  60.4  &   61.5 \\
\bottomrule
  \end{tabular}
  }
  \vspace{-10pt}
\end{table*}

\begin{table*}
  \setlength\tabcolsep{18pt}
  \centering
  \small
  \caption{\small{Image-level anomaly classification and pixel-level anomaly segmentation results on \textbf{Real-IAD} with zero-shot and one-shot AdaptCLIP.}}\label{tab:realiad01}
  \vspace{-5pt}
  \resizebox{1.0\textwidth}{!}{
    \begin{tabular}{cc  ccc  c ccc}
    \toprule
     \multirow{2}{*}{\textbf{Shot}}  &\multirow{2}{*}{\textbf{Catergoies}} &\multicolumn{3}{c}{\textbf{Anomaly Classification}} & &\multicolumn{3}{c}{\textbf{Anomaly Segmentation}} \\
     \cline{3-5} \cline{7-9}\\
      & &\textbf{I-AUROC}	&\textbf{I-AUPR}& \textbf{I-$\text{F1}_{\text{max}}$}	&	&\textbf{P-AUROC}	&\textbf{P-AUPR}	&\textbf{P-$\text{F1}_{\text{max}}$}	\\
    \midrule
    \multirow{30}{*}{\textbf{0-shot}}
        & audiojack & 63.5 & 48.2 & 53.4 && 94.9 & 11.3 & 20.9 \\
        & bottle\_cap & 66.2 & 63.1 & 61.2 && 98.0 & 26.6 & 36.8 \\
        & button\_battery & 70.6 & 78.0 & 73.4 && 98.2 & 50.8 & 54.2 \\
        & end\_cap & 66.1 & 70.9 & 74.0 && 95.2 & 10.0 & 18.2 \\
        & eraser & 86.4 & 85.8 & 75.9 && 99.4 & 37.2 & 45.6 \\
\bottomrule
  \end{tabular}
  }
  \vspace{-10pt}
\end{table*}

\begin{table*}
  \setlength\tabcolsep{18pt}
  \centering
  \small
  \caption{\small{Image-level anomaly classification and pixel-level anomaly segmentation results on \textbf{Real-IAD} with 2-shot and 4-shot AdaptCLIP.}}\label{tab:realiad24}
  \vspace{-5pt}
  \resizebox{1.0\textwidth}{!}{
    \begin{tabular}{ccc  ccc  c ccc}
    \toprule
   \multirow{2}{*}{\textbf{Shot}}  &\multirow{2}{*}{\textbf{Catergoies}} &\multicolumn{3}{c}{\textbf{Anomaly Classification}} & &\multicolumn{3}{c}{\textbf{Anomaly Segmentation}} \\
     \cline{3-5} \cline{7-9}\\
      & &\textbf{I-AUROC}	&\textbf{I-AUPR}& \textbf{I-$\text{F1}_{\text{max}}$}	&	&\textbf{P-AUROC}	&\textbf{P-AUPR}	&\textbf{P-$\text{F1}_{\text{max}}$}	\\
    \midrule
    \multirow{30}{*}{\textbf{2-shot}}
        & audiojack & \pmerror{74.7}{1.1} & \pmerror{66.2}{0.9} & \pmerror{58.5}{1.4} && \pmerror{96.9}{0.0} & \pmerror{29.4}{4.2} & \pmerror{36.8}{4.3} \\
        & bottle\_cap & \pmerror{80.5}{0.6} & \pmerror{77.8}{0.8} & \pmerror{71.9}{0.7} && \pmerror{98.8}{0.0} & \pmerror{32.7}{0.8} & \pmerror{39.2}{0.6} \\
        & button\_battery & \pmerror{76.3}{1.0} & \pmerror{83.6}{0.6} & \pmerror{74.4}{0.5} && \pmerror{98.5}{0.0} & \pmerror{61.3}{0.4} & \pmerror{60.1}{0.4} \\
        & end\_cap & \pmerror{77.3}{0.4} & \pmerror{83.0}{0.1} & \pmerror{76.3}{0.3} && \pmerror{97.0}{0.0} & \pmerror{27.3}{0.2} & \pmerror{34.5}{0.5} \\
        & eraser & \pmerror{88.9}{1.0} & \pmerror{87.1}{0.9} & \pmerror{76.2}{1.3} && \pmerror{99.5}{0.0} & \pmerror{49.9}{0.0} & \pmerror{50.5}{0.0} \\
\bottomrule
  \end{tabular}
  }
  \vspace{-10pt}
\end{table*}

\begin{table*}
  \setlength\tabcolsep{15pt}
  \centering
  \small
  \caption{\small{Image-level anomaly classification and pixel-level anomaly segmentation results on \textbf{BTAD} with zero-shot and few-shot AdaptCLIP.}}\label{tab:btad}
  \vspace{-5pt}
  \resizebox{1.0\textwidth}{!}{
    \begin{tabular}{cc  ccc  c ccc}
    \toprule
     \multirow{2}{*}{\textbf{Shot}}  &\multirow{2}{*}{\textbf{Catergoies}} &\multicolumn{3}{c}{\textbf{Anomaly Classification}} & &\multicolumn{3}{c}{\textbf{Anomaly Segmentation}} \\
     \cline{3-5} \cline{7-9}\\
       & &\textbf{I-AUROC}	&\textbf{I-AUPR}& \textbf{I-$\text{F1}_{\text{max}}$}	&	&\textbf{P-AUROC}	&\textbf{P-AUPR}	&\textbf{P-$\text{F1}_{\text{max}}$}	\\
    \midrule
     \multirow{4}{*}{\textbf{0-shot}}
        & 01 & 96.1 & 98.6 & 95.8 && 93.3 & 51.1 & 54.0 \\
        & 02 & 81.1 & 96.9 & 93.2 && 93.4 & 60.5 & 61.6 \\
        & 03 & 95.9 & 81.0 & 79.5 && 94.7 & 13.8 & 23.4 \\
         \rowcolor[HTML]{EFEFEF} & mean & 91.0 & 92.2 & 89.5 && 93.8 & 41.8 & 46.3 \\
    \midrule
     \multirow{4}{*}{\textbf{1-shot}}
        & 01 & \pmerror{97.8}{0.5} & \pmerror{99.2}{0.2} & \pmerror{96.5}{1.0} && \pmerror{96.4}{0.1} & \pmerror{61.2}{0.3} & \pmerror{60.3}{0.5} \\
\bottomrule
  \end{tabular}
  }
  \vspace{-10pt}
\end{table*}

\begin{table*}
 \setlength\tabcolsep{15pt}
  \centering
  \small
  \caption{\small{Image-level anomaly classification and pixel-level anomaly segmentation results on \textbf{KSDD} with zero-shot and few-shot AdaptCLIP.}}\label{tab:ksdd}
  \vspace{-5pt}
  \resizebox{1.0\textwidth}{!}{
    \begin{tabular}{cc  ccc  c ccc}
    \toprule
     \multirow{2}{*}{\textbf{Shot}}  &\multirow{2}{*}{\textbf{Catergoies}} &\multicolumn{3}{c}{\textbf{Anomaly Classification}} & &\multicolumn{3}{c}{\textbf{Anomaly Segmentation}} \\
     \cline{3-5} \cline{7-9}\\
       & &\textbf{I-AUROC}	&\textbf{I-AUPR}& \textbf{I-$\text{F1}_{\text{max}}$}	&	&\textbf{P-AUROC}	&\textbf{P-AUPR}	&\textbf{P-$\text{F1}_{\text{max}}$}	\\
    \midrule
     {\textbf{0-shot}}
        &electrical commutators &   98.1    &  95.7  &   92.3    &&   98.1    &  58.3  &   58.1    \\
    \midrule
     {\textbf{1-shot}}
        & electrical commutators &   \pmerror{96.9}{0.3}    & \pmerror{91.8}{0.2}  & \pmerror{89.6}{0.5}    &&   \pmerror{98.2}{0.1}    & \pmerror{57.8}{1.2}   & \pmerror{58.4}{0.5}    \\
    \midrule
     {\textbf{2-shot}}
        & electrical commutators &   \pmerror{97.2}{0.0}    & \pmerror{92.4}{0.4}  & \pmerror{89.4}{0.9}    &&   \pmerror{98.1}{0.1}    & \pmerror{57.5}{1.1}   & \pmerror{58.1}{0.6}    \\
    \midrule
     {\textbf{4-shot}}
        & electrical commutators &   \pmerror{97.0}{0.2}    & \pmerror{91.7}{0.9}  & \pmerror{88.6}{1.7}    &&   \pmerror{98.0}{0.1}    & \pmerror{56.4}{1.4}   & \pmerror{57.2}{0.7}    \\
    \bottomrule
  \end{tabular}
  }
  \vspace{-10pt}
\end{table*}

\begin{table*}
  \setlength\tabcolsep{15pt}
  \centering
  \small
  \caption{\small{Image-level anomaly classification and pixel-level anomaly segmentation results on four \textbf{medical datasets, Br35H, Covid, Kvasir and Endo} with zero-shot and few-shot AdaptCLIP.}}\label{tab:medical}
  \vspace{-5pt}
  \resizebox{1.0\textwidth}{!}{
    \begin{tabular}{cc  ccc  c ccc}
    \toprule
    \multirow{2}{*}{\textbf{Shot}}  &\multirow{2}{*}{\textbf{Datasets}} &\multicolumn{3}{c}{\textbf{Anomaly Classification}} & &\multicolumn{3}{c}{\textbf{Anomaly Segmentation}} \\
     \cline{3-5} \cline{7-9}\\
       & &\textbf{I-AUROC}	&\textbf{I-AUPR}& \textbf{I-$\text{F1}_{\text{max}}$}	&	&\textbf{P-AUROC}	&\textbf{P-AUPR}	&\textbf{P-$\text{F1}_{\text{max}}$}	\\
    \midrule
     \multirow{4}{*}{\textbf{0-shot}}
        & Br35H & 94.8 & 95.1 & 87.7 && - & - & - \\
        & Covid & 86.5 & 54.4 & 55.4 && - & - & - \\
        & Kvasir & - & - & - && 82.1 & 45.3 & 50.6 \\
        & Endo & - & - & - && 86.5 & 52.0 & 54.5 \\
    \midrule
     \multirow{4}{*}{\textbf{1-shot}}
        & Br35H & \pmerror{93.7}{2.4} & \pmerror{92.1}{3.1} & \pmerror{87.8}{3.2} && - & - & - \\
\bottomrule
  \end{tabular}
  }
  \vspace{-10pt}
\end{table*}

